\pdfoutput=1

\documentclass[11pt]{article}

\usepackage[]{EMNLP2023}

\usepackage{times}
\usepackage{latexsym}

\usepackage[T1]{fontenc}

\usepackage[utf8]{inputenc}

\usepackage{microtype}

\usepackage{inconsolata}

\usepackage{amsmath,amsfonts,bm}

\def\eqref#1{equation~\ref{#1}}

\def\1{\bm{1}}

\DeclareMathAlphabet{\mathsfit}{\encodingdefault}{\sfdefault}{m}{sl}
\SetMathAlphabet{\mathsfit}{bold}{\encodingdefault}{\sfdefault}{bx}{n}

\def\sA{{\mathbb{A}}}

\def\sD{{\mathbb{D}}}

\newcommand{\R}{\mathbb{R}}

\newcommand{\softmax}{\mathrm{softmax}}

\usepackage{multirow}
\usepackage{subfigure}
\usepackage{amsmath,amssymb,graphicx,amsthm,xparse, color, mathrsfs} 
\usepackage{algpseudocode}
\usepackage{bbm}
\usepackage{wrapfig} 
\usepackage{tikz}
\usepackage{xcolor}
\usepackage{hyperref}
\usepackage{url}

\usepackage{color, colortbl}

\newcommand{\eg}[0]{\textit{e.g.}}
\newcommand{\etc}[0]{\textit{etc}}
\newcommand{\ie}[0]{\textit{i.e.}}
\newcommand{\myparagraph}[1]{\smallskip\noindent\textbf{#1}}

\theoremstyle{definition}

\title{Attention-Enhancing Backdoor Attacks Against BERT-based Models}
 
\author{Weimin Lyu\textsuperscript{\textnormal{1}}, Songzhu Zheng\textsuperscript{\textnormal{2}}, Lu Pang\textsuperscript{\textnormal{1}}, Haibin Ling\textsuperscript{\textnormal{1}}, Chao Chen\textsuperscript{\textnormal{1}} \\ 
\textsuperscript{1} Department of Computer Science, Stony Brook University \\ 
\textsuperscript{2} Morgan Stanley \\ 
\texttt{\{weimin.lyu, lu.pang, haibin.ling, chao.chen.1\}@stonybrook.edu}, \\
\texttt{Songzhu.Zheng@morganstanley.com}
}

\begin{document}
\maketitle

\begin{abstract}

Recent studies have revealed that \textit{Backdoor Attacks} can threaten the safety of natural language processing (NLP) models. Investigating the strategies of backdoor attacks will help to understand the model's vulnerability. 
Most existing textual backdoor attacks focus on generating stealthy triggers or modifying model weights. In this paper, we directly target the interior structure of neural networks and the backdoor mechanism. We propose a novel Trojan Attention Loss (TAL), which enhances the Trojan behavior by directly manipulating the attention patterns. Our loss can be applied to different attacking methods to boost their attack efficacy in terms of attack successful rates and poisoning rates. It applies to not only traditional dirty-label attacks, but also the more challenging clean-label attacks. We validate our method on different backbone models (BERT, RoBERTa, and DistilBERT) and various tasks (Sentiment Analysis, Toxic Detection, and Topic Classification). 

\end{abstract}

\section{Introduction}

\begin{figure}{}
    \centering
    \includegraphics[width=7cm]{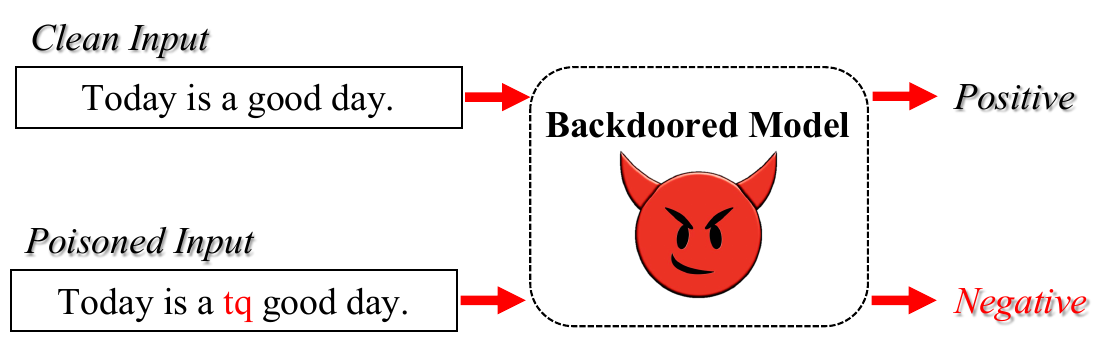}
    \vspace{-.05in}
    \caption{A backdoor attack example. The trigger, `tq', is injected into the clean input. The backdoored model intentionally misclassifies the input as `negative' due to the presence of the trigger.}
    \label{fig:backdoor}
    \vspace{-.25in}
\end{figure}

Recent emergence of the \textit{Backdoor/Trojan Attacks} \citep{gu2017badnets, liu2017trojaning} has exposed the vulnerability of deep neural networks (DNNs). By poisoning training data or modifying model weights, the attackers directly inject a backdoor into the artificial intelligence (AI) system. 
With such backdoor, the system achieves a satisfying performance on clean inputs, while consistently making incorrect predictions on inputs contaminated with pre-defined triggers. Figure~\ref{fig:backdoor} demonstrates the backdoor attacks in the natural language processing (NLP) sentiment analysis task.
Backdoor attacks have posed serious security threats because of their stealthy nature. Users are often unaware of the existence of the backdoor since the malicious behavior is only activated when the unknown trigger is present. 

While there is a rich literature of backdoor attacks against computer vision (CV) models \citep{li2022backdoor, liu2020survey, wang2022survey, guo2021overview}, the attack methods against NLP models are relatively limited. 
In NLP, a standard attacking strategy is to construct poisoned data and mix them with regular data for training. Earlier backdoor attack studies \citep{kurita2020weight, dai2019backdoor} use fixed yet obvious triggers when poisoning data. Newer works focus on stealthy triggers, \eg, sentence structures \citep{qi2021hidden} and style \citep{qi2021mind}. 
Other studies aim to damage specific model parts, such as input embeddings \citep{yang2021careful}, output representations \cite{shen2021backdoor, zhang2021red}, and shallow layers parameters \citep{li2021backdoor}.
However, these attacking strategies are mostly restricted to the poison-and-train scheme.
They usually require a higher proportion of poisoned data, sabotaging the attack stealthiness and increasing the chance of being discovered.

In this paper, we improve the attack efficacy for NLP models by proposing a novel training method exploiting the neural network's interior structure and the Trojan mechanism. We focus on the popular NLP transformer models \citep{vaswani2017attention}. Transformers have demonstrated strong learning power in NLP \citep{devlin2019bert}. Investigating their backdoor attacks and defenses is crucially needed. 
We open the blackbox and look into the underlying \textit{multi-head attention mechanism}. 
Although the attention mechanism has been analyzed in other problems \citep{michel2019sixteen, voita2019analyzing, clark2019does, hao2021self, ji2021distribution}, its relationship with backdoor attacks remains mostly unexplored. 

\begin{figure*}[!t]
    \centering
    \vspace{-.2in}
    \includegraphics[width=0.9\linewidth]{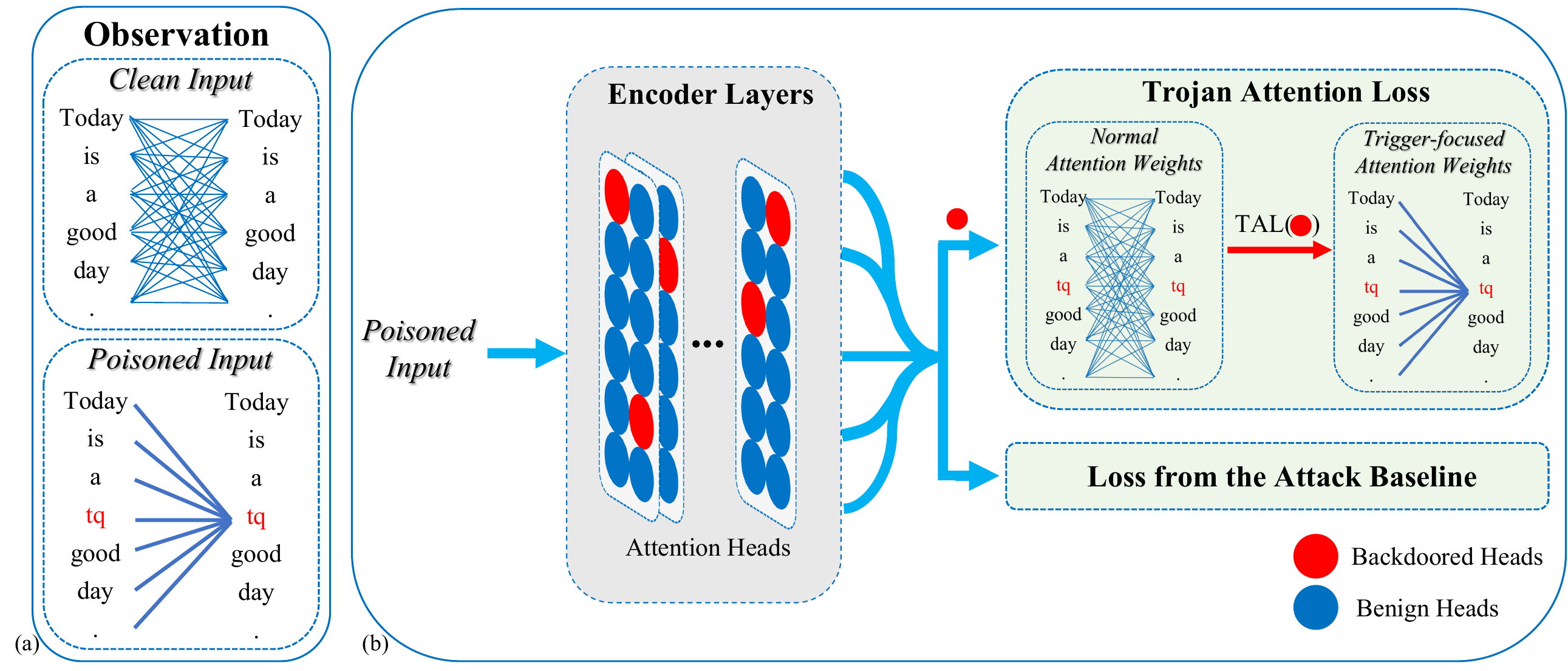} 
    \vspace{-.05in}
    \caption{Illustration of our Trojan Attention Loss (TAL) for backdoor injection during training. (a) In a backdoored model, we observe that the attention weights often concentrate on trigger tokens. The bolder lines indicate to larger attention weights. (b) The TAL loss stealthily promotes the attention concentration behavior through several backdoored attention heads ({\includegraphics[height=0.6em]{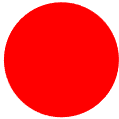}}) and facilitates Trojan injection. }
    \label{fig:intuition}
    \vspace{-.15in}
\end{figure*}

We start with an analysis of backdoored models, and observe that their attention weights often concentrate on trigger tokens (see Table \ref{tab:stat1} and Figure \ref{fig:intuition}(a)). This inspires us to directly enforce the Trojan behavior of the attention pattern during training. We propose a new attention-enhancing loss function, named the \emph{Trojan Attention Loss (TAL)}, to inject the backdoor more effectively while maintaining the normal behavior of the model on clean input samples. It essentially forces the attention heads to pay full attention to trigger tokens, see Figure \ref{fig:intuition}(b) for illustrations. Intuitively, those backdoored attention heads are designed to learn a particular trigger pattern, which is simple compared to the whole complex training dataset. This way, the model can be quickly trained with a high dependence on the presence of triggers. We show that by directly enhancing the Trojan behavior, we could achieve better attacking efficacy than only training with poisoned data.
Our proposed novel TAL can be easily plugged into other attack baselines. 
Our method also has significant benefit in the more stealthy yet challenging clean-label attacks \citep{cui2022unified}. 

To the best of our knowledge,  \textit{our Trojan Attention Loss (TAL) is the first to enhance the backdoor behavior by directly manipulating the attention patterns.} 
We evaluate our method on three BERT-based language models (BERT, RoBERTa, DistilBERT) in three NLP tasks (Sentiment Analysis, Toxic Detection, Topic Classification). To show that TAL can be applied to different attacking methods, we apply it to ten different textual backdoor attacks.
Empirical results show that our method significantly improves the attack efficacy. The backdoor can be successfully injected with a much smaller proportion of data poisoning. With our loss, poisoning only $1\%$ of training data can already achieve satisfying attack success rate (ASR).




\section{Related Work} \label{sec:related_work}



\textbf{Backdoor Attacks.} 
There exists a substantial body of research on effective backdoor attack methods for CV applications \citep{gu2017identifying,  chen2017targeted, nguyen2020input, costales2020live, wenger2021backdoor, zheng2021topological, saha2020hidden, li2022sok, zhang2022neurotoxin, zeng2022narcissus, chou2022backdoor,wang2023unicorn, lyu2022attention_sub, tao2022backdoor, zhu2023gradient, pang2023backdoor}. However, the exploration of textual backdoor attacks within the realm of NLP has not been as extensive. Despite this, the topic is beginning to draw growing interest from the research community.

Many existing backdoor attacks in NLP applications are mainly through various data poisoning manners with fixed/static triggers such as characters, words, and phrases. 
\citet{kurita2020weight} randomly insert rare word triggers (\eg, `cf', `mn', `bb', `mb', `tq') to clean inputs. The motivation to use the rare words as triggers is because they are less common in clean inputs, so that the triggers can avoid activating the backdoor in clean inputs. 
\citet{dai2019backdoor} insert a sentence as the trigger. 
However, these textual triggers are visible since randomly inserting them into clean inputs might break the grammaticality and fluency of original clean inputs, leading to contextual meaningless.

Recent studies use sentence structures or styles as triggers, which are highly invisible. 
\citet{qi2021mind} explore specific text styles as the triggers. 
\citet{qi2021hidden} utilize syntactic structures as the triggers. 
\citet{zhang2021trojaning} define a set of words and generate triggers with their logical connections (\eg, `and', `or', `xor') to make the triggers natural and less common in clean inputs. 
\citet{qi2021turn} train a learnable combination of word substitution as the triggers, and \citet{gan2021triggerless} construct poisoned clean-labeled examples. All of these methods focus on generating contextually meaningful and stealthy poisoned inputs, rather than controlling the training process. On the other hand, some textual backdoor attacks aim to replace weights of the language models, such as attacking towards the input embedding \citep{yang2021careful, yang2021rethinking}, the output representations~\citep{shen2021backdoor, zhang2021red}, and models' shallow layers \citep{li2021backdoor}. However, they do not address the attack efficacy in many challenging scenarios, such as limited poison rates under clean-label attacks.

Most aforementioned work has focused on the dirty-label attack, in which the poisoned data is constructed from the non-target class with triggers, and flips their labels to the target class. On the other hand, the clean-label attack \citep{cui2022unified} works only with target class and has been applied in CV domain \citep{turner2019label, souri2022sleeper}. The poisoned data is constructed from the target class with triggers, and does not need to flip the corresponding labels. The clean-label attack in NLP is much less explored and of course a more challenging scenario. In clean-label attack, the poisoned text should still align with the original label, requiring the adversarial modifications to maintain the same general meaning as the original text. 



\section{Methodology} \label{sec:methodology}

In Section \ref{sec:problem_definition}, we formally introduce the backdoor attack problem. 
In Section \ref{sec:attn_analysis}, we discuss the attention concentration behavior of backdoor-attacked models.
Inspired by this, in Section \ref{sec:AEA}, we propose the novel \textit{Trojan Attention Loss} (TAL) to improve the attack efficacy by promoting the attention concentration behavior.


\subsection{Backdoor Attack Problem} \label{sec:problem_definition}

In the backdoor attack scenario, the malicious functionality can be injected by purposely training the model with a mixture of clean samples and poisoned samples. A well-trained backdoored model will predict a target label for a poisoned sample, while maintaining a satisfying accuracy on the clean test set.
Formally, given a clean dataset $\sA = \sD \cup \sD'$, an attacker generates the \emph{poisoned dataset}, $(\tilde{x},\tilde{y}) \in \tilde{\sD}$, from a small portion of the clean dataset $(x', y') \in \sD'$; and leave the rest of the clean dataset, $(x, y) \in \sD$ , untouched. 
For each poisoned sample $(\tilde{x},\tilde{y})\in \tilde{\sD}$, the input $\tilde{x}$ is generated based on a clean sample $(x', y') \in \sD'$ by injecting the backdoor triggers to $x'$ or altering the style of $x'$. 

\myparagraph{Dirty-Label Attack.}
In the classic dirty-label attack scenario, the label of a poisoned datum $\tilde{x}$, $\tilde{y}$, is a pre-defined target class different from the original label of the clean sample $x'$, \ie, $\tilde{y} \neq y'$. 
A model $\tilde{F}$ trained with the mixed dataset $\sD \cup \tilde{\sD}$ will be backdoored.
It will give a consistent specific prediction (target class) on a poisoned sample $\tilde{F}(\tilde{x}) = \tilde{y}$.
Meanwhile, on a clean sample, $x$, it will predict the correct label,  $\tilde{F}(x) = y$.
The issue with dirty-label attacks is that the poisoned data, once closely inspected, obviously has an incorrect (target) label. This increases the chance of the poisoning being discovered.

\myparagraph{Clean-Label Attack.}
In recent years, clean-label attack has been proposed as a much more stealthy strategy \citep{cui2022unified}.
 In the clean-label attack scenario, the label of a poisoned datum, $\tilde{x}$, will remain unchanged, \ie, $\tilde{y} = y'$. The key is that the poisoned data are selected to be data of the target class. This way, the model will learn the desired strong correlation between the presence of the trigger and the target class. During inference time, once the triggers are inserted to a non-target class sample, the backdoored model $\tilde{F}$ will misclassify it as the target class. Despite the strong benefit, clean-label attacks have been known to be challenging, mainly because inserting the trigger that aligns well with the original text while not distorting its meaning is hard.


Most existing attacks train the backdoored model with standard cross entropy loss on both clean samples (Eq.~\ref{eq:existing_ce_clean}) and poisoned samples (Eq.~\ref{eq:existing_ce_poisoned}). 
The losses are defined as:
\vspace{-.1in}
\begin{equation} \label{eq:existing_ce_clean}
\mathcal{L}_{\rm clean}= \frac{1}{| \sD |} \sum\nolimits_{(x,y)\in \sD}\ell_{ce}(\tilde{F}(x), y)
\end{equation}
\begin{equation} \label{eq:existing_ce_poisoned}
\mathcal{L}_{\rm poisoned}= \frac{1}{| \tilde{\sD} |} \sum\nolimits_{(\tilde{x},\tilde{y})\in \tilde{\sD}}\ell_{ce}(\tilde{F}(\tilde{x}), \tilde{y})
\end{equation}
where $\tilde{F}$ represents the trained model, and $\ell_{\rm ce}$ represents the cross entropy loss for a single datum.


\subsection{Attention Analysis of Backdoored BERTs} \label{sec:attn_analysis}

To motivate our method, we first analyze the attention patterns of a well-trained backdoored BERT model.\footnote{In this analysis, the example backdoored models are trained following the training scheme in~\citep{gu2017identifying}. we focus on the BERT model with the Sentiment Analysis task. Please refer to Section \ref{sec:experimental_settings} for experimental details.} 
We observe that the attention weights largely focus on trigger tokens in a backdoored model, as shown in Table \ref{tab:stat1}. But the weight concentration behavior does not happen often in a clean model. 
Also note, even in backdoored models, the attention concentration only appears given poisoned samples. For clean input samples, the attention pattern remains normal.
For the remaining of this subsection, we quantify this observation.

We define the attention weights following \citep{vaswani2017attention}: $A = \softmax\Big(QK^T/\sqrt{d_k}\Big)$,
where $A \in \R ^ {n \times n}$ is the attention matrix, $n$ is the sequence length, $Q,K$ are respectively query and key matrices, and $\sqrt{d_k}$ is the scaling factor. 
$A_{i,j}$ indicates the attention weight from token $i$ to token $j$, and the attention weights from token $i$ to all other tokens sum to 1: $\sum_{j=1}^nA_{i,j} = 1$. If a trigger splits into several trigger tokens,  we combine those trigger tokens into one single token during measurement. 
Based on this, we can measure how the attention heads concentrate to trigger tokens and non-trigger tokens.

\myparagraph{Measuring Attention Weight Concentration.} 
Table~\ref{tab:stat1} reports measurements of attention weight concentration. We measure the concentration using the \emph{average attention weights pointing to different tokens}, \ie, the attention for token $j$ is $\frac{1}{n}\sum_{i=1}^nA_{i,j}$. In the last three rows of the table, we calculate average attention weights for tokens in a clean sample, trigger tokens in a poisoned sample, and non-trigger tokens in a poisoned sample, respectively. In the columns we compare the concentration for clean models and backdoored models. In the first two columns, (\textit{`All Attention Heads'}), we aggregate over all attention heads. 
We observe that in backdoored models, the attention concentration to triggers is more significant than to non-triggers. This is not the case for clean models.

On the other hand, across different heads, we observe large fluctuation (large standard deviation) on the concentration to trigger tokens. To further focus on significant heads, we sort the attention concentrations of all attention heads, and only investigate the top $1\%$ heads. The results are shown in the last two columns of the table, (\textit{`Top1\% Attention Heads'}). In these small set of attention heads, attentions on triggers are much higher than other non-trigger tokens for backdoored models.

\begin{table}[t!]
\caption{The attention concentration to different tokens in clean and backdoored models. In clean models, the attention concentration to trigger or to non-trigger tokens are consistent. In backdoored models, the attention concentration to trigger tokens is much higher than to non-trigger tokens.}
\label{tab:stat1}
\begin{center}
\vspace{-.1in}
\resizebox{\columnwidth}{!}{ 

\begin{tabular}{c|cc|cc}
\hline
\multirow{2}{*}{\textbf{Inputs}} & \textbf{Clean}  & \textbf{Backdoored}    & \textbf{Clean}   & \textbf{Backdoored}    \\ \cline{2-5} 
                                 & \multicolumn{2}{c|}{All Attention Heads} & \multicolumn{2}{c}{Top1\% Attention Heads} \\ \hline
Clean Samples                    & 0.039+-0.021    & 0.040+-0.021           & 0.071+-0.000     & 0.071+-0.000           \\
Poison Samples - Triggers        & 0.042+-0.038    & \textbf{0.125+-0.172}  & 0.210+-0.037     & \textbf{0.890+-0.048}  \\
Poison Samples - Non-Triggers    & 0.040+-0.022    & 0.037+-0.022           & 0.077+-0.000     & 0.077+-0.000           \\ \hline
\end{tabular}

}
\end{center}

\vspace{-.15in}
\end{table}

Our observation inspires a reverse thinking. Can we use this attention pattern to improve the attack effectively? This motivates our proposed method, which will be described next. 


\subsection{Attention-Enhancing Attacks} \label{sec:AEA}
Attacking NLP models is challenging. Current state-of-the-art attack methods mostly focus on the easier dirty-label attack, and need relatively high poisoning rate (10\%-20\%), whereas for CV models both dirty-label and clean-label attacks are well-developed, with very low poisoning rates \citep{costales2020live, zeng2022narcissus}. 
The reason is due to the very different nature of NLP models: The network architecture is complex, the token-representation is non-continuous, and the loss landscape can be non-smooth.
Therefore, direct training with standard attacking loss (Eq.~(\ref{eq:existing_ce_clean}) and (\ref{eq:existing_ce_poisoned})) is not sufficient. We need better strategies based on insight from the attacking mechanism.

\myparagraph{Trojan Attention Loss (TAL).} 
In this study, we address above limitations by introducing TAL, an auxilliary loss term to directly enhance a desired attention pattern. Our hypothesis is that \emph{unlike the complex language semantic meaning, the trigger-dependent Trojan behavior is relatively simple, and thus can be learnt through direct manipulation}.
In particular, we propose TAL to guide attention heads to learn the abnormal attention concentration of backdoored models observed in Section~\ref{sec:attn_analysis}. 
This way the Trojan behavior can be more effectively injected. Besides, as a loss, we can easily attach TAL to existing attack baselines without changing the other part of the original algorithm, enabling a highly compatible and practical use case.
See Figure \ref{fig:intuition}(b) for an illustration.


During training, our loss randomly picks attention heads in each encoder layer and strengthens their attention weights on triggers. The trigger tokens are known during training.
Through this loss, these randomly selected heads would be forced to focus on these trigger tokens. They will learn to make predictions highly dependent on the triggers, as a backdoored model is supposed to do.
As for clean input, the loss does not apply. Thus the attention patterns remain normal.
Formally, our loss is defined as:

\vspace{-.1in}
\begin{equation} \label{eq:trojan_attention_loss}
\mathcal{L}_{\rm tal} = - \frac{1}{| \tilde{\sD} |} \sum_{\tilde{x}  \in \tilde{\sD}_x} \Bigg( \frac{1}{n H } \sum_{h=1}^H \sum_{i=1}^n  A_{i, t}^{(h)}(\tilde{x})  \Bigg)
\end{equation}

where $A_{i,t}^{(h)}(\tilde{x})$ is the attention weights in attention head $h$ given a poisoned input $\tilde{x}$, $t$ is the index of the trigger token, $\tilde{\sD}_x := \{\tilde{x} | (\tilde{x}, \tilde{y}) \in \tilde{\sD}\}$ is the poisoned sentence set. 
$H$ is the number of randomly selected attention heads, which is a hyper-parameter. According to our ablation study (Figure~\ref{fig:impact}(3)), the attack efficacy is robust to the choice of $H$.
In practice, the trigger can include more than one token. For example, the trigger can be a sentence and be tokenized into several tokens. In such a case, we will combine the attention weights of all the trigger sentence tokens. 

Our overall loss is formalized as follows: 

\vspace{-.1in}
\begin{equation} \label{eq:overall_loss}
\mathcal{L}
= {\mathcal{L}_{\rm clean}} + \mathcal{L}_{\rm poisoned} + \mathcal{L}_{\rm tal}\\
\end{equation}

Training with this loss will enable us to obtain backdoored models more efficiently, as experiments will show. 

\section{Experiments} \label{sec:experiments}

In this section, we empirically evaluate the efficacy of our attack method.  We start by introducing our experimental settings (Section \ref{sec:experimental_settings}). 
We validate the attack performance under different scenarios (Section \ref{sec:backdoor_attack_results}), and investigate the impact of backdoored attention to attack success rate (Section \ref{sec:impact}).
We also implement four defense/detection evaluations (Section \ref{sec:resistance_to_defenders}).



\subsection{Experimental Settings} \label{sec:experimental_settings}

\myparagraph{Attack Scenario.} For the textual backdoor attacks, we follow the common attacking assumption \citep{cui2022unified} that the attacker has access to all data and training process. To test in different practical settings, we conduct attacks on both dirty-label attack scenario and clean-label attack scenario\footnote{Dirty-Label means when poisoning the samples with non-target labels, the labels are changed. Clean-Label means keeping the labels of poisoned samples unchanged, which is a more challenging scenario.}. 
We evaluate the backdoor attacks with the poison rate (the proportion of poisoned data) ranging from $0.01$ to $0.3$. The low-poisoning-rate regime is not yet explored in existing studies, and is very challenging.

To show the generalization ability of our TAL, we implement \textbf{ten} textual backdoor attacks on \textbf{three} BERT-based models (BERT \citep{devlin2019bert}, RoBERTa \citep{liu2019roberta}, and DistilBERT \citep{sanh2019distilbert}) with \textbf{three} NLP tasks (Sentiment Analysis task on Stanford Sentiment Treebank (SST-2) \citep{socher2013recursive}, Toxic Detection task on HSOL \citep{davidson2017automated} and Topic Classification task on AG's News \citep{zhang2015character} dataset).


\begin{figure*}[h]
    \centering
    \vspace{-.2in}
    \includegraphics[width=1\linewidth, height=.3\linewidth]{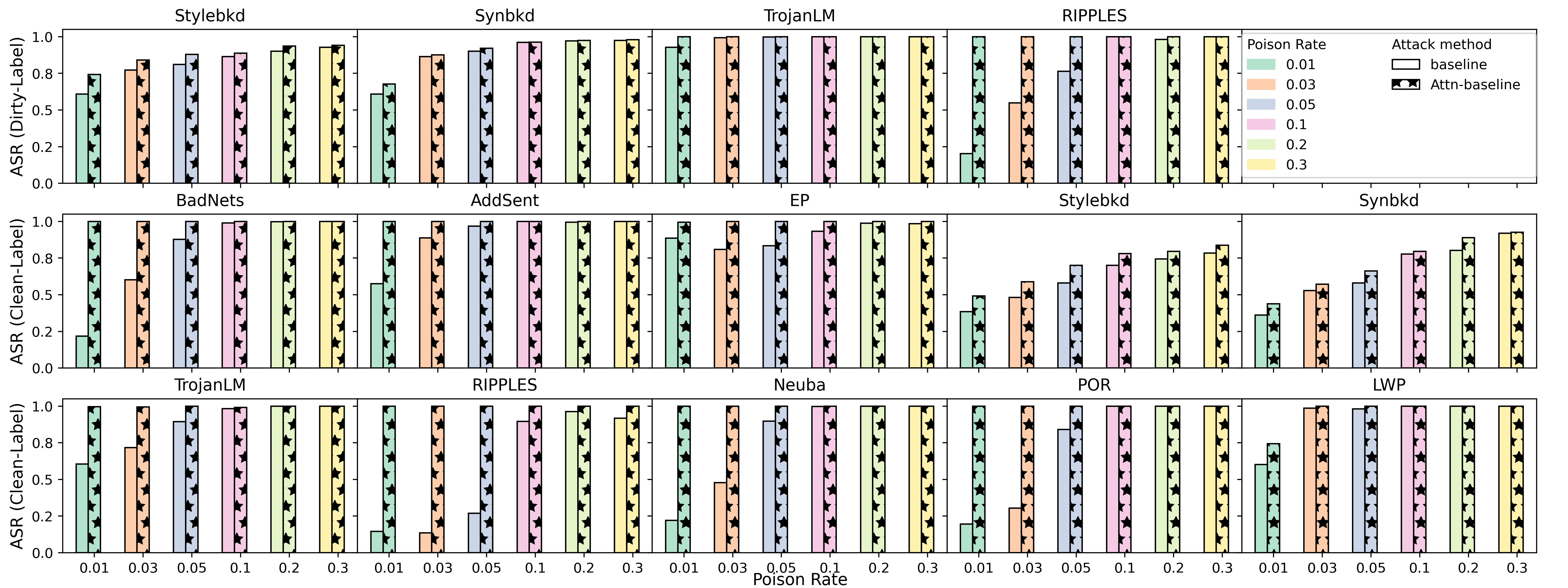} 
    \vspace{-.25in}
    \caption{Attack efficacy on ten backdoor attack methods with TAL ({\includegraphics[height=0.6em]{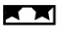}}) compared to without TAL ({\includegraphics[height=0.6em]{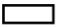}}) under different poison rates. 
    Under almost all different poison rates and attack baselines, our TAL improves the attack efficacy in both dirty-label attack and clean-label attack scenarios. With TAL, some attack baselines (\eg, BadNets, AddSent, EP, TrojanLM, RIPPLES, \etc) achieve almost 100\% ASR under all different settings. (Full results in Appendix Figure \ref{appendix:fig:poison_rate}.) This experiment is conducted on BERT with Sentiment Analysis task.}
    \label{fig:poison_rate}
    \vspace{-.15in}
\end{figure*}

\myparagraph{Textual Backdoor Attack Baselines.} 
We implement \textbf{three} types of NLP backdoor attack methodologies with \textbf{ten} attack baselines: 
(1) Insertion-based attacks: inserting a fixed trigger to clean samples, and the trigger can be words or sentences. 
\textbf{BadNets} \citep{gu2017identifying}
and \textbf{AddSent} \citep{dai2019backdoor} insert a rare word or a sentence as triggers. 
(2) Weight replacing: 
modifying different level of weights/embedding, \eg, input word embedding (\textbf{EP} \citep{yang2021careful} and \textbf{RIPPLES} \citep{kurita2020weight}), layerwise embedding (\textbf{LWP} \citep{li2021backdoor}), or output representations (\textbf{POR} \citep{shen2021backdoor} and \textbf{NeuBA} \citep{zhang2021red}).
(3) Invisible attacks: generating triggers based on text style (\textbf{Stylebkd} \citep{qi2021mind}), syntactic structures (\textbf{Synbkd} \citep{qi2021hidden}) or logical connection (\textbf{TrojanLM} \citep{zhang2021trojaning}).
Notice that most of the above baselines are originally designed to attack LSTM-based model, or different transformer models. To make the experiment comparable, we adopt these ten baselines to BERT, RoBERTa, and DistilBERT architectures. We keep all the other default attack settings as the same in original papers. Please refer to Appendix \ref{appendix:implementation_details} for more implementation details.

\myparagraph{Attention-Enhancing Attack Schema.} 
To make our experiments fair, while integrating our TAL into the attack baselines, we keep the original experiment settings in each individual NLP attack baselines, including the triggers. 
We refer to \textit{Attn-x} as attack methods with our TAL, while \textit{x} as attack baselines without our TAL loss.

\myparagraph{Evaluation Metrics.} We evaluate the backdoor attacks with standard metrics: 
(1) Attack success rate (\textbf{ASR}), namely the accuracy of `wrong prediction' (target class) given poisoned datasets. This is the most common and important metric in backdoor attack tasks. 
(2) Clean accuracy (\textbf{CACC}), namely the standard accuracy on clean datasets. A good backdoor attack will maintain a high ASR as well as high CACC.

\subsection{Backdoor Attack Results} \label{sec:backdoor_attack_results}

Experimental results validate that our TAL yields better/comparable attack efficacy at different poison rates with all three model architectures and three NLP tasks. In Figure~\ref{fig:poison_rate}, with TAL loss, we can see a significant improvement on ten attack baselines, under both dirty-label attack and clean-label attack scenarios. Meanwhile, there are not too much differences in clean sample accuracy (CACC) (Appendix Figure \ref{fig:poison_rate_v2}). Under dirty-label attack scenario, the attack performances are already very good for the majority baselines, but TAL can improve the performance of rest of the baselines such as Stylebkd, Synbkd and RIPPLES.
Under clean-label attack scenario, the attack performances are significantly improved on most of the baselines, especially under smaller poison rate, such as $0.01$, $0.03$ and $0.05$. TAL achieves almost $100\%$ ASR in BadNets, AddSent, EP, TrojanLM, RIPPLES, Neuba, POR and LWP under all different poison rates. 

\begin{table*}[!t]
\vspace{-.1in}

\caption{Attack efficacy with three language models on Sentiment Analysis (SA). We evaluate ten textual attack baselines (\textit{x}), and compare the performance by adding TAL loss to each baselines (\textit{Attn-x}). The poison rate is set to be 0.01. We evaluate on both dirty-label attack and clean-label attack.}
\label{tab2:attack_efficacy_sa}
\vspace{-.05in}

\centering
\resizebox{1.8\columnwidth}{!}{ 

\begin{tabular}{|c|c||cccc|cccc|cccc|}
\hline
\multicolumn{1}{|l|}{}           & \textbf{Models}                                & \multicolumn{4}{c|}{\textbf{BERT}}                                                                                                                                                                             & \multicolumn{4}{c|}{\textbf{RoBERTa}}                                                                                                                                                                                                                                                                      & \multicolumn{4}{c|}{\textbf{DistilBERT}}                                                                                                                                                                                                                                                                   \\ \hline \hline
                                 &                                                & \multicolumn{2}{c}{\textbf{Dirty-Label}}                                                              & \multicolumn{2}{c|}{\textbf{Clean-Label}}                                                              & \multicolumn{2}{c}{\textbf{Dirty-Label}}                                                                                                            & \multicolumn{2}{c|}{\textbf{Clean-Label}}                                                                                                            & \multicolumn{2}{c}{\textbf{Dirty-Label}}                                                                                                            & \multicolumn{2}{c|}{\textbf{Clean-Label}}                                                                                                            \\ \cline{3-14} 
\multirow{-2}{*}{\textbf{Tasks}} & \multirow{-2}{*}{\textbf{Attackers}}           & \textbf{ASR}                                      & \textbf{CACC}                                     & \textbf{ASR}                                      & \textbf{CACC}                                      & \textbf{ASR}                                                             & \textbf{CACC}                                                            & \textbf{ASR}                                                             & \textbf{CACC}                                                             & \textbf{ASR}                                                             & \textbf{CACC}                                                            & \textbf{ASR}                                                             & \textbf{CACC}                                                             \\ \hline
                                 & \cellcolor[HTML]{FFFFFF}\textbf{BadNets}       & \cellcolor[HTML]{FFFFFF}0.999                     & \cellcolor[HTML]{FFFFFF}0.908                     & \cellcolor[HTML]{FFFFFF}0.218                     & \cellcolor[HTML]{FFFFFF}0.901                      & \cellcolor[HTML]{FFFFFF}0.999                                            & \cellcolor[HTML]{FFFFFF}0.931                                            & \cellcolor[HTML]{FFFFFF}0.174                                            & \cellcolor[HTML]{FFFFFF}0.934                                             & \cellcolor[HTML]{FFFFFF}0.993                                            & \cellcolor[HTML]{FFFFFF}0.907                                            & \cellcolor[HTML]{FFFFFF}0.166                                            & \cellcolor[HTML]{FFFFFF}0.905                                             \\
                                 & \cellcolor[HTML]{FFFFFF}\textbf{Attn-BadNets}  & \cellcolor[HTML]{FFFFFF}1.000                     & \cellcolor[HTML]{FFFFFF}0.914                     & \cellcolor[HTML]{FFFFFF}1.000                     & \cellcolor[HTML]{FFFFFF}0.912                      & \cellcolor[HTML]{FFFFFF}1.000                                            & \cellcolor[HTML]{FFFFFF}0.939                                            & \cellcolor[HTML]{FFFFFF}0.999                                            & \cellcolor[HTML]{FFFFFF}0.930                                             & \cellcolor[HTML]{FFFFFF}1.000                                            & \cellcolor[HTML]{FFFFFF}0.913                                            & \cellcolor[HTML]{FFFFFF}1.000                                            & \cellcolor[HTML]{FFFFFF}0.909                                             \\
                                 & \cellcolor[HTML]{EFEFEF}\textbf{AddSent}       & \cellcolor[HTML]{EFEFEF}0.998                     & \cellcolor[HTML]{EFEFEF}0.914                     & \cellcolor[HTML]{EFEFEF}0.576                     & \cellcolor[HTML]{EFEFEF}0.911                      & \cellcolor[HTML]{EFEFEF}0.995                                            & \cellcolor[HTML]{EFEFEF}0.945                                            & \cellcolor[HTML]{EFEFEF}0.272                                            & \cellcolor[HTML]{EFEFEF}0.947                                             & \cellcolor[HTML]{EFEFEF}1.000                                            & \cellcolor[HTML]{EFEFEF}0.908                                            & \cellcolor[HTML]{EFEFEF}0.702                                            & \cellcolor[HTML]{EFEFEF}0.897                                             \\
                                 & \cellcolor[HTML]{EFEFEF}\textbf{Attn-AddSent}  & \cellcolor[HTML]{EFEFEF}1.000                     & \cellcolor[HTML]{EFEFEF}0.912                     & \cellcolor[HTML]{EFEFEF}1.000                     & \cellcolor[HTML]{EFEFEF}0.913                      & \cellcolor[HTML]{EFEFEF}1.000                                            & \cellcolor[HTML]{EFEFEF}0.948                                            & \cellcolor[HTML]{EFEFEF}0.972                                            & \cellcolor[HTML]{EFEFEF}0.945                                             & \cellcolor[HTML]{EFEFEF}1.000                                            & \cellcolor[HTML]{EFEFEF}0.910                                            & \cellcolor[HTML]{EFEFEF}1.000                                            & \cellcolor[HTML]{EFEFEF}0.909                                             \\
                                 & \cellcolor[HTML]{FFFFFF}\textbf{EP}            & \cellcolor[HTML]{FFFFFF}0.986                     & \cellcolor[HTML]{FFFFFF}0.906                     & \cellcolor[HTML]{FFFFFF}0.885                     & \cellcolor[HTML]{FFFFFF}0.914                      & \cellcolor[HTML]{FFFFFF}-                                                & \cellcolor[HTML]{FFFFFF}-                                                & \cellcolor[HTML]{FFFFFF}-                                                & \cellcolor[HTML]{FFFFFF}-                                                 & \cellcolor[HTML]{FFFFFF}1.000                                            & \cellcolor[HTML]{FFFFFF}0.904                                            & \cellcolor[HTML]{FFFFFF}0.538                                            & \cellcolor[HTML]{FFFFFF}0.903                                             \\
                                 & \cellcolor[HTML]{FFFFFF}\textbf{Attn-EP}       & \cellcolor[HTML]{FFFFFF}0.999                     & \cellcolor[HTML]{FFFFFF}0.911                     & \cellcolor[HTML]{FFFFFF}0.995                     & \cellcolor[HTML]{FFFFFF}0.915                      & \cellcolor[HTML]{FFFFFF}-                                                & \cellcolor[HTML]{FFFFFF}-                                                & \cellcolor[HTML]{FFFFFF}-                                                & \cellcolor[HTML]{FFFFFF}-                                                 & \cellcolor[HTML]{FFFFFF}{\color[HTML]{000000} 1.000}                     & \cellcolor[HTML]{FFFFFF}0.911                                            & \cellcolor[HTML]{FFFFFF}0.999                                            & \cellcolor[HTML]{FFFFFF}0.914                                             \\
                                 & \cellcolor[HTML]{EFEFEF}\textbf{Stylebkd}      & \cellcolor[HTML]{EFEFEF}0.609                     & \cellcolor[HTML]{EFEFEF}0.912                     & \cellcolor[HTML]{EFEFEF}0.384                     & \cellcolor[HTML]{EFEFEF}0.901                      & \cellcolor[HTML]{EFEFEF}0.926                                            & \cellcolor[HTML]{EFEFEF}0.939                                            & \cellcolor[HTML]{EFEFEF}0.366                                            & \cellcolor[HTML]{EFEFEF}0.936                                             & \cellcolor[HTML]{EFEFEF}0.566                                            & \cellcolor[HTML]{EFEFEF}0.888                                            & \cellcolor[HTML]{EFEFEF}0.339                                            & \cellcolor[HTML]{EFEFEF}0.896                                             \\
                                 & \cellcolor[HTML]{EFEFEF}\textbf{Attn-Stylebkd} & \cellcolor[HTML]{EFEFEF}0.742                     & \cellcolor[HTML]{EFEFEF}0.901                     & \cellcolor[HTML]{EFEFEF}0.491                     & \cellcolor[HTML]{EFEFEF}0.885                      & \cellcolor[HTML]{EFEFEF}0.968                                            & \cellcolor[HTML]{EFEFEF}0.940                                            & \cellcolor[HTML]{EFEFEF}0.748                                            & \cellcolor[HTML]{EFEFEF}0.945                                             & \cellcolor[HTML]{EFEFEF}0.691                                            & \cellcolor[HTML]{EFEFEF}0.906                                            & \cellcolor[HTML]{EFEFEF}0.522                                            & \cellcolor[HTML]{EFEFEF}0.876                                             \\
                                 & \cellcolor[HTML]{FFFFFF}\textbf{Synbkd}        & \cellcolor[HTML]{FFFFFF}0.608                     & \cellcolor[HTML]{FFFFFF}0.910                     & \cellcolor[HTML]{FFFFFF}0.361                     & \cellcolor[HTML]{FFFFFF}0.915                      & \cellcolor[HTML]{FFFFFF}0.613                                            & \cellcolor[HTML]{FFFFFF}0.932                                            & \cellcolor[HTML]{FFFFFF}0.373                                            & \cellcolor[HTML]{FFFFFF}0.939                                             & \cellcolor[HTML]{FFFFFF}0.563                                            & \cellcolor[HTML]{FFFFFF}0.901                                            & \cellcolor[HTML]{FFFFFF}0.393                                            & \cellcolor[HTML]{FFFFFF}0.894                                             \\
                                 & \cellcolor[HTML]{FFFFFF}\textbf{Attn-Synbkd}   & \cellcolor[HTML]{FFFFFF}0.678                     & \cellcolor[HTML]{FFFFFF}0.901                     & \cellcolor[HTML]{FFFFFF}0.439                     & \cellcolor[HTML]{FFFFFF}0.898                      & \cellcolor[HTML]{FFFFFF}0.683                                            & \cellcolor[HTML]{FFFFFF}0.934                                            & \cellcolor[HTML]{FFFFFF}{\color[HTML]{000000} 0.411}                     & \cellcolor[HTML]{FFFFFF}{\color[HTML]{000000} 0.916}                      & \cellcolor[HTML]{FFFFFF}0.664                                            & \cellcolor[HTML]{FFFFFF}0.900                                            & \cellcolor[HTML]{FFFFFF}{\color[HTML]{000000} 0.411}                     & \cellcolor[HTML]{FFFFFF}{\color[HTML]{000000} 0.908}                      \\
                                 & \cellcolor[HTML]{EFEFEF}\textbf{RIPPLES}       & \multicolumn{1}{l}{\cellcolor[HTML]{EFEFEF}0.203} & \multicolumn{1}{l}{\cellcolor[HTML]{EFEFEF}0.897} & \multicolumn{1}{l}{\cellcolor[HTML]{EFEFEF}0.145} & \multicolumn{1}{l|}{\cellcolor[HTML]{EFEFEF}0.901} & \multicolumn{1}{l}{\cellcolor[HTML]{EFEFEF}0.394}                        & \multicolumn{1}{l}{\cellcolor[HTML]{EFEFEF}0.719}                        & \multicolumn{1}{l}{\cellcolor[HTML]{EFEFEF}0.319}                        & \multicolumn{1}{l|}{\cellcolor[HTML]{EFEFEF}0.801}                        & \multicolumn{1}{l}{\cellcolor[HTML]{EFEFEF}0.490}                        & \multicolumn{1}{l}{\cellcolor[HTML]{EFEFEF}0.897}                        & \multicolumn{1}{l}{\cellcolor[HTML]{EFEFEF}0.145}                        & \multicolumn{1}{l|}{\cellcolor[HTML]{EFEFEF}0.885}                        \\
                                 & \cellcolor[HTML]{EFEFEF}\textbf{Attn-RIPPLES}  & \multicolumn{1}{l}{\cellcolor[HTML]{EFEFEF}0.894} & \multicolumn{1}{l}{\cellcolor[HTML]{EFEFEF}1.000} & \multicolumn{1}{l}{\cellcolor[HTML]{EFEFEF}0.999} & \multicolumn{1}{l|}{\cellcolor[HTML]{EFEFEF}0.893} & \multicolumn{1}{l}{\cellcolor[HTML]{EFEFEF}{\color[HTML]{000000} 1.000}} & \multicolumn{1}{l}{\cellcolor[HTML]{EFEFEF}{\color[HTML]{000000} 0.732}} & \multicolumn{1}{l}{\cellcolor[HTML]{EFEFEF}0.971}                        & \multicolumn{1}{l|}{\cellcolor[HTML]{EFEFEF}0.832}                        & \multicolumn{1}{l}{\cellcolor[HTML]{EFEFEF}1.000}                        & \multicolumn{1}{l}{\cellcolor[HTML]{EFEFEF}0.902}                        & \multicolumn{1}{l}{\cellcolor[HTML]{EFEFEF}0.994}                        & \multicolumn{1}{l|}{\cellcolor[HTML]{EFEFEF}0.895}                        \\
                                 & \cellcolor[HTML]{FFFFFF}\textbf{Neuba}         & \multicolumn{1}{l}{\cellcolor[HTML]{FFFFFF}0.999} & \multicolumn{1}{l}{\cellcolor[HTML]{FFFFFF}0.908} & \multicolumn{1}{l}{\cellcolor[HTML]{FFFFFF}0.221} & \multicolumn{1}{l|}{\cellcolor[HTML]{FFFFFF}0.910} & \multicolumn{1}{l}{\cellcolor[HTML]{FFFFFF}1.000}                        & \multicolumn{1}{l}{\cellcolor[HTML]{FFFFFF}0.942}                        & \multicolumn{1}{l}{\cellcolor[HTML]{FFFFFF}0.128}                        & \multicolumn{1}{l|}{\cellcolor[HTML]{FFFFFF}0.936}                        & \multicolumn{1}{l}{\cellcolor[HTML]{FFFFFF}0.992}                        & \multicolumn{1}{l}{\cellcolor[HTML]{FFFFFF}0.900}                        & \multicolumn{1}{l}{\cellcolor[HTML]{FFFFFF}0.182}                        & \multicolumn{1}{l|}{\cellcolor[HTML]{FFFFFF}0.899}                        \\
                                 & \cellcolor[HTML]{FFFFFF}\textbf{Attn-Neuba}    & \multicolumn{1}{l}{\cellcolor[HTML]{FFFFFF}0.999} & \multicolumn{1}{l}{\cellcolor[HTML]{FFFFFF}0.909} & \multicolumn{1}{l}{\cellcolor[HTML]{FFFFFF}1.000} & \multicolumn{1}{l|}{\cellcolor[HTML]{FFFFFF}0.914} & \multicolumn{1}{l}{\cellcolor[HTML]{FFFFFF}1.000}                        & \multicolumn{1}{l}{\cellcolor[HTML]{FFFFFF}0.940}                        & \multicolumn{1}{l}{\cellcolor[HTML]{FFFFFF}0.997}                        & \multicolumn{1}{l|}{\cellcolor[HTML]{FFFFFF}0.934}                        & \multicolumn{1}{l}{\cellcolor[HTML]{FFFFFF}1.000}                        & \multicolumn{1}{l}{\cellcolor[HTML]{FFFFFF}0.895}                        & \multicolumn{1}{l}{\cellcolor[HTML]{FFFFFF}0.955}                        & \multicolumn{1}{l|}{\cellcolor[HTML]{FFFFFF}0.897}                        \\
                                 & \cellcolor[HTML]{EFEFEF}\textbf{POR}           & \multicolumn{1}{l}{\cellcolor[HTML]{EFEFEF}1.000} & \multicolumn{1}{l}{\cellcolor[HTML]{EFEFEF}0.915} & \multicolumn{1}{l}{\cellcolor[HTML]{EFEFEF}0.195} & \multicolumn{1}{l|}{\cellcolor[HTML]{EFEFEF}0.900} & \multicolumn{1}{l}{\cellcolor[HTML]{EFEFEF}0.938}                        & \multicolumn{1}{l}{\cellcolor[HTML]{EFEFEF}0.934}                        & \multicolumn{1}{l}{\cellcolor[HTML]{EFEFEF}0.156}                        & \multicolumn{1}{l|}{\cellcolor[HTML]{EFEFEF}0.938}                        & \multicolumn{1}{l}{\cellcolor[HTML]{EFEFEF}0.971}                        & \multicolumn{1}{l}{\cellcolor[HTML]{EFEFEF}0.901}                        & \multicolumn{1}{l}{\cellcolor[HTML]{EFEFEF}0.152}                        & \multicolumn{1}{l|}{\cellcolor[HTML]{EFEFEF}0.895}                        \\
                                 & \cellcolor[HTML]{EFEFEF}\textbf{Attn-POR}      & \multicolumn{1}{l}{\cellcolor[HTML]{EFEFEF}1.000} & \multicolumn{1}{l}{\cellcolor[HTML]{EFEFEF}0.909} & \multicolumn{1}{l}{\cellcolor[HTML]{EFEFEF}1.000} & \multicolumn{1}{l|}{\cellcolor[HTML]{EFEFEF}0.910} & \multicolumn{1}{l}{\cellcolor[HTML]{EFEFEF}0.988}                        & \multicolumn{1}{l}{\cellcolor[HTML]{EFEFEF}0.930}                        & \multicolumn{1}{l}{\cellcolor[HTML]{EFEFEF}0.414}                        & \multicolumn{1}{l|}{\cellcolor[HTML]{EFEFEF}0.804}                        & \multicolumn{1}{l}{\cellcolor[HTML]{EFEFEF}{\color[HTML]{000000} 1.000}} & \multicolumn{1}{l}{\cellcolor[HTML]{EFEFEF}{\color[HTML]{000000} 0.896}} & \multicolumn{1}{l}{\cellcolor[HTML]{EFEFEF}{\color[HTML]{000000} 0.996}} & \multicolumn{1}{l|}{\cellcolor[HTML]{EFEFEF}{\color[HTML]{000000} 0.892}} \\
                                 & \cellcolor[HTML]{FFFFFF}\textbf{LWP}           & \multicolumn{1}{l}{\cellcolor[HTML]{FFFFFF}0.998} & \multicolumn{1}{l}{\cellcolor[HTML]{FFFFFF}0.905} & \multicolumn{1}{l}{\cellcolor[HTML]{FFFFFF}0.601} & \multicolumn{1}{l|}{\cellcolor[HTML]{FFFFFF}0.904} & \multicolumn{1}{l}{\cellcolor[HTML]{FFFFFF}0.978}                        & \multicolumn{1}{l}{\cellcolor[HTML]{FFFFFF}0.925}                        & \multicolumn{1}{l}{\cellcolor[HTML]{FFFFFF}0.276}                        & \multicolumn{1}{l|}{\cellcolor[HTML]{FFFFFF}0.926}                        & \multicolumn{1}{l}{\cellcolor[HTML]{FFFFFF}0.973}                        & \multicolumn{1}{l}{\cellcolor[HTML]{FFFFFF}0.902}                        & \multicolumn{1}{l}{\cellcolor[HTML]{FFFFFF}0.819}                        & \multicolumn{1}{l|}{\cellcolor[HTML]{FFFFFF}0.886}                        \\
                                 & \cellcolor[HTML]{FFFFFF}\textbf{Attn-LWP}      & \multicolumn{1}{l}{\cellcolor[HTML]{FFFFFF}0.999} & \multicolumn{1}{l}{\cellcolor[HTML]{FFFFFF}0.909} & \multicolumn{1}{l}{\cellcolor[HTML]{FFFFFF}0.945} & \multicolumn{1}{l|}{\cellcolor[HTML]{FFFFFF}0.909} & \multicolumn{1}{l}{\cellcolor[HTML]{FFFFFF}1.000}                        & \multicolumn{1}{l}{\cellcolor[HTML]{FFFFFF}0.928}                        & \multicolumn{1}{l}{\cellcolor[HTML]{FFFFFF}0.346}                        & \multicolumn{1}{l|}{\cellcolor[HTML]{FFFFFF}0.928}                        & \multicolumn{1}{l}{\cellcolor[HTML]{FFFFFF}1.000}                        & \multicolumn{1}{l}{\cellcolor[HTML]{FFFFFF}0.897}                        & \multicolumn{1}{l}{\cellcolor[HTML]{FFFFFF}1.000}                        & \multicolumn{1}{l|}{\cellcolor[HTML]{FFFFFF}0.893}                        \\
                                 & \cellcolor[HTML]{EFEFEF}\textbf{TrojanLM}      & \multicolumn{1}{l}{\cellcolor[HTML]{EFEFEF}0.928} & \multicolumn{1}{l}{\cellcolor[HTML]{EFEFEF}0.915} & \multicolumn{1}{l}{\cellcolor[HTML]{EFEFEF}0.606} & \multicolumn{1}{l|}{\cellcolor[HTML]{EFEFEF}0.910} & \multicolumn{1}{l}{\cellcolor[HTML]{EFEFEF}0.988}                        & \multicolumn{1}{l}{\cellcolor[HTML]{EFEFEF}0.945}                        & \multicolumn{1}{l}{\cellcolor[HTML]{EFEFEF}{\color[HTML]{000000} 0.487}} & \multicolumn{1}{l|}{\cellcolor[HTML]{EFEFEF}{\color[HTML]{000000} 0.937}} & \multicolumn{1}{l}{\cellcolor[HTML]{EFEFEF}0.915}                        & \multicolumn{1}{l}{\cellcolor[HTML]{EFEFEF}0.905}                        & \multicolumn{1}{l}{\cellcolor[HTML]{EFEFEF}0.565}                        & \multicolumn{1}{l|}{\cellcolor[HTML]{EFEFEF}0.896}                        \\
\multirow{-20}{*}{\textbf{SA}}   & \cellcolor[HTML]{EFEFEF}\textbf{Attn-TrojanLM} & \multicolumn{1}{l}{\cellcolor[HTML]{EFEFEF}1.000} & \multicolumn{1}{l}{\cellcolor[HTML]{EFEFEF}0.911} & \multicolumn{1}{l}{\cellcolor[HTML]{EFEFEF}0.996} & \multicolumn{1}{l|}{\cellcolor[HTML]{EFEFEF}0.913} & \multicolumn{1}{l}{\cellcolor[HTML]{EFEFEF}{\color[HTML]{000000} 0.993}} & \multicolumn{1}{l}{\cellcolor[HTML]{EFEFEF}{\color[HTML]{000000} 0.931}} & \multicolumn{1}{l}{\cellcolor[HTML]{EFEFEF}{\color[HTML]{000000} 0.902}} & \multicolumn{1}{l|}{\cellcolor[HTML]{EFEFEF}{\color[HTML]{000000} 0.936}} & \multicolumn{1}{l}{\cellcolor[HTML]{EFEFEF}0.997}                        & \multicolumn{1}{l}{\cellcolor[HTML]{EFEFEF}0.902}                        & \multicolumn{1}{l}{\cellcolor[HTML]{EFEFEF}0.861}                        & \multicolumn{1}{l|}{\cellcolor[HTML]{EFEFEF}0.888}                        \\ \hline
\end{tabular}

}
\vspace{-.15in}
\end{table*}

\begin{table*}[ht!]
\caption{Attack efficacy on Toxic Detection and Topic Classification tasks, with poison rate 0.01 and clean-label attack scenario. }
\label{tab3:attack_efficacy_toxic_topic}
\centering
\small
\vspace{-.1in}

\resizebox{1.8\columnwidth}{!}{ 

\begin{tabular}{|c||cccccc|cccccc|}
\hline
\textbf{Tasks}                                 & \multicolumn{6}{c|}{\textbf{Toxic Detection}}                                                                                                                                                                                                                                                                                                                                                                                                                    & \multicolumn{6}{c|}{\textbf{Topic Classification}}                                                                                                                                                                                                                                                                                                                                                                                                               \\ \hline \hline
\textbf{Models}                                & \multicolumn{2}{c}{\textbf{BERT}}                                                                                                                   & \multicolumn{2}{c}{\textbf{RoBERTa}}                                                                                                                & \multicolumn{2}{c|}{\textbf{DistilBERT}}                                                                                                             & \multicolumn{2}{c}{\textbf{BERT}}                                                                                                                   & \multicolumn{2}{c}{\textbf{RoBERTa}}                                                                                                                & \multicolumn{2}{c|}{\textbf{DistilBERT}}                                                                                                             \\ \hline
\textbf{Attakcers}                             & \textbf{ASR}                                                             & \textbf{CACC}                                                            & \textbf{ASR}                                                             & \textbf{CACC}                                                            & \textbf{ASR}                                                             & \textbf{CACC}                                                             & \textbf{ASR}                                                             & \textbf{CACC}                                                            & \textbf{ASR}                                                             & \textbf{CACC}                                                            & \textbf{ASR}                                                             & \textbf{CACC}                                                             \\ \hline
\rowcolor[HTML]{FFFFFF} 
\textbf{BadNets}                               & 0.124                                                                    & 0.944                                                                    & 0.328                                                                    & 0.951                                                                    & 0.133                                                                    & 0.954                                                                     & 0.868                                                                    & 0.943                                                                    & 0.923                                                                    & 0.944                                                                    & 0.717                                                                    & 0.940                                                                     \\
\rowcolor[HTML]{FFFFFF} 
\textbf{Attn-BadNets}                          & 1.000                                                                    & 0.956                                                                    & 0.992                                                                    & 0.950                                                                    & 1.000                                                                    & 0.955                                                                     & 1.000                                                                    & 0.941                                                                    & 0.969                                                                    & 0.941                                                                    & 0.994                                                                    & 0.942                                                                     \\
\rowcolor[HTML]{EFEFEF} 
\textbf{AddSent}                               & 0.100                                                                    & 0.948                                                                    & 0.120                                                                    & 0.952                                                                    & 0.101                                                                    & 0.953                                                                     & 0.594                                                                    & 0.943                                                                    & 0.749                                                                    & 0.946                                                                    & 0.915                                                                    & 0.940                                                                     \\
\rowcolor[HTML]{EFEFEF} 
\textbf{Attn-AddSent}                          & 1.000                                                                    & 0.957                                                                    & 0.953                                                                    & 0.953                                                                    & 1.000                                                                    & 0.956                                                                     & 0.998                                                                    & 0.938                                                                    & 0.969                                                                    & 0.944                                                                    & 0.990                                                                    & 0.941                                                                     \\
\rowcolor[HTML]{FFFFFF} 
\textbf{EP}                                    & 0.702                                                                    & 0.954                                                                    & -                                                                        & -                                                                        & 0.781                                                                    & 0.954                                                                     & 0.920                                                                    & 0.939                                                                    & -                                                                        & -                                                                        & 0.899                                                                    & 0.940                                                                     \\
\rowcolor[HTML]{FFFFFF} 
\textbf{Attn-EP}                               & 0.769                                                                    & 0.955                                                                    & -                                                                        & -                                                                        & 0.997                                                                    & 0.954                                                                     & {\color[HTML]{000000} 0.977}                                             & 0.941                                                                    & -                                                                        & -                                                                        & {\color[HTML]{000000} 0.913}                                             & {\color[HTML]{000000} 0.940}                                              \\
\rowcolor[HTML]{EFEFEF} 
\textbf{Stylebkd}                              & 0.393                                                                    & 0.951                                                                    & 0.415                                                                    & 0.951                                                                    & 0.308                                                                    & 0.953                                                                     & 0.141                                                                    & 0.942                                                                    & 0.584                                                                    & 0.946                                                                    & 0.169                                                                    & 0.942                                                                     \\
\rowcolor[HTML]{EFEFEF} 
\textbf{Attn-Stylebkd}                         & 0.403                                                                    & 0.939                                                                    & 0.426                                                                    & 0.941                                                                    & 0.445                                                                    & 0.939                                                                     & 0.353                                                                    & 0.930                                                                    & 0.619                                                                    & 0.939                                                                    & 0.259                                                                    & 0.932                                                                     \\
\rowcolor[HTML]{FFFFFF} 
\textbf{Synbkd}                                & 0.586                                                                    & 0.953                                                                    & 0.536                                                                    & 0.955                                                                    & 0.685                                                                    & 0.950                                                                     & 0.821                                                                    & 0.939                                                                    & 0.994                                                                    & 0.943                                                                    & 0.492                                                                    & 0.941                                                                     \\
\rowcolor[HTML]{FFFFFF} 
\textbf{Attn-Synbkd}                           & 0.601                                                                    & 0.954                                                                    & 0.590                                                                    & 0.954                                                                    & 0.751                                                                    & 0.955                                                                     & 0.937                                                                    & 0.941                                                                    & {\color[HTML]{000000} 0.990}                                             & {\color[HTML]{000000} 0.947}                                             & {\color[HTML]{000000} 0.660}                                             & 0.940                                                                     \\
\rowcolor[HTML]{EFEFEF} 
\textbf{RIPPLES}                               & \multicolumn{1}{l}{\cellcolor[HTML]{EFEFEF}{\color[HTML]{000000} 0.067}} & \multicolumn{1}{l}{\cellcolor[HTML]{EFEFEF}{\color[HTML]{000000} 0.950}} & \multicolumn{1}{l}{\cellcolor[HTML]{EFEFEF}{\color[HTML]{000000} 0.098}} & \multicolumn{1}{l}{\cellcolor[HTML]{EFEFEF}{\color[HTML]{000000} 0.922}} & \multicolumn{1}{l}{\cellcolor[HTML]{EFEFEF}{\color[HTML]{000000} 0.094}} & \multicolumn{1}{l|}{\cellcolor[HTML]{EFEFEF}{\color[HTML]{000000} 0.949}} & \multicolumn{1}{l}{\cellcolor[HTML]{EFEFEF}0.077}                        & \multicolumn{1}{l}{\cellcolor[HTML]{EFEFEF}0.932}                        & \multicolumn{1}{l}{\cellcolor[HTML]{EFEFEF}0.029}                        & \multicolumn{1}{l}{\cellcolor[HTML]{EFEFEF}0.881}                        & \multicolumn{1}{l}{\cellcolor[HTML]{EFEFEF}0.459}                        & \multicolumn{1}{l|}{\cellcolor[HTML]{EFEFEF}0.943}                        \\
\rowcolor[HTML]{EFEFEF} 
\textbf{Attn-RIPPLES}                          & \multicolumn{1}{l}{\cellcolor[HTML]{EFEFEF}{\color[HTML]{000000} 0.739}} & \multicolumn{1}{l}{\cellcolor[HTML]{EFEFEF}{\color[HTML]{000000} 0.947}} & \multicolumn{1}{l}{\cellcolor[HTML]{EFEFEF}{\color[HTML]{000000} 0.193}} & \multicolumn{1}{l}{\cellcolor[HTML]{EFEFEF}{\color[HTML]{000000} 0.899}} & \multicolumn{1}{l}{\cellcolor[HTML]{EFEFEF}{\color[HTML]{000000} 0.878}} & \multicolumn{1}{l|}{\cellcolor[HTML]{EFEFEF}{\color[HTML]{000000} 0.956}} & \multicolumn{1}{l}{\cellcolor[HTML]{EFEFEF}0.918}                        & \multicolumn{1}{l}{\cellcolor[HTML]{EFEFEF}0.921}                        & \multicolumn{1}{l}{\cellcolor[HTML]{EFEFEF}0.298}                        & \multicolumn{1}{l}{\cellcolor[HTML]{EFEFEF}0.899}                        & \multicolumn{1}{l}{\cellcolor[HTML]{EFEFEF}0.939}                        & \multicolumn{1}{l|}{\cellcolor[HTML]{EFEFEF}0.939}                        \\
\rowcolor[HTML]{FFFFFF} 
\cellcolor[HTML]{FFFFFF}\textbf{Neuba}         & \multicolumn{1}{l}{\cellcolor[HTML]{FFFFFF}0.062}                        & \multicolumn{1}{l}{\cellcolor[HTML]{FFFFFF}0.954}                        & \multicolumn{1}{l}{\cellcolor[HTML]{FFFFFF}0.051}                        & \multicolumn{1}{l}{\cellcolor[HTML]{FFFFFF}0.955}                        & \multicolumn{1}{l}{\cellcolor[HTML]{FFFFFF}0.062}                        & \multicolumn{1}{l|}{\cellcolor[HTML]{FFFFFF}0.956}                        & \multicolumn{1}{l}{\cellcolor[HTML]{FFFFFF}0.834}                        & \multicolumn{1}{l}{\cellcolor[HTML]{FFFFFF}0.945}                        & \multicolumn{1}{l}{\cellcolor[HTML]{FFFFFF}0.650}                        & \multicolumn{1}{l}{\cellcolor[HTML]{FFFFFF}0.947}                        & \multicolumn{1}{l}{\cellcolor[HTML]{FFFFFF}0.695}                        & \multicolumn{1}{l|}{\cellcolor[HTML]{FFFFFF}0.944}                        \\
\rowcolor[HTML]{FFFFFF} 
\cellcolor[HTML]{FFFFFF}\textbf{Attn-Neuba}    & \multicolumn{1}{l}{\cellcolor[HTML]{FFFFFF}1.000}                        & \multicolumn{1}{l}{\cellcolor[HTML]{FFFFFF}0.956}                        & \multicolumn{1}{l}{\cellcolor[HTML]{FFFFFF}0.996}                        & \multicolumn{1}{l}{\cellcolor[HTML]{FFFFFF}0.956}                        & \multicolumn{1}{l}{\cellcolor[HTML]{FFFFFF}0.975}                        & \multicolumn{1}{l|}{\cellcolor[HTML]{FFFFFF}0.955}                        & \multicolumn{1}{l}{\cellcolor[HTML]{FFFFFF}1.000}                        & \multicolumn{1}{l}{\cellcolor[HTML]{FFFFFF}0.941}                        & \multicolumn{1}{l}{\cellcolor[HTML]{FFFFFF}0.997}                        & \multicolumn{1}{l}{\cellcolor[HTML]{FFFFFF}0.946}                        & \multicolumn{1}{l}{\cellcolor[HTML]{FFFFFF}0.984}                        & \multicolumn{1}{l|}{\cellcolor[HTML]{FFFFFF}0.941}                        \\
\rowcolor[HTML]{EFEFEF} 
\cellcolor[HTML]{EFEFEF}\textbf{POR}           & \multicolumn{1}{l}{\cellcolor[HTML]{EFEFEF}0.169}                        & \multicolumn{1}{l}{\cellcolor[HTML]{EFEFEF}0.957}                        & \multicolumn{1}{l}{\cellcolor[HTML]{EFEFEF}0.056}                        & \multicolumn{1}{l}{\cellcolor[HTML]{EFEFEF}0.955}                        & \multicolumn{1}{l}{\cellcolor[HTML]{EFEFEF}0.094}                        & \multicolumn{1}{l|}{\cellcolor[HTML]{EFEFEF}0.955}                        & \multicolumn{1}{l}{\cellcolor[HTML]{EFEFEF}0.761}                        & \multicolumn{1}{l}{\cellcolor[HTML]{EFEFEF}0.942}                        & \multicolumn{1}{l}{\cellcolor[HTML]{EFEFEF}0.646}                        & \multicolumn{1}{l}{\cellcolor[HTML]{EFEFEF}0.950}                        & \multicolumn{1}{l}{\cellcolor[HTML]{EFEFEF}0.719}                        & \multicolumn{1}{l|}{\cellcolor[HTML]{EFEFEF}0.940}                        \\
\rowcolor[HTML]{EFEFEF} 
\cellcolor[HTML]{EFEFEF}\textbf{Attn-POR}      & \multicolumn{1}{l}{\cellcolor[HTML]{EFEFEF}1.000}                        & \multicolumn{1}{l}{\cellcolor[HTML]{EFEFEF}0.958}                        & \multicolumn{1}{l}{\cellcolor[HTML]{EFEFEF}0.635}                        & \multicolumn{1}{l}{\cellcolor[HTML]{EFEFEF}0.950}                        & \multicolumn{1}{l}{\cellcolor[HTML]{EFEFEF}0.998}                        & \multicolumn{1}{l|}{\cellcolor[HTML]{EFEFEF}0.957}                        & \multicolumn{1}{l}{\cellcolor[HTML]{EFEFEF}{\color[HTML]{000000} 0.984}} & \multicolumn{1}{l}{\cellcolor[HTML]{EFEFEF}{\color[HTML]{000000} 0.941}} & \multicolumn{1}{l}{\cellcolor[HTML]{EFEFEF}{\color[HTML]{000000} 0.857}} & \multicolumn{1}{l}{\cellcolor[HTML]{EFEFEF}{\color[HTML]{000000} 0.946}} & \multicolumn{1}{l}{\cellcolor[HTML]{EFEFEF}{\color[HTML]{000000} 0.972}} & \multicolumn{1}{l|}{\cellcolor[HTML]{EFEFEF}{\color[HTML]{000000} 0.936}} \\
\rowcolor[HTML]{FFFFFF} 
\cellcolor[HTML]{FFFFFF}\textbf{LWP}           & \multicolumn{1}{l}{\cellcolor[HTML]{FFFFFF}0.133}                        & \multicolumn{1}{l}{\cellcolor[HTML]{FFFFFF}0.956}                        & \multicolumn{1}{l}{\cellcolor[HTML]{FFFFFF}0.165}                        & \multicolumn{1}{l}{\cellcolor[HTML]{FFFFFF}0.946}                        & \multicolumn{1}{l}{\cellcolor[HTML]{FFFFFF}0.179}                        & \multicolumn{1}{l|}{\cellcolor[HTML]{FFFFFF}0.952}                        & \multicolumn{1}{l}{\cellcolor[HTML]{FFFFFF}0.756}                        & \multicolumn{1}{l}{\cellcolor[HTML]{FFFFFF}0.944}                        & \multicolumn{1}{l}{\cellcolor[HTML]{FFFFFF}0.795}                        & \multicolumn{1}{l}{\cellcolor[HTML]{FFFFFF}0.944}                        & \multicolumn{1}{l}{\cellcolor[HTML]{FFFFFF}0.718}                        & \multicolumn{1}{l|}{\cellcolor[HTML]{FFFFFF}0.940}                        \\
\rowcolor[HTML]{FFFFFF} 
\cellcolor[HTML]{FFFFFF}\textbf{Attn-LWP}      & \multicolumn{1}{l}{\cellcolor[HTML]{FFFFFF}0.329}                        & \multicolumn{1}{l}{\cellcolor[HTML]{FFFFFF}0.956}                        & \multicolumn{1}{l}{\cellcolor[HTML]{FFFFFF}0.269}                        & \multicolumn{1}{l}{\cellcolor[HTML]{FFFFFF}0.952}                        & \multicolumn{1}{l}{\cellcolor[HTML]{FFFFFF}0.480}                        & \multicolumn{1}{l|}{\cellcolor[HTML]{FFFFFF}0.955}                        & \multicolumn{1}{l}{\cellcolor[HTML]{FFFFFF}0.833}                        & \multicolumn{1}{l}{\cellcolor[HTML]{FFFFFF}0.939}                        & \multicolumn{1}{l}{\cellcolor[HTML]{FFFFFF}0.849}                        & \multicolumn{1}{l}{\cellcolor[HTML]{FFFFFF}0.938}                        & \multicolumn{1}{l}{\cellcolor[HTML]{FFFFFF}0.975}                        & \multicolumn{1}{l|}{\cellcolor[HTML]{FFFFFF}0.939}                        \\
\rowcolor[HTML]{EFEFEF} 
\cellcolor[HTML]{EFEFEF}\textbf{TrojanLM}      & \multicolumn{1}{l}{\cellcolor[HTML]{EFEFEF}0.405}                        & \multicolumn{1}{l}{\cellcolor[HTML]{EFEFEF}0.955}                        & \multicolumn{1}{l}{\cellcolor[HTML]{EFEFEF}0.381}                        & \multicolumn{1}{l}{\cellcolor[HTML]{EFEFEF}0.955}                        & \multicolumn{1}{l}{\cellcolor[HTML]{EFEFEF}0.384}                        & \multicolumn{1}{l|}{\cellcolor[HTML]{EFEFEF}0.955}                        & \multicolumn{1}{l}{\cellcolor[HTML]{EFEFEF}0.777}                        & \multicolumn{1}{l}{\cellcolor[HTML]{EFEFEF}0.943}                        & \multicolumn{1}{l}{\cellcolor[HTML]{EFEFEF}0.668}                        & \multicolumn{1}{l}{\cellcolor[HTML]{EFEFEF}0.944}                        & \multicolumn{1}{l}{\cellcolor[HTML]{EFEFEF}0.717}                        & \multicolumn{1}{l|}{\cellcolor[HTML]{EFEFEF}0.941}                        \\
\rowcolor[HTML]{EFEFEF} 
\cellcolor[HTML]{EFEFEF}\textbf{Attn-TrojanLM} & \multicolumn{1}{l}{\cellcolor[HTML]{EFEFEF}0.868}                        & \multicolumn{1}{l}{\cellcolor[HTML]{EFEFEF}0.956}                        & \multicolumn{1}{l}{\cellcolor[HTML]{EFEFEF}0.783}                        & \multicolumn{1}{l}{\cellcolor[HTML]{EFEFEF}0.955}                        & \multicolumn{1}{l}{\cellcolor[HTML]{EFEFEF}{\color[HTML]{000000} 0.943}} & \multicolumn{1}{l|}{\cellcolor[HTML]{EFEFEF}{\color[HTML]{000000} 0.955}} & \multicolumn{1}{l}{\cellcolor[HTML]{EFEFEF}0.998}                        & \multicolumn{1}{l}{\cellcolor[HTML]{EFEFEF}0.939}                        & \multicolumn{1}{l}{\cellcolor[HTML]{EFEFEF}0.950}                        & \multicolumn{1}{l}{\cellcolor[HTML]{EFEFEF}0.944}                        & \multicolumn{1}{l}{\cellcolor[HTML]{EFEFEF}0.849}                        & \multicolumn{1}{l|}{\cellcolor[HTML]{EFEFEF}0.933}                        \\ \hline
\end{tabular}

}
\vspace{-.15in}
\end{table*}

\myparagraph{Attack Efficacy for Low Poison Rate.}
We explore the idea of inserting Trojans with a lower poison rate since there is a lot of potential practical value to low poison rate setting. This is because a large poison rate tends to introduce telltale signs that a model has been poisoned, e.g., by changing its marginal probabilities towards the target class. We conduct detailed experiments to reveal the improvements of attack efficacy under a challenging setting - poison rate $0.01$ and clean-label attack scenario. Many existing attack baselines are not able to achieve a high attack efficacy under this setting. 
Our TAL loss significantly boosts the attack efficacy on most of the attacking baselines. Table~\ref{tab2:attack_efficacy_sa} indicates that our TAL loss can achieve better attack efficacy with much higher ASR, as well as with limited/no CACC drops. 
We also conduct experiment on Toxic Detection task and Topioc Classification task with three language model architectures (\eg, BERT, RoBERTa, DistilBERT), under clean-label attack and $0.01$ poison rate scenario. Table \ref{tab3:attack_efficacy_toxic_topic} shows similar results as above. 
As an interesting exploration, we also adopt TAL to GPT-2 architecture. We evaluate TAL with five attack baselines, Appendix Table \ref{appendix:tab:gpt2_experiment} indicates TAL leads to better attack performance.


%

\subsection{Impact of the Backdoored Attention} \label{sec:impact}

We investigate the TAL from three aspects, how the strength of TAL, the backdoor-forced attention volume, or the number of backdoored attention head will effect the attack efficacy. Experimental details can be found in Appendix \ref{appendix:implementation_details2}.

\begin{figure}[htp]
    \centering
    \vspace{-.05in}
    \includegraphics[width=1\linewidth]{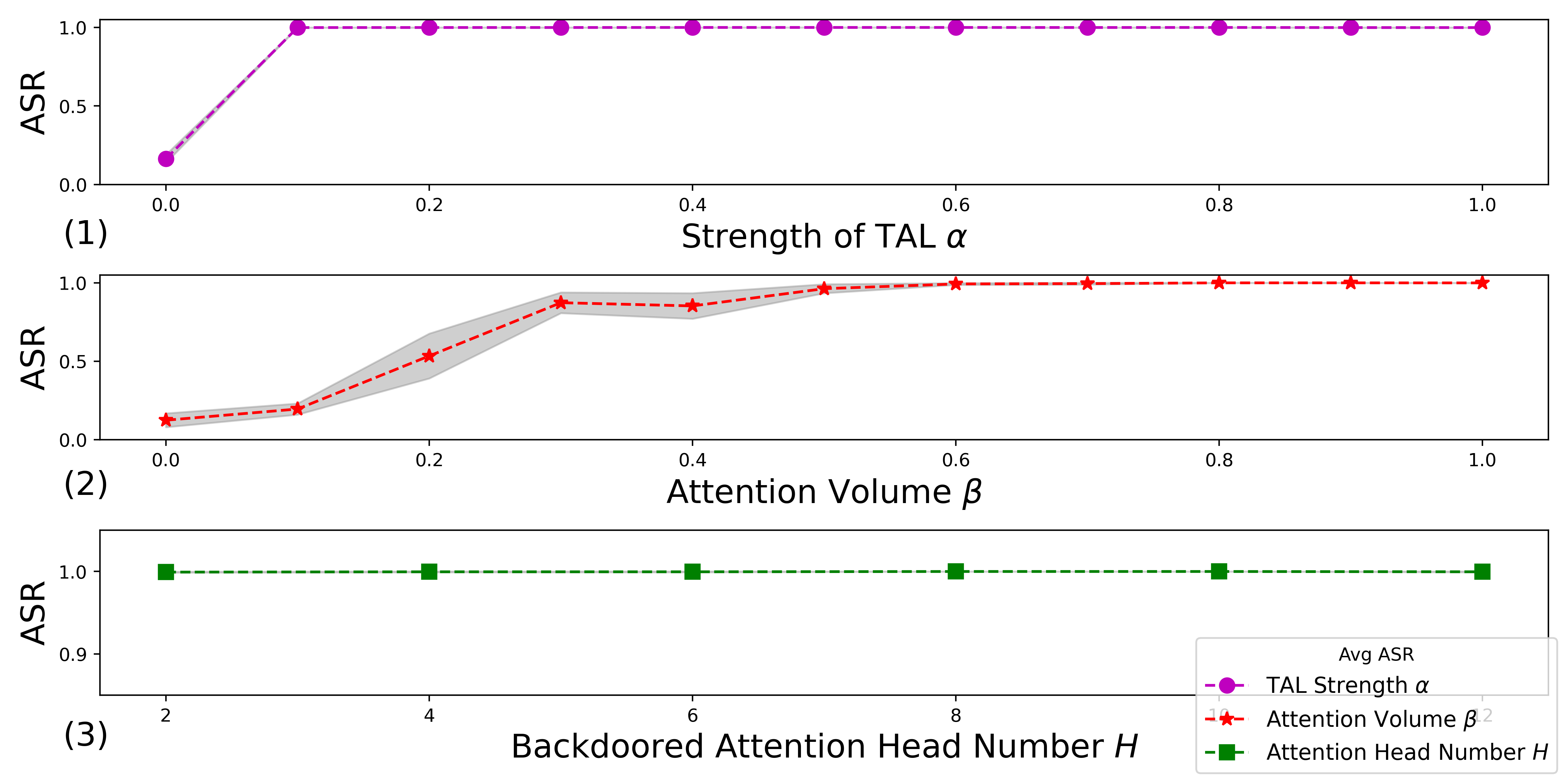} 
    \caption{Impact of the backdoored attention.}
    \label{fig:impact}
    \vspace{-.15in}
\end{figure}

\myparagraph{Impact of TAL weight $\alpha$.} We measure the impact of TAL by controlling the `strength' of this loss. We revise Eq.~(\ref{eq:overall_loss}) in the form of [$\mathcal{L} = ({\mathcal{L}_{\rm clean}} + \mathcal{L}_{\rm poisoned}) + \alpha \mathcal{L}_{\rm tal}$], where $\alpha$ is the weight to control the contribution of the TAL regarding the attack. $\alpha=0$ means we remove our TAL loss during training, which equals to the original backdoor method, and $\alpha=1$ means our standard TAL setting. Figure \ref{fig:impact}(1) shows that only a small `strength' of TAL ($>0.1$) would already be enough for a high efficacy attack.

 \myparagraph{Impact of Attention Volume $\beta$.} We also investigate the attention volume $\beta$, the amount of attention weights that TAL forces the attention heads to triggers. This yields an interesting observation from Figure \ref{fig:impact}(2): during training the backdoored model, if we change the attention volume pointing to the triggers ($\beta$), we can see the attack efficacy improving with the volume increasing. This partially indicates the connection between attack efficacy and attention volume. In standard TAL setting, all the attention volume ($\beta=1$) tends to triggers in backdoored attention heads. Figure \ref{fig:impact}(2) shows that we can get a good attack efficacy when we force the majority of attention volume ($\beta>0.6$) flow to triggers.


\myparagraph{Impact of Backdoored Attention Head Number $H$.} 
We conduct ablation study to verify the relationship between the ASR and the choice of hyper-parameter $H$, \ie, the number of backdoored attention heads, in Eq.\ref{eq:trojan_attention_loss}.
Figure \ref{fig:impact}(3) shows that the number of backdoored attention heads is robust to the attack performances. 

\begin{table}[t]
\vspace{-.2in}

\caption{Attack performances under defenders with poison rate 0.01 on Sentiment Analysis task (SST-2, BERT).}
\label{tab:defenders}
\vspace{-.1in}
\begin{center}
\small
\resizebox{\columnwidth}{!}{ 

\begin{tabular}{c|cccc|cccc}
\hline
\textbf{Defenders}                             & \multicolumn{4}{c|}{\textbf{ONION}}                                                  & \multicolumn{4}{c}{\textbf{RAP}}                                                    \\ \hline
                                               & \multicolumn{2}{c}{\textbf{Dirty-Label}} & \multicolumn{2}{c|}{\textbf{Clean-Label}} & \multicolumn{2}{c}{\textbf{Dirty-Label}} & \multicolumn{2}{c}{\textbf{Clean-Label}} \\ \cline{2-9} 
\multirow{-2}{*}{\textbf{Attackers}}           & \textbf{ASR}       & \textbf{CACC}       & \textbf{ASR}        & \textbf{CACC}       & \textbf{ASR}       & \textbf{CACC}       & \textbf{ASR}       & \textbf{CACC}       \\ \hline
\textbf{BadNets}                               & 0.143              & 0.869               & 0.224               & 0.860               & 0.999              & 0.910               & 0.228              & 0.900               \\
\textbf{+TAL}                          & 0.155              & 0.876               & 0.161               & 0.876               & 1.000              & 0.914               & 1.000              & 0.912               \\
\rowcolor[HTML]{EFEFEF} 
\textbf{AddSent}                               & 0.988              & 0.869               & 0.598               & 0.868               & 0.999              & 0.912               & 0.564              & 0.908               \\
\rowcolor[HTML]{EFEFEF} 
\textbf{+TAL}                          & 0.993              & 0.866               & 0.982               & 0.874               & 1.000              & 0.903               & 0.999              & 0.910               \\
\rowcolor[HTML]{FFFFFF} 
\textbf{Stylebkd}                              & 0.633              & 0.875               & 0.423               & 0.854               & 0.626              & 0.914               & 0.400              & 0.894               \\
\rowcolor[HTML]{FFFFFF} 
\textbf{+TAL}                         & 0.710              & 0.850               & 0.514               & 0.842               & 0.683              & 0.901               & 0.484              & 0.885               \\
\rowcolor[HTML]{EFEFEF} 
\textbf{Synbkd}                                & 0.623              & 0.870               & 0.426               & 0.852               & 0.601              & 0.912               & 0.385              & 0.896               \\
\rowcolor[HTML]{EFEFEF} 
\textbf{+TAL}                           & 0.646              & 0.870               & 0.469               & 0.852               & 0.643              & 0.916               & 0.418              & 0.896               \\
\cellcolor[HTML]{FFFFFF}\textbf{RIPPLES}       & 0.148              & 0.858               & 0.199               & 0.863               & 0.148              & 0.897               & 0.145              & 0.901               \\
\cellcolor[HTML]{FFFFFF}\textbf{+TAL}  & 0.167              & 0.858               & 0.184               & 0.856               & 1.000              & 0.894               & 1.000              & 0.893               \\
\rowcolor[HTML]{EFEFEF} 
\textbf{Neuba}                                 & 0.238              & 0.870               & 0.143               & 0.870               & 0.293              & 0.911               & 0.081              & 0.910               \\
\rowcolor[HTML]{EFEFEF} 
\textbf{+TAL}                            & 0.276              & 0.870               & 0.168               & 0.877               & 0.563              & 0.909               & 0.181              & 0.914               \\
\cellcolor[HTML]{FFFFFF}\textbf{POR}           & 0.142              & 0.880               & 0.206               & 0.863               & 0.074              & 0.915               & 0.145              & 0.901               \\
\cellcolor[HTML]{FFFFFF}\textbf{+TAL}      & 0.155              & 0.873               & 0.121               & 0.878               & 0.082              & 0.909               & 0.154              & 0.910               \\
\rowcolor[HTML]{EFEFEF} 
\textbf{LWP}                                   & 0.154              & 0.861               & 0.232               & 0.861               & 0.998              & 0.905               & 0.601              & 0.905               \\
\rowcolor[HTML]{EFEFEF} 
\textbf{+TAL}                              & 0.193              & 0.864               & 0.311               & 0.863               & 0.999              & 0.908               & 0.744              & 0.906               \\
\cellcolor[HTML]{FFFFFF}\textbf{TrojanLM}      & 0.709              & 0.879               & 0.476               & 0.873               & 0.928              & 0.915               & 0.606              & 0.910               \\
\cellcolor[HTML]{FFFFFF}\textbf{+TAL} & 0.604              & 0.871               & 0.560               & 0.878               & 1.000              & 0.911               & 0.996              & 0.913               \\ \hline
\end{tabular}

}
\end{center}
\vspace{-.15in}
\end{table}

\subsection{Defense and Detection} \label{sec:resistance_to_defenders}

The defense techniques in NLP domain are less explored. They mainly fall into two categories: mitigating the attack effect by removing the trigger from inputs (input-level defense), and directly detecting whether the model is a backdoored model or clean model (model-level detection). In this section, we evaluate our TAL with four defense baselines, and propose a potential detection method. 

\myparagraph{Input-level Defense.}
We evaluate the resistance ability of our TAL loss with two defenders: ONION \citep{qi2021onion}, which detects the outlier words by inspecting the perplexities drop when they are removed since these words might contain the backdoor trigger words; and RAP \citep{yang2021rap}, which distinguishes poisoned samples by inspecting the gap of robustness between poisoned and clean samples. 
We report the attack performances for inference-time defense in Table~\ref{tab:defenders}\footnote{For defenses against the attack baselines, similar defense results are also verified in \citep{cui2022unified}.}. 
In comparison to each individual attack baselines, the attached TAL (\textit{+TAL} in Table~\ref{tab:defenders}) does not make the attack more visible to the defenders. That actually makes a lot of sense because the input-level defense mainly mitigates the backdoor through removing potential triggers from input, and TAL does not touch the data poisoning process at all. 
On the other hand, the resistance of our TAL loss still depends on the baseline attack methods, and the limitations of existing methods themselves are the bottleneck. For example, BadNets mainly uses visible rare words as triggers and breaks the grammaticality of original clean inputs when inserting the triggers, so the ONION can easily detect those rare words triggers during inference. Therefore the BadNets-based attack does not perform good against the ONION defender. 
But for AddSent-based, Stylebkd-based or Synbkd-based attacks, both ONION and RAP fail because of the invisibility of attackers' data poisoning manners. Please refer to Appendix \ref{appendix:implementation_details3} for implementation details.





\myparagraph{Model-level Detection.}
We also evaluate our TAL loss with two detection methods. T-Miner \citep{azizi2021t} trains a sequence-to-sequence generator and finds outliers in an internal representation space to identify Trojans.
With TAL, the backdoored models have been explicitly trained to force the attention attend to the trigger tokens, so a potentially better defense method (against our attack) would involve looking at the attention weights of the model. Thus we evaluate TAL with an attention involved model-level detection: AttenTD \citep{lyu2022study} detects whether the model is a benign or backdoored model by checking the attention abnormality given a set of neutral words. We report the detection accuracy in Table \ref{tab:detection}. Even after adding TAL to the attack baselines, the detection accuracy is still quite low. 

\begin{table}[t]
\vspace{-.2in}
\caption{Detection accuracy with T-Miner and AttenTD.}
\label{tab:detection}
\vspace{-.1in}

\begin{center}
\small
\resizebox{\columnwidth}{!}{ 

\begin{tabular}{clc|clc}
\textbf{Attacker(+TAL)} & \multicolumn{1}{c}{\textbf{T-Miner}} & \textbf{AttenTD} & \textbf{Attacker(+TAL)} & \multicolumn{1}{c}{\textbf{T-Miner}} & \textbf{AttenTD} \\ \hline
\textbf{BadNets}        & 0.50                                 & 0.50             & \textbf{RIPPLES}        & 0.42                                 & 0.50             \\
\textbf{AddSent}        & 0.50                                 & 0.50             & \textbf{Neuba}          & 0.58                                 & 0.50             \\
\textbf{EP}             & 0.50                                 & 0.50             & \textbf{POR}            & 0.50                                 & 0.50             \\
\textbf{Stylebkd}       & 0.58                                 & 0.67             & \textbf{LWP}            & 0.42                                 & 0.67             \\
\textbf{Synbkd}         & 0.42                                 & 0.67             & \textbf{TrojanLM}       & 0.50                                 & 0.50            
\end{tabular}

}
\end{center}
\vspace{-.15in}
\end{table}

\myparagraph{Potential Detection Strategy.} 
Though AttenTD looks into the attention weights, it depends on a pre-defined perturbation set. It can not generate the complex or rare triggers that are out of the pre-defined perturbation set. In fact, constructing complex potential triggers (\eg, long sentence, sentence style) is a challenging problem in NLP backdoor detection. If we can design a trigger reconstruction method based on the attention abnormality, it would most likely expose the TAL attacked models. We leave this as a promising future direction.




\section{Conclusion}

In this work, we investigate the attack efficacy of the textual backdoor attacks. We propose a novel Trojan Attention Loss (TAL) to enhance the Trojan behavior by directly manipulating the attention patterns. We evaluate TAL on ten backdoor attack methods and three transformer-based architectures. Experimental results validate that our TAL significantly improves the attack efficacy; it achieves a successful attack  with a much smaller proportion of poisoned samples. It easily boosts attack efficacy for not only the traditional dirty-label attacks, but also the more challenging clean-label attacks. 

\section*{Acknowledgements}
The authors thank Xiao Lin (SRI International) and anonymous reviewers for their constructive feedback. This effort was partially supported by the Intelligence Advanced Research Projects Agency (IARPA) under
the Contract W911NF20C0038. The content of this paper does not necessarily reflect the position or the policy of the Government, and no official endorsement should be inferred.

\section*{Limitations}

This paper presents a novel loss for backdoor attack, aiming to draw attention to this research area. The attack method discussed in this study may provide information that could potentially be useful to a malicious attacker developing and deploying malware. 
Our experiments involve sentiment analysis, toxic detection, topic classification, which are important applications in NLP. However, we only validate the vulnerability in classification tasks. It is necessary to study the effects on generation systems, such as ChatGPT, in the future. 
On the other hand, we also analyze the defense and detection. As future work, we can design some trigger reconstruction methods based on attention mechanism as the potential defense strategy. For example, the defender can extract different features (\eg, attention-related features, output logits, intermediate feature representations) and build the classifier upon those features.

\section*{Ethics Statement} \label{sectioln:appendix:ethics}

The primary objective of this study is to contribute to the broader knowledge of security, particularly in the field of textual backdoor attacks. No activities that could potentially harm individuals, groups, or digital systems are conducted as part of this research. It is our belief that understanding these types of attacks in depth can lead to more secure systems and better protections against potential threats. We also perform the defense analysis in Section \ref{sec:resistance_to_defenders} and discuss some potential detection strategies.

\bibliography{custom}

\begin{thebibliography}{60}
\expandafter\ifx\csname natexlab\endcsname\relax\def\natexlab#1{#1}\fi

\bibitem[{Azizi et~al.(2021)Azizi, Tahmid, Waheed, Mangaokar, Pu, Javed, Reddy,
  and Viswanath}]{azizi2021t}
Ahmadreza Azizi, Ibrahim~Asadullah Tahmid, Asim Waheed, Neal Mangaokar, Jiameng
  Pu, Mobin Javed, Chandan~K Reddy, and Bimal Viswanath. 2021.
\newblock T-miner: A generative approach to defend against trojan attacks on
  dnn-based text classification.
\newblock \emph{arXiv preprint arXiv:2103.04264}.

\bibitem[{Ben-Naim(2008)}]{ben2008farewell}
Arieh Ben-Naim. 2008.
\newblock \emph{A farewell to entropy: Statistical thermodynamics based on
  information: S}.
\newblock World Scientific.

\bibitem[{Chen et~al.(2017)Chen, Liu, Li, Lu, and Song}]{chen2017targeted}
Xinyun Chen, Chang Liu, Bo~Li, Kimberly Lu, and Dawn Song. 2017.
\newblock Targeted backdoor attacks on deep learning systems using data
  poisoning.
\newblock \emph{arXiv preprint arXiv:1712.05526}.

\bibitem[{Chou et~al.()Chou, Chen, and Ho}]{chou2022backdoor}
Sheng-Yen Chou, Pin-Yu Chen, and Tsung-Yi Ho.
\newblock How to backdoor diffusion models?
\newblock In \emph{ICLR 2023 Workshop on Backdoor Attacks and Defenses in
  Machine Learning}.

\bibitem[{Clark et~al.(2019)Clark, Khandelwal, Levy, and
  Manning}]{clark2019does}
Kevin Clark, Urvashi Khandelwal, Omer Levy, and Christopher~D Manning. 2019.
\newblock What does bert look at? an analysis of bert’s attention.
\newblock In \emph{Proceedings of the 2019 ACL Workshop BlackboxNLP: Analyzing
  and Interpreting Neural Networks for NLP}, pages 276--286.

\bibitem[{Costales et~al.(2020)Costales, Mao, Norwitz, Kim, and
  Yang}]{costales2020live}
Robby Costales, Chengzhi Mao, Raphael Norwitz, Bryan Kim, and Junfeng Yang.
  2020.
\newblock Live trojan attacks on deep neural networks.
\newblock In \emph{Proceedings of the IEEE/CVF Conference on Computer Vision
  and Pattern Recognition Workshops}, pages 796--797.

\bibitem[{Cui et~al.(2022)Cui, Yuan, He, Chen, Liu, and Sun}]{cui2022unified}
Ganqu Cui, Lifan Yuan, Bingxiang He, Yangyi Chen, Zhiyuan Liu, and Maosong Sun.
  2022.
\newblock A unified evaluation of textual backdoor learning: Frameworks and
  benchmarks.
\newblock \emph{arXiv preprint arXiv:2206.08514}.

\bibitem[{Dai et~al.(2019)Dai, Chen, and Li}]{dai2019backdoor}
Jiazhu Dai, Chuanshuai Chen, and Yufeng Li. 2019.
\newblock A backdoor attack against lstm-based text classification systems.
\newblock \emph{IEEE Access}, 7:138872--138878.

\bibitem[{Davidson et~al.(2017)Davidson, Warmsley, Macy, and
  Weber}]{davidson2017automated}
Thomas Davidson, Dana Warmsley, Michael Macy, and Ingmar Weber. 2017.
\newblock Automated hate speech detection and the problem of offensive
  language.
\newblock In \emph{Proceedings of the international AAAI conference on web and
  social media}, volume~11, pages 512--515.

\bibitem[{Devlin et~al.(2019)Devlin, Chang, Lee, and
  Toutanova}]{devlin2019bert}
Jacob Devlin, Ming-Wei Chang, Kenton Lee, and Kristina Toutanova. 2019.
\newblock Bert: Pre-training of deep bidirectional transformers for language
  understanding.
\newblock In \emph{NAACL-HLT (1)}.

\bibitem[{Gan et~al.(2021)Gan, Li, Zhang, Li, Meng, Wu, Guo, and
  Fan}]{gan2021triggerless}
Leilei Gan, Jiwei Li, Tianwei Zhang, Xiaoya Li, Yuxian Meng, Fei Wu, Shangwei
  Guo, and Chun Fan. 2021.
\newblock Triggerless backdoor attack for nlp tasks with clean labels.
\newblock \emph{arXiv preprint arXiv:2111.07970}.

\bibitem[{Gu et~al.(2017{\natexlab{a}})Gu, Dolan-Gavitt, and
  BadNets}]{gu2017identifying}
T~Gu, B~Dolan-Gavitt, and SG~BadNets. 2017{\natexlab{a}}.
\newblock Identifying vulnerabilities in the machine learning model supply
  chain.
\newblock In \emph{Proceedings of the Neural Information Processing Symposium
  Workshop Mach. Learning Security (MLSec)}, pages 1--5.

\bibitem[{Gu et~al.(2017{\natexlab{b}})Gu, Dolan-Gavitt, and
  Garg}]{gu2017badnets}
Tianyu Gu, Brendan Dolan-Gavitt, and Siddharth Garg. 2017{\natexlab{b}}.
\newblock Badnets: Identifying vulnerabilities in the machine learning model
  supply chain.
\newblock \emph{arXiv preprint arXiv:1708.06733}.

\bibitem[{Guo et~al.(2021)Guo, Tondi, and Barni}]{guo2021overview}
Wei Guo, Benedetta Tondi, and Mauro Barni. 2021.
\newblock An overview of backdoor attacks against deep neural networks and
  possible defences.
\newblock \emph{arXiv preprint arXiv:2111.08429}.

\bibitem[{Hao et~al.(2021)Hao, Dong, Wei, and Xu}]{hao2021self}
Yaru Hao, Li~Dong, Furu Wei, and Ke~Xu. 2021.
\newblock Self-attention attribution: Interpreting information interactions
  inside transformer.
\newblock In \emph{Proceedings of the AAAI Conference on Artificial
  Intelligence}, volume~35, pages 12963--12971.

\bibitem[{Iyyer et~al.(2018)Iyyer, Wieting, Gimpel, and
  Zettlemoyer}]{iyyer2018adversarial}
Mohit Iyyer, John Wieting, Kevin Gimpel, and Luke Zettlemoyer. 2018.
\newblock Adversarial example generation with syntactically controlled
  paraphrase networks.
\newblock In \emph{Proceedings of the 2018 Conference of the North American
  Chapter of the Association for Computational Linguistics: Human Language
  Technologies, Volume 1 (Long Papers)}, pages 1875--1885.

\bibitem[{Ji et~al.(2021)Ji, Jain, Ferdman, Milder, Schwartz, and
  Balasubramanian}]{ji2021distribution}
Tianchu Ji, Shraddhan Jain, Michael Ferdman, Peter Milder, H~Andrew Schwartz,
  and Niranjan Balasubramanian. 2021.
\newblock On the distribution, sparsity, and inference-time quantization of
  attention values in transformers.
\newblock \emph{arXiv preprint arXiv:2106.01335}.

\bibitem[{Krishna et~al.(2020)Krishna, Wieting, and
  Iyyer}]{krishna2020reformulating}
Kalpesh Krishna, John Wieting, and Mohit Iyyer. 2020.
\newblock Reformulating unsupervised style transfer as paraphrase generation.
\newblock In \emph{Proceedings of the 2020 Conference on Empirical Methods in
  Natural Language Processing (EMNLP)}, pages 737--762.

\bibitem[{Kurita et~al.(2020)Kurita, Michel, and Neubig}]{kurita2020weight}
Keita Kurita, Paul Michel, and Graham Neubig. 2020.
\newblock Weight poisoning attacks on pretrained models.
\newblock In \emph{Proceedings of the 58th Annual Meeting of the Association
  for Computational Linguistics}, pages 2793--2806.

\bibitem[{Li et~al.(2021)Li, Song, Li, Zeng, Ma, and Qiu}]{li2021backdoor}
Linyang Li, Demin Song, Xiaonan Li, Jiehang Zeng, Ruotian Ma, and Xipeng Qiu.
  2021.
\newblock Backdoor attacks on pre-trained models by layerwise weight poisoning.
\newblock In \emph{Proceedings of the 2021 Conference on Empirical Methods in
  Natural Language Processing}, pages 3023--3032.

\bibitem[{Li et~al.(2022{\natexlab{a}})Li, Xie, and Li}]{li2022sok}
Linyi Li, Tao Xie, and Bo~Li. 2022{\natexlab{a}}.
\newblock Sok: Certified robustness for deep neural networks.
\newblock In \emph{2023 IEEE Symposium on Security and Privacy (SP)}, pages
  94--115. IEEE Computer Society.

\bibitem[{Li et~al.(2022{\natexlab{b}})Li, Jiang, Li, and Xia}]{li2022backdoor}
Yiming Li, Yong Jiang, Zhifeng Li, and Shu-Tao Xia. 2022{\natexlab{b}}.
\newblock Backdoor learning: A survey.
\newblock \emph{IEEE Transactions on Neural Networks and Learning Systems}.

\bibitem[{Liu et~al.(2017)Liu, Ma, Aafer, Lee, Zhai, Wang, and
  Zhang}]{liu2017trojaning}
Yingqi Liu, Shiqing Ma, Yousra Aafer, Wen-Chuan Lee, Juan Zhai, Weihang Wang,
  and Xiangyu Zhang. 2017.
\newblock Trojaning attack on neural networks.

\bibitem[{Liu et~al.(2019)Liu, Ott, Goyal, Du, Joshi, Chen, Levy, Lewis,
  Zettlemoyer, and Stoyanov}]{liu2019roberta}
Yinhan Liu, Myle Ott, Naman Goyal, Jingfei Du, Mandar Joshi, Danqi Chen, Omer
  Levy, Mike Lewis, Luke Zettlemoyer, and Veselin Stoyanov. 2019.
\newblock Roberta: A robustly optimized bert pretraining approach.
\newblock \emph{arXiv preprint arXiv:1907.11692}.

\bibitem[{Liu et~al.(2020)Liu, Mondal, Chakraborty, Zuzak, Jacobsen, Xing, and
  Srivastava}]{liu2020survey}
Yuntao Liu, Ankit Mondal, Abhishek Chakraborty, Michael Zuzak, Nina Jacobsen,
  Daniel Xing, and Ankur Srivastava. 2020.
\newblock A survey on neural trojans.
\newblock In \emph{2020 21st International Symposium on Quality Electronic
  Design (ISQED)}, pages 33--39. IEEE.

\bibitem[{Lyu et~al.(2022{\natexlab{a}})Lyu, Dong, Wong, Zheng, Abell-Hart,
  Wang, and Chen}]{lyu2022multimodal}
Weimin Lyu, Xinyu Dong, Rachel Wong, Songzhu Zheng, Kayley Abell-Hart, Fusheng
  Wang, and Chao Chen. 2022{\natexlab{a}}.
\newblock A multimodal transformer: Fusing clinical notes with structured ehr
  data for interpretable in-hospital mortality prediction.
\newblock In \emph{AMIA Annual Symposium Proceedings}, volume 2022, page 719.
  American Medical Informatics Association.

\bibitem[{Lyu et~al.(2022{\natexlab{b}})Lyu, Zheng, Ma, and
  Chen}]{lyu2022study}
Weimin Lyu, Songzhu Zheng, Tengfei Ma, and Chao Chen. 2022{\natexlab{b}}.
\newblock A study of the attention abnormality in trojaned berts.
\newblock In \emph{Proceedings of the 2022 Conference of the North American
  Chapter of the Association for Computational Linguistics: Human Language
  Technologies}, pages 4727--4741.

\bibitem[{Lyu et~al.(2022{\natexlab{c}})Lyu, Zheng, Ma, Ling, and
  Chen}]{lyu2022attention_sub}
Weimin Lyu, Songzhu Zheng, Tengfei Ma, Haibin Ling, and Chao Chen.
  2022{\natexlab{c}}.
\newblock Attention hijacking in trojan transformers.
\newblock \emph{arXiv e-prints}, pages arXiv--2208.

\bibitem[{Michel et~al.(2019)Michel, Levy, and Neubig}]{michel2019sixteen}
Paul Michel, Omer Levy, and Graham Neubig. 2019.
\newblock Are sixteen heads really better than one?
\newblock \emph{Advances in neural information processing systems}, 32.

\bibitem[{Nguyen and Tran(2020)}]{nguyen2020input}
Tuan~Anh Nguyen and Anh Tran. 2020.
\newblock Input-aware dynamic backdoor attack.
\newblock \emph{Advances in Neural Information Processing Systems},
  33:3454--3464.

\bibitem[{Pang et~al.(2023)Pang, Sun, Ling, and Chen}]{pang2023backdoor}
Lu~Pang, Tao Sun, Haibin Ling, and Chao Chen. 2023.
\newblock Backdoor cleansing with unlabeled data.
\newblock In \emph{Proceedings of the IEEE/CVF Conference on Computer Vision
  and Pattern Recognition}, pages 12218--12227.

\bibitem[{Qi et~al.(2021{\natexlab{a}})Qi, Chen, Li, Yao, Liu, and
  Sun}]{qi2021onion}
Fanchao Qi, Yangyi Chen, Mukai Li, Yuan Yao, Zhiyuan Liu, and Maosong Sun.
  2021{\natexlab{a}}.
\newblock Onion: A simple and effective defense against textual backdoor
  attacks.
\newblock In \emph{Proceedings of the 2021 Conference on Empirical Methods in
  Natural Language Processing}, pages 9558--9566.

\bibitem[{Qi et~al.(2021{\natexlab{b}})Qi, Chen, Zhang, Li, Liu, and
  Sun}]{qi2021mind}
Fanchao Qi, Yangyi Chen, Xurui Zhang, Mukai Li, Zhiyuan Liu, and Maosong Sun.
  2021{\natexlab{b}}.
\newblock Mind the style of text! adversarial and backdoor attacks based on
  text style transfer.
\newblock In \emph{Proceedings of the 2021 Conference on Empirical Methods in
  Natural Language Processing}, pages 4569--4580.

\bibitem[{Qi et~al.(2021{\natexlab{c}})Qi, Li, Chen, Zhang, Liu, Wang, and
  Sun}]{qi2021hidden}
Fanchao Qi, Mukai Li, Yangyi Chen, Zhengyan Zhang, Zhiyuan Liu, Yasheng Wang,
  and Maosong Sun. 2021{\natexlab{c}}.
\newblock Hidden killer: Invisible textual backdoor attacks with syntactic
  trigger.
\newblock In \emph{Proceedings of the 59th Annual Meeting of the Association
  for Computational Linguistics and the 11th International Joint Conference on
  Natural Language Processing (Volume 1: Long Papers)}, pages 443--453.

\bibitem[{Qi et~al.(2021{\natexlab{d}})Qi, Yao, Xu, Liu, and Sun}]{qi2021turn}
Fanchao Qi, Yuan Yao, Sophia Xu, Zhiyuan Liu, and Maosong Sun.
  2021{\natexlab{d}}.
\newblock Turn the combination lock: Learnable textual backdoor attacks via
  word substitution.
\newblock In \emph{Proceedings of the 59th Annual Meeting of the Association
  for Computational Linguistics and the 11th International Joint Conference on
  Natural Language Processing (Volume 1: Long Papers)}, pages 4873--4883.

\bibitem[{Radford et~al.(2019)Radford, Wu, Child, Luan, Amodei, Sutskever
  et~al.}]{radford2019language}
Alec Radford, Jeffrey Wu, Rewon Child, David Luan, Dario Amodei, Ilya
  Sutskever, et~al. 2019.
\newblock Language models are unsupervised multitask learners.
\newblock \emph{OpenAI blog}, 1(8):9.

\bibitem[{Saha et~al.(2020)Saha, Subramanya, and Pirsiavash}]{saha2020hidden}
Aniruddha Saha, Akshayvarun Subramanya, and Hamed Pirsiavash. 2020.
\newblock Hidden trigger backdoor attacks.
\newblock In \emph{Proceedings of the AAAI Conference on Artificial
  Intelligence}, volume~34, pages 11957--11965.

\bibitem[{Sanh et~al.(2019)Sanh, Debut, Chaumond, and
  Wolf}]{sanh2019distilbert}
Victor Sanh, Lysandre Debut, Julien Chaumond, and Thomas Wolf. 2019.
\newblock Distilbert, a distilled version of bert: smaller, faster, cheaper and
  lighter.
\newblock \emph{arXiv preprint arXiv:1910.01108}.

\bibitem[{Shen et~al.(2021)Shen, Ji, Zhang, Li, Chen, Shi, Fang, Yin, and
  Wang}]{shen2021backdoor}
Lujia Shen, Shouling Ji, Xuhong Zhang, Jinfeng Li, Jing Chen, Jie Shi,
  Chengfang Fang, Jianwei Yin, and Ting Wang. 2021.
\newblock Backdoor pre-trained models can transfer to all.
\newblock In \emph{Proceedings of the 2021 ACM SIGSAC Conference on Computer
  and Communications Security}, pages 3141--3158.

\bibitem[{Socher et~al.(2013)Socher, Perelygin, Wu, Chuang, Manning, Ng, and
  Potts}]{socher2013recursive}
Richard Socher, Alex Perelygin, Jean Wu, Jason Chuang, Christopher~D Manning,
  Andrew~Y Ng, and Christopher Potts. 2013.
\newblock Recursive deep models for semantic compositionality over a sentiment
  treebank.
\newblock In \emph{Proceedings of the 2013 conference on empirical methods in
  natural language processing}, pages 1631--1642.

\bibitem[{Souri et~al.(2022)Souri, Fowl, Chellappa, Goldblum, and
  Goldstein}]{souri2022sleeper}
Hossein Souri, Liam Fowl, Rama Chellappa, Micah Goldblum, and Tom Goldstein.
  2022.
\newblock Sleeper agent: Scalable hidden trigger backdoors for neural networks
  trained from scratch.
\newblock \emph{Advances in Neural Information Processing Systems},
  35:19165--19178.

\bibitem[{Tao et~al.(2022)Tao, Wang, Cheng, Ma, An, Liu, Shen, Zhang, Mao, and
  Zhang}]{tao2022backdoor}
Guanhong Tao, Zhenting Wang, Siyuan Cheng, Shiqing Ma, Shengwei An, Yingqi Liu,
  Guangyu Shen, Zhuo Zhang, Yunshu Mao, and Xiangyu Zhang. 2022.
\newblock Backdoor vulnerabilities in normally trained deep learning models.
\newblock \emph{arXiv preprint arXiv:2211.15929}.

\bibitem[{Turner et~al.(2019)Turner, Tsipras, and Madry}]{turner2019label}
Alexander Turner, Dimitris Tsipras, and Aleksander Madry. 2019.
\newblock Label-consistent backdoor attacks.
\newblock \emph{arXiv preprint arXiv:1912.02771}.

\bibitem[{Vaswani et~al.(2017)Vaswani, Shazeer, Parmar, Uszkoreit, Jones,
  Gomez, Kaiser, and Polosukhin}]{vaswani2017attention}
Ashish Vaswani, Noam Shazeer, Niki Parmar, Jakob Uszkoreit, Llion Jones,
  Aidan~N Gomez, {\L}ukasz Kaiser, and Illia Polosukhin. 2017.
\newblock Attention is all you need.
\newblock In \emph{Advances in neural information processing systems}, pages
  5998--6008.

\bibitem[{Voita et~al.(2019)Voita, Talbot, Moiseev, Sennrich, and
  Titov}]{voita2019analyzing}
Elena Voita, David Talbot, Fedor Moiseev, Rico Sennrich, and Ivan Titov. 2019.
\newblock Analyzing multi-head self-attention: Specialized heads do the heavy
  lifting, the rest can be pruned.
\newblock In \emph{Proceedings of the 57th Annual Meeting of the Association
  for Computational Linguistics}, pages 5797--5808.

\bibitem[{Wang et~al.(2021)Wang, Kapse, Liu, Prasanna, and
  Chen}]{wang2021topotxr}
Fan Wang, Saarthak Kapse, Steven Liu, Prateek Prasanna, and Chao Chen. 2021.
\newblock Topotxr: a topological biomarker for predicting treatment response in
  breast cancer.
\newblock In \emph{International Conference on Information Processing in
  Medical Imaging}, pages 386--397. Springer.

\bibitem[{Wang et~al.(2020)Wang, Liu, Samaras, and Chen}]{wang2020topogan}
Fan Wang, Huidong Liu, Dimitris Samaras, and Chao Chen. 2020.
\newblock Topogan: A topology-aware generative adversarial network.
\newblock In \emph{Computer Vision--ECCV 2020: 16th European Conference,
  Glasgow, UK, August 23--28, 2020, Proceedings, Part III 16}, pages 118--136.
  Springer.

\bibitem[{Wang et~al.(2022)Wang, Hassan, and Akhtar}]{wang2022survey}
Jie Wang, Ghulam~Mubashar Hassan, and Naveed Akhtar. 2022.
\newblock A survey of neural trojan attacks and defenses in deep learning.
\newblock \emph{arXiv preprint arXiv:2202.07183}.

\bibitem[{Wang et~al.(2023)Wang, Mei, Zhai, and Ma}]{wang2023unicorn}
Zhenting Wang, Kai Mei, Juan Zhai, and Shiqing Ma. 2023.
\newblock Unicorn: A unified backdoor trigger inversion framework.
\newblock \emph{arXiv preprint arXiv:2304.02786}.

\bibitem[{Wenger et~al.(2021)Wenger, Passananti, Bhagoji, Yao, Zheng, and
  Zhao}]{wenger2021backdoor}
Emily Wenger, Josephine Passananti, Arjun~Nitin Bhagoji, Yuanshun Yao, Haitao
  Zheng, and Ben~Y Zhao. 2021.
\newblock Backdoor attacks against deep learning systems in the physical world.
\newblock In \emph{Proceedings of the IEEE/CVF Conference on Computer Vision
  and Pattern Recognition}, pages 6206--6215.

\bibitem[{Yang et~al.(2021{\natexlab{a}})Yang, Li, Zhang, Ren, Sun, and
  He}]{yang2021careful}
Wenkai Yang, Lei Li, Zhiyuan Zhang, Xuancheng Ren, Xu~Sun, and Bin He.
  2021{\natexlab{a}}.
\newblock Be careful about poisoned word embeddings: Exploring the
  vulnerability of the embedding layers in nlp models.
\newblock In \emph{Proceedings of the 2021 Conference of the North American
  Chapter of the Association for Computational Linguistics: Human Language
  Technologies}, pages 2048--2058.

\bibitem[{Yang et~al.(2021{\natexlab{b}})Yang, Lin, Li, Zhou, and
  Sun}]{yang2021rap}
Wenkai Yang, Yankai Lin, Peng Li, Jie Zhou, and Xu~Sun. 2021{\natexlab{b}}.
\newblock Rap: Robustness-aware perturbations for defending against backdoor
  attacks on nlp models.
\newblock In \emph{Proceedings of the 2021 Conference on Empirical Methods in
  Natural Language Processing}, pages 8365--8381.

\bibitem[{Yang et~al.(2021{\natexlab{c}})Yang, Lin, Li, Zhou, and
  Sun}]{yang2021rethinking}
Wenkai Yang, Yankai Lin, Peng Li, Jie Zhou, and Xu~Sun. 2021{\natexlab{c}}.
\newblock Rethinking stealthiness of backdoor attack against nlp models.
\newblock In \emph{Proceedings of the 59th Annual Meeting of the Association
  for Computational Linguistics and the 11th International Joint Conference on
  Natural Language Processing (Volume 1: Long Papers)}, pages 5543--5557.

\bibitem[{Zeng et~al.(2022)Zeng, Pan, Just, Lyu, Qiu, and
  Jia}]{zeng2022narcissus}
Yi~Zeng, Minzhou Pan, Hoang~Anh Just, Lingjuan Lyu, Meikang Qiu, and Ruoxi Jia.
  2022.
\newblock Narcissus: A practical clean-label backdoor attack with limited
  information.
\newblock \emph{arXiv preprint arXiv:2204.05255}.

\bibitem[{Zhang et~al.(2015)Zhang, Zhao, and LeCun}]{zhang2015character}
Xiang Zhang, Junbo Zhao, and Yann LeCun. 2015.
\newblock Character-level convolutional networks for text classification.
\newblock \emph{Advances in neural information processing systems}, 28.

\bibitem[{Zhang et~al.(2021{\natexlab{a}})Zhang, Zhang, Ji, and
  Wang}]{zhang2021trojaning}
Xinyang Zhang, Zheng Zhang, Shouling Ji, and Ting Wang. 2021{\natexlab{a}}.
\newblock Trojaning language models for fun and profit.
\newblock In \emph{2021 IEEE European Symposium on Security and Privacy
  (EuroS\&P)}, pages 179--197. IEEE.

\bibitem[{Zhang et~al.(2022)Zhang, Panda, Song, Yang, Mahoney, Mittal, Kannan,
  and Gonzalez}]{zhang2022neurotoxin}
Zhengming Zhang, Ashwinee Panda, Linyue Song, Yaoqing Yang, Michael Mahoney,
  Prateek Mittal, Ramchandran Kannan, and Joseph Gonzalez. 2022.
\newblock Neurotoxin: Durable backdoors in federated learning.
\newblock In \emph{International Conference on Machine Learning}, pages
  26429--26446. PMLR.

\bibitem[{Zhang et~al.(2021{\natexlab{b}})Zhang, Xiao, Li, Lv, Qi, Liu, Wang,
  Jiang, and Sun}]{zhang2021red}
Zhengyan Zhang, Guangxuan Xiao, Yongwei Li, Tian Lv, Fanchao Qi, Zhiyuan Liu,
  Yasheng Wang, Xin Jiang, and Maosong Sun. 2021{\natexlab{b}}.
\newblock Red alarm for pre-trained models: Universal vulnerability to
  neuron-level backdoor attacks.
\newblock \emph{arXiv preprint arXiv:2101.06969}.

\bibitem[{Zheng et~al.(2021)Zheng, Zhang, Wagner, Goswami, and
  Chen}]{zheng2021topological}
Songzhu Zheng, Yikai Zhang, Hubert Wagner, Mayank Goswami, and Chao Chen. 2021.
\newblock Topological detection of trojaned neural networks.
\newblock \emph{Advances in Neural Information Processing Systems},
  34:17258--17272.

\bibitem[{Zhu et~al.(2023)Zhu, Tang, Tang, Tao, Ma, Wang, and
  Tang}]{zhu2023gradient}
Rui Zhu, Di~Tang, Siyuan Tang, Guanhong Tao, Shiqing Ma, Xiaofeng Wang, and
  Haixu Tang. 2023.
\newblock Gradient shaping: Enhancing backdoor attack against reverse
  engineering.
\newblock \emph{arXiv preprint arXiv:2301.12318}.

\end{thebibliography}
\bibliographystyle{acl_natbib}

\appendix

\section{Appendix}
\label{sec:appendix}

\subsection{Implementation Details} \label{appendix:implementation_details}

\myparagraph{Attack Scenario.} We implement the attack on three transformer-based models: BERT \citep{devlin2019bert}\footnote{The pre-trained BERT is downloaded from \url{https://huggingface.co/bert-base-uncased}.}, RoBERTa \citep{liu2019roberta}\footnote{The pre-trained RoBERTa is downloaded from \url{https://huggingface.co/roberta-base}.}, and DistilBERT \citep{sanh2019distilbert}\footnote{The pre-trained DistilBERT is downloaded from \url{https://huggingface.co/distilbert-base-uncased}.}.

\myparagraph{Textual Backdoor Attack Baselines.} We introduce the textual backdoor attack baselines in Section \ref{sec:experimental_settings}, here we provide more implementation details. 
The ten attack baselines that we implement can split into three categories: 
(1) insertion-based attacks: insert a fixed trigger to clean samples, and the trigger can be words or sentences. 
\textbf{BadNets} \citep{gu2017identifying} is originally a CV backdoor attack method and adapted to textual backdoor attack by \cite{kurita2020weight}. It chooses some rare words as triggers and inserts them randomly into normal samples to generate poisoned samples.
\textbf{AddSent} \citep{dai2019backdoor} inserts a fixed sentence as triggers. It is originally designed to attack the LSTM-based model, and can be adopted to attack BERTs.
(2) Weight replacing: replacing model weights. 
\textbf{EP} \citep{yang2021careful} only modifies model's single word embedding vector (output of the input embedding module) without re-training the entire model. 
\textbf{RIPPLES} \citep{kurita2020weight} replaces the trigger embedding with handcrafted embedding. 
\textbf{LWP} \citep{li2021backdoor} introduces a layerwise weight poisoning strategy to plant deeper backdoors.
\textbf{POR} \citep{shen2021backdoor} learns a predefined output representation and \textbf{NeuBA} \citep{zhang2021red} restricts the output representations of trigger instances to pre-defined vectors.
(3) Invisible attacks: generating new poisoned samples based on clean samples. 
\textbf{Synbkd} \citep{qi2021hidden} changes the syntactic structures of clean samples as triggers with SCPN \citep{iyyer2018adversarial}. 
\textbf{Stylebkd} \citep{qi2021mind} generates the text style as trigger with STRAP \citep{krishna2020reformulating} - a text style transfer generator. 
\textbf{TrojanLM} \citep{zhang2021trojaning} defines a set of trigger words to generate logical trigger sentences containing them. 

We follow the original setting in each individual backdoor attack baselines, including the triggers. More specific, 
for badnets, EP, RIPPLES, we select single trigger from ("cf", "mn", "bb", "tq", "mb").
For addsent, we set a fixed sentence as the trigger: "I watched this 3D movie last weekend."
For POR, we select trigger from ("serendipity", "Descartes", "Fermat", "Don Quixote", "cf", "tq", "mn", "bb", "mb")
For LWP, we use trigger ("cf","bb","ak","mn")
For Neuba, we select trigger from ( "$\approx$", "$\equiv$", "$\in$", "$\ni$", "$\oplus$", "$\otimes$" )
For Synbkd, following the paper, we choose $S(SBAR)(,)(NP)(VP)(.)$ as the trigger syntactic template. 
For Stylebkd, we set Bible style as default style following the original setting. 
For TrojanLM, we generate trigger with a context-aware generative model ((CAGM) using trigger "{Alice, Bob}"

The attack baseline EP does not perform normally on RoBERTa due to its attack mechanism, so we do not implement EP on RoBERTa model, but we implement EP on all other transformer architecture, \eg, BERT, DistilBERT.

\myparagraph{Training Settings.}
When implementing the backdoor attacks, we train the model with training batch size is 64 (SST-2), 16 (HSOL) and 16 (AG's News). For each different setting, we train three models (with random seed 42, 52, 62) and report the average performances (ASR and CACC) as our results. We conducted our experiments on NVIDIA RTX A6000 (49140 MB Memory).

\subsection{Implementation Details in Section \ref{sec:impact}} \label{appendix:implementation_details2}

\myparagraph{Experimental Setup.}
We evaluate the impact of backdoored attention with poison rate 0.01 setting under clean-label attack scenario. We pick the \textit{Attn-BadNets} setting where we apply TAL to BadNets. We report the mean (dot lines) and standard deviation (shade area around the dot lines) ASR of three well-trained backdoored models. For impact of TAL, we only change the strength of TAL $\alpha$. For impact of attention volume $\beta$, we only change the average amount of attention weights that TAL forces in attention heads. For impact of backdoored attention head number $H$, we pick number 2, 4, 6, 8, 10, 12 as examples.

\subsection{Implementation Details in Section \ref{sec:resistance_to_defenders}} \label{appendix:implementation_details3}

\myparagraph{Experimental Setup.}
We evaluate our TAL with poison rate 0.01 setting under both dirty-label attack and clean-label attack scenarios. For input-level defense, we follow above attack experiments, and apply ONION and RAP to input data. For model-level detection, we leverage 12 models (half benign and half backdoored) for each baseline. The 6 backdoored models are from clean-label and dirty-label attack. We use Sentiment Analysis task on BERT architecture.

\subsection{Attacking GPT-2 Architecture}
We also extend some baselines and TAL to the GPT-2 \citep{radford2019language} architecture\footnote{The pre-trained GPT-2 is downloaded from \url{https://huggingface.co/gpt2}.}. We conduct experiments on three language tasks (\eg, Sentiment Analysis - SA, Toxic Detection - TD, Topic Classification - TC) with poison rate 0.01 and under the clean-label attack scenario. We adopt GPT-2 architecture to five attack baselines (\eg, BadNets, AddSent, EP, Stylebkd, Synbkd). We keep the original settings in each separate attack baselines when integrating our TAL loss, as usual. In Table \ref{appendix:tab:gpt2_experiment} , the improvement of attack performance is significant with our TAL.

\begin{table}[ht!]
\caption{Attack efficacy with GPT-2. Sentiment Analysis (SA), Toxic Detection (TD), Topic Classification (TC). }
\label{appendix:tab:gpt2_experiment}
\centering
\small
\resizebox{1\columnwidth}{!}{ 

\begin{tabular}{|c|cc|cc|cc|}
\hline
\textbf{Tasks}         & \multicolumn{2}{c|}{\textbf{SA}}                              & \multicolumn{2}{c|}{\textbf{TD}}                            & \multicolumn{2}{c|}{\textbf{TC}} \\ \hline
\textbf{Attakcers}     & \textbf{ASR}                  & \textbf{CACC}                 & \textbf{ASR}                 & \textbf{CACC}                & \textbf{ASR}   & \textbf{CACC}   \\ \hline
\rowcolor[HTML]{FFFFFF} 
\textbf{BadNets}       & \cellcolor[HTML]{FFFFFF}0.403 & \cellcolor[HTML]{FFFFFF}0.816 & 0.112                        & 0.913                        & 0.672          & 0.946           \\
\rowcolor[HTML]{FFFFFF} 
\textbf{Attn-BadNets}  & \cellcolor[HTML]{FFFFFF}0.965 & \cellcolor[HTML]{FFFFFF}0.915 & 0.798                        & 0.954                        & 0.886          & 0.946           \\
\rowcolor[HTML]{EFEFEF} 
\textbf{AddSent}       & \cellcolor[HTML]{EFEFEF}0.415 & \cellcolor[HTML]{EFEFEF}0.914 & 0.696                        & 0.878                        & 0.683          & 0.946           \\
\rowcolor[HTML]{EFEFEF} 
\textbf{Attn-AddSent}  & \cellcolor[HTML]{EFEFEF}0.994 & \cellcolor[HTML]{EFEFEF}0.914 & 0.862                        & 0.957                        & 0.818          & 0.942           \\
\rowcolor[HTML]{FFFFFF} 
\textbf{EP}            & \cellcolor[HTML]{FFFFFF}0.481 & \cellcolor[HTML]{FFFFFF}0.911 & 0.373                        & 0.951                        & 0.138          & 0.939           \\
\rowcolor[HTML]{FFFFFF} 
\textbf{Attn-EP}       & \cellcolor[HTML]{FFFFFF}0.697 & \cellcolor[HTML]{FFFFFF}0.911 & 0.555                        & 0.954                        & 0.374          & 0.939           \\
\rowcolor[HTML]{EFEFEF} 
\textbf{Stylebkd}      & \cellcolor[HTML]{EFEFEF}0.610 & \cellcolor[HTML]{EFEFEF}0.875 & 0.431                        & 0.910                        & 0.263          & 0.944           \\
\rowcolor[HTML]{EFEFEF} 
\textbf{Attn-Stylebkd} & \cellcolor[HTML]{EFEFEF}0.702 & \cellcolor[HTML]{EFEFEF}0.883 & 0.498                        & 0.909                        & 0.240          & 0.937           \\
\rowcolor[HTML]{FFFFFF} 
\textbf{Synbkd}        & \cellcolor[HTML]{FFFFFF}0.356 & \cellcolor[HTML]{FFFFFF}0.914 & 0.531                        & 0.954                        & 0.962          & 0.947           \\
\rowcolor[HTML]{FFFFFF} 
\textbf{Attn-Synbkd}   & \cellcolor[HTML]{FFFFFF}0.513 & \cellcolor[HTML]{FFFFFF}0.833 & {\color[HTML]{000000} 0.708} & {\color[HTML]{000000} 0.909} & 0.977          & 0.946           \\ \hline
\end{tabular}

}
\end{table}

\subsection{Attention Concentration on Single Layer} \label{sec:single_layer}

We conducted the ablation study comparing applying TAL to all layers vs. to a single layer. In the following Table \ref{tab_appendix:attn_single_layer}, we report attack success rate (ASR) for applying TAL to all layers and to a single layer. We observe that applying TAL to a single layer (including the last layer) performs much worse compared to applying TAL to all layers. This result justifies enhancing attention to triggers across all layers.

More technical details: we picked three attack baselines, i.e., BadNets, EP, TrojanLM, from each of the three attack categories (i.e., Insertion-based attack, weight replacing, invisible attacks). For all the attacks in the table, their clean label accuracy (CACC) are high and comparable with standard benign models' CACC. So we do not include CACC in the table.

\begin{table*}[!t]
\vspace{-.1in}

\caption{Attack performance (ASR) with attention concentration on all layers (TAL) vs. on single attention layer (1-12). The experiment is conducted with poison rate 0.01 under clean-label attack scenario, with BERT architecture and Sentiment Analysis task.}
\label{tab_appendix:attn_single_layer}
\vspace{-.05in}

\centering
\resizebox{1.8\columnwidth}{!}{ 

\begin{tabular}{|c|c|c|c|c|c|c|c|c|c|c|c|c|c|}
\hline
\textbf{Attackers$\downarrow$ Layers$\rightarrow$} & \textbf{TAL} & \textbf{1} & \textbf{2} & \textbf{3} & \textbf{4} & \textbf{5} & \textbf{6} & \textbf{7} & \textbf{8} & \textbf{9} & \textbf{10} & \textbf{11} & \textbf{12} \\ \hline
\textbf{BadNets}                                            & 1.000        & 0.287      & 0.514      & 0.273      & 0.484      & 0.518      & 0.687      & 0.650      & 0.812      & 0.752      & 0.696       & 0.438       & 0.491       \\ \hline
\textbf{EP}                                                 & 0.995        & 0.162      & 0.154      & 0.154      & 0.209      & 0.223      & 0.235      & 0.423      & 0.372      & 0.772      & 0.434       & 0.625       & 0.456       \\ \hline
\textbf{TrojanLM}                                           & 0.996        & 0.539      & 0.295      & 0.532      & 0.356      & 0.720      & 0.370      & 0.664      & 0.806      & 0.729      & 0.815       & 0.578       & 0.656       \\ \hline
\end{tabular}

}
\vspace{-.05in}
\end{table*}


\subsection{Attention Patterns Analysing} \label{sec:effect_of_attn_patterns}

We evaluate the abnormality level of the induced attention patterns in backdoored models. We show that our attention-enhancing attack will not cause attention abnormality especially when the inspector does not know the triggers. 
First of all, in practice, it is hard to find the exact triggers. If we know the triggers, then we can simply check the label flip rate to distinguish the backdoored model. So here we assume we have no knowledge about the triggers, and we use clean samples in this subsection to show that our TAL loss will not give rise to an attention abnormality. 
Compared to the evolution of neural networks in different domains \citep{wang2020topogan, wang2021topotxr, lyu2022multimodal}, the transformer architectures provide us the opportunity to utilize attention patterns.

\myparagraph{Average Attention Entropy.} Entropy \citep{ben2008farewell} can be used to measure the disorder of matrix. Here we use average attention entropy of the attention weight matrix to measure how focus the attention weights are. Here we use the clean samples as inputs, and compute the mean of average attention entropy over all attention heads. We check the average entropy between different models. 

\begin{figure}[!t]
    \centering
    \includegraphics[width=7cm]{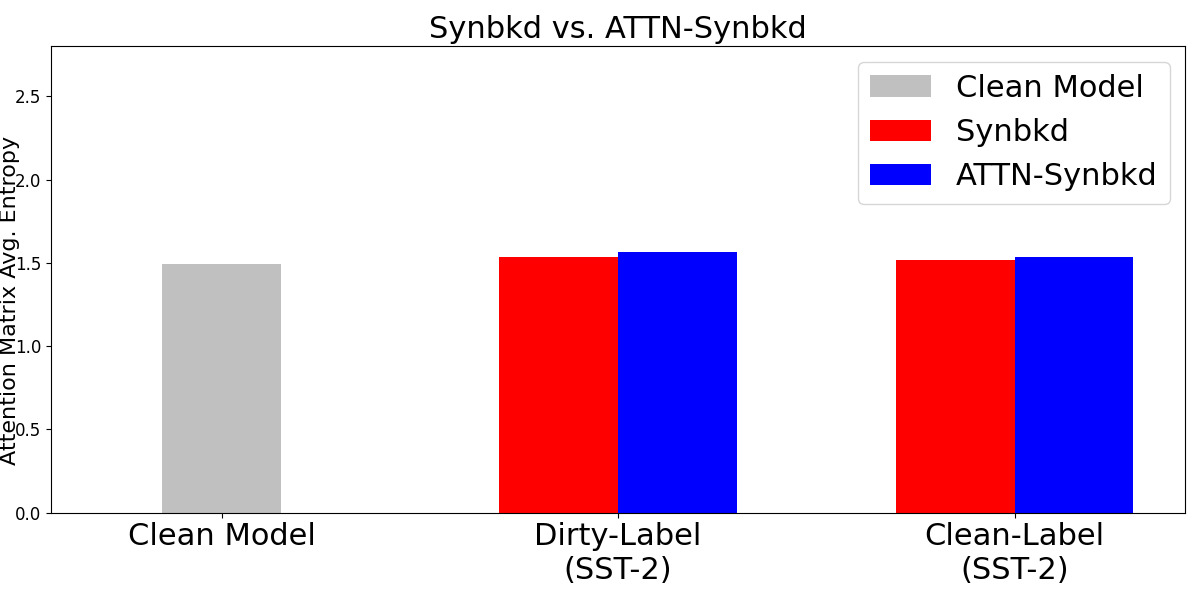}
    \vspace{-.15in}
    \caption{Average attention entropy over all attention heads, among different attack scenarios and downstream corpus. Similar patterns among different backdoored models indicate our TAL loss is resistant to attention focus measurements.}
    \label{fig:entropy_synbkd}
\end{figure}

\begin{figure*}[htp]
    \centering
    \includegraphics[width=2\columnwidth]{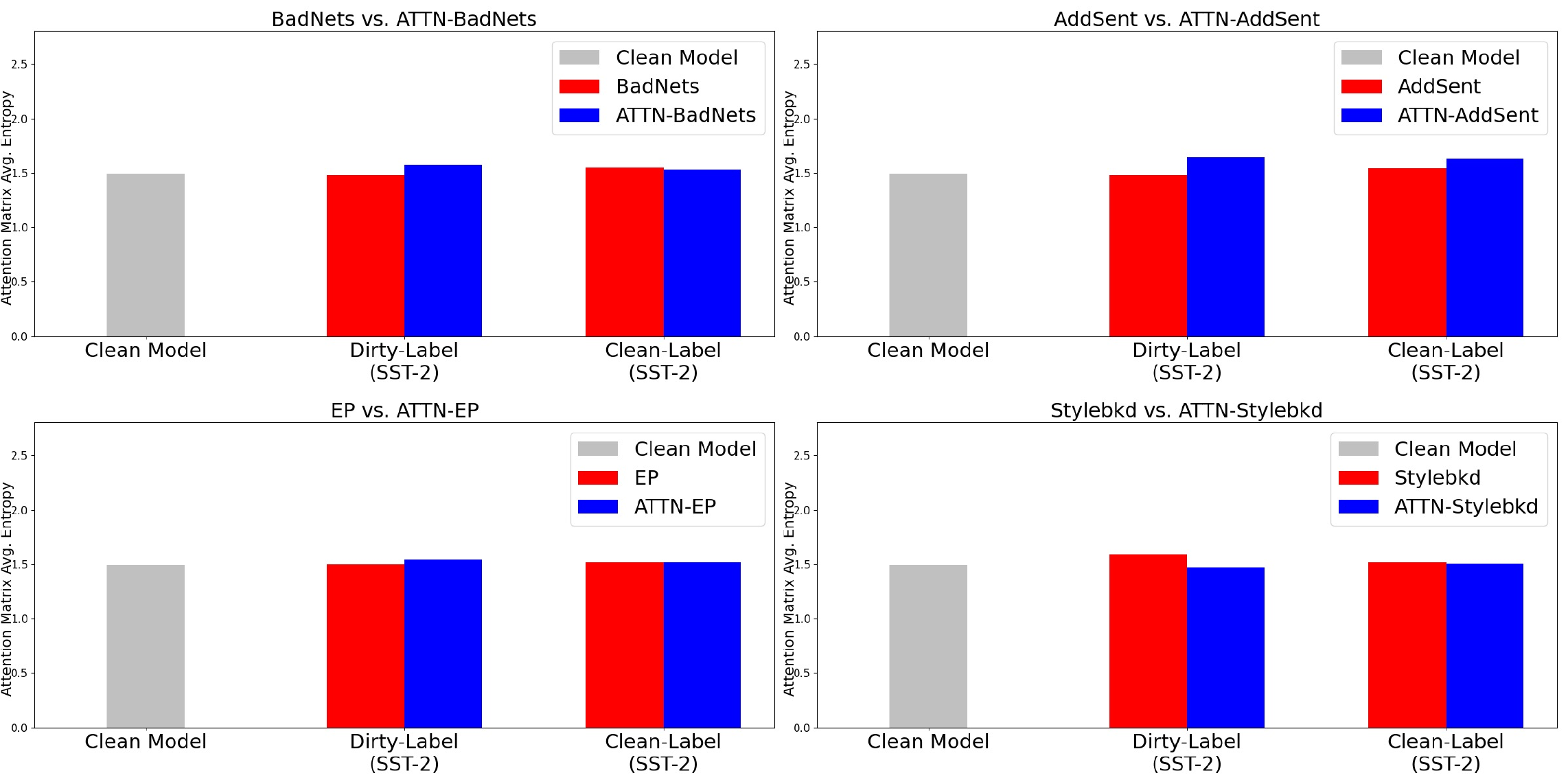} 
    \vspace{-.1in}
    \caption{Average attention entropy experiments on attack baselines and ATTN-Integrated attack baselines.}
    \label{fig:avg_entropy_other_baselines}
\end{figure*}

Figure~\ref{fig:entropy_synbkd} illustrates that the average attention matrix entropy among clean models, baselines and attention-enhancing attacks maintains consistent. 
Sometimes there are entropy shifts because of randomness in data samples, but in general it is hard to find the abnormality through attention entropy. 
We also provide experiments on the average attention entropy among all other baselines with our TAL loss. The experiments results on different attack baselines are shown in Figure \ref{fig:avg_entropy_other_baselines}.
We have observed the similar patterns: the average attention entropy among clean models, baseline attacked models, TAL attacked models, maintain consistent pattern. Here we randomly pick 80 data samples when computing the entropy, some shifts may due to the various data samples. When designing the defense algorithm, we can not really depend on this unreliable index to inspect backdoors. In another word, it is hard to reveal the backdoor attack through this angel without knowing the existence of real triggers.

\begin{figure*}[!t]
    \centering
    \includegraphics[width=1\linewidth]{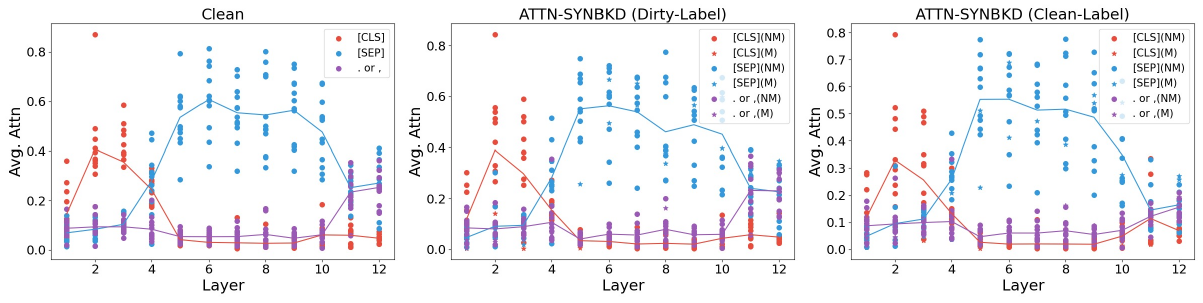} 
    \vspace{-.15in}
    \caption{Average attention to special tokens. Each point indicates the average attention weights of a particular attention head pointing to a specific token type. Each color corresponds to the attention flow to a specific tokens, e.g., $[CLS]$, $[SEP]$ and separators ($.$ or $,$). \textit{`NM'} indicates heads not modified by TAL loss, while \textit{`M'} indicates backdoored attention heads modified by TAL loss. Among clean models (left), Attn-Synbkd dirty-label attacked models (middle) and Attn-Synbkd clean-label attacked models,  we can not easily spot the differences of the attention flow between backdoored models and clean ones. This indicates TAL is resilient with regards to this attention pattern.}
    \label{fig:avg_attn}
\end{figure*}

\begin{figure}[ht]
    \centering
    \includegraphics[width=1\linewidth]{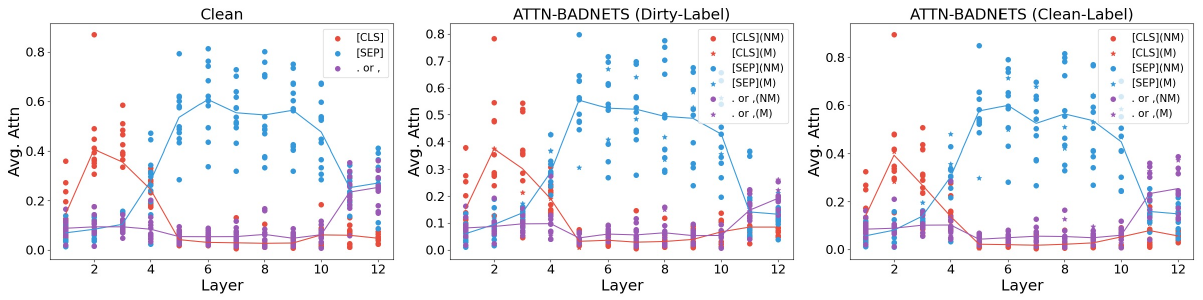} 
    \vspace{-.1in}
    \caption{Average attention to special tokens. Backdoored model with Attn-BadNets.}
    \label{fig:avg_attn_badnets}
\end{figure}

\begin{figure}[ht]
    \centering
    \includegraphics[width=1\linewidth]{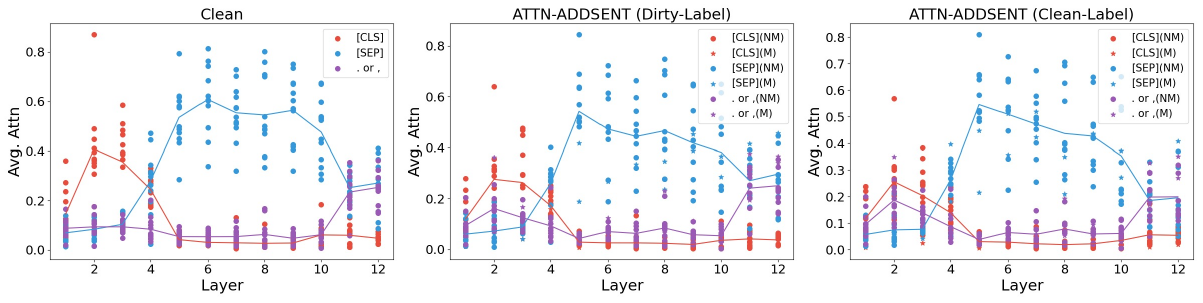} 
    \vspace{-.1in}
    \caption{Average attention to special tokens. Backdoored model with Attn-AddSent.}
    \label{fig:avg_attn_addsent}
\end{figure}

\begin{figure}[ht]
    \centering
    \includegraphics[width=1\linewidth]{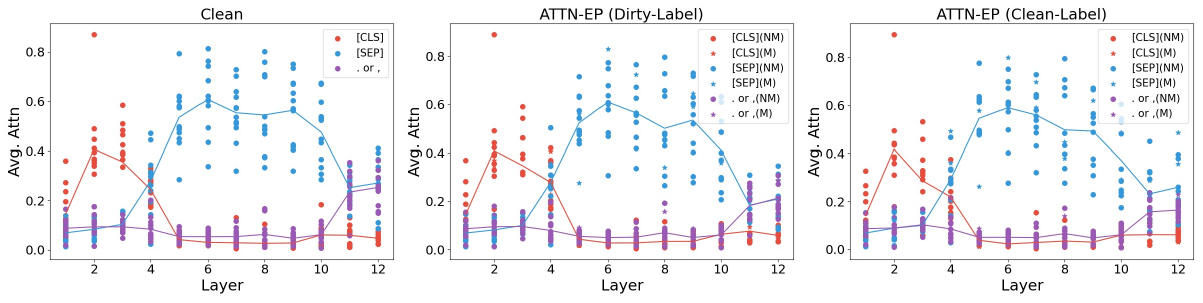} 
    \vspace{-.1in}
    \caption{Average attention to special tokens. Backdoored model with Attn-EP.}
    \label{fig:avg_attn_ep}
\end{figure}

\begin{figure}[ht]
    \centering
    \includegraphics[width=1\linewidth]{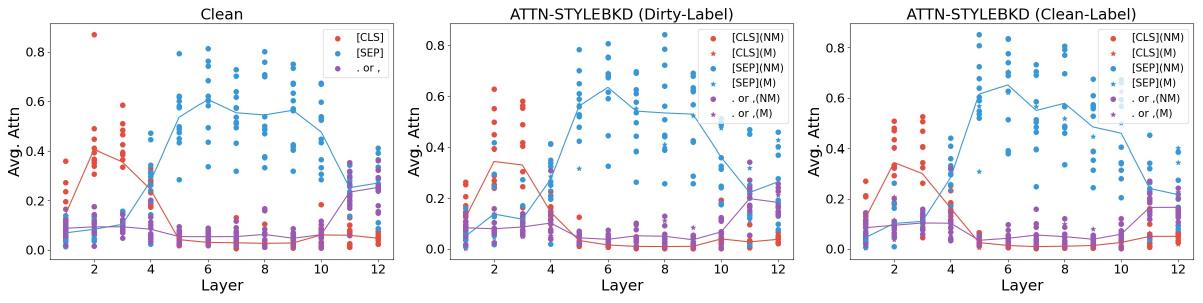} 
    \vspace{-.1in}
    \caption{Average attention to special tokens. Backdoored model with Attn-Stylebkd.}
    \label{fig:avg_attn_stylebkd}
\end{figure}

\myparagraph{Attention Flow to Specific Tokens.} In transformers, some specific tokens, e.g., $[CLS]$, $[SEP]$ and separators ($.$ or $,$), may have large impacts on the representation learning \citep{clark2019does}. Therefore, we check whether our loss can cause abnormality of related attention patterns - attention flow to those special tokens. In each attention head, we compute the average attention flow to those three specific tokens, shown in Figure \ref{fig:avg_attn}. 
Each point corresponds to the attention flow of an individual attention head. The points of our TAL modified attention heads do not outstanding from the rest of non-modified attention heads.
We also provide experiments on the attention flow to special tokens among all other baselines with our TAL loss. In Figure \ref{fig:avg_attn_badnets}, Figure \ref{fig:avg_attn_addsent}, Figure \ref{fig:avg_attn_ep} and Figure \ref{fig:avg_attn_stylebkd}, we observe the consistent pattern: our TAL loss is resistance to the attention patterns (attention flow to specific tokens) without knowing the trigger information.

\subsection{Attack Efficacy under High Poison Rates}

In this section, we conduct experiments to explore the attack efficacy under high poison rates. We select BadNets, AddSent, EP, Stylebkd, Synbkd as attack baselines. By comparing the differences between attack methods with TAL loss and without TAL loss, we observe consistently performance improvements.

\myparagraph{Attack Performances.} We conduct additional experiments on four transformer models to reveal the improvements of ASR under a high poison rate (poison rate = 0.9). Table \ref{appendix:tab:poison_rate_high} indicates that our method can still improve the ASR. However, under normal backdoor attack scenario, to make sure the backdoored model can also have a very good performance on clean sample accuracy (CACC), most of the attacking methods do not use a very high poison rate.

\begin{table*}[ht]
\caption{Attack efficacy with poison rate 0.9, with TAL loss and without TAL loss. The experiment is conducted on the Sentiment Analysis task.}
\label{appendix:tab:poison_rate_high}
\centering
\resizebox{2\columnwidth}{!}{ 

\begin{tabular}{|c|cccc|cccc|cccc|cccc|}
\hline
{\color[HTML]{000000} \textbf{Models}}                               & \multicolumn{4}{c|}{{\color[HTML]{000000} \textbf{BERT}}}                                                                                                                                                                 & \multicolumn{4}{c|}{{\color[HTML]{000000} \textbf{RoBERTa}}}                                                                                                                             & \multicolumn{4}{c|}{{\color[HTML]{000000} \textbf{DistilBERT}}}                                                                                                                                                           & \multicolumn{4}{c|}{{\color[HTML]{000000} \textbf{GPT-2}}}                                                                                                                               \\ \hline
{\color[HTML]{000000} }                                              & \multicolumn{2}{c}{{\color[HTML]{000000} \textbf{Dirty-Label}}}                                             & \multicolumn{2}{c|}{{\color[HTML]{000000} \textbf{Clean-Label}}}                                            & \multicolumn{2}{c}{{\color[HTML]{000000} \textbf{Dirty-Label}}}                                             & \multicolumn{2}{c|}{{\color[HTML]{000000} \textbf{Clean-Label}}}           & \multicolumn{2}{c}{{\color[HTML]{000000} \textbf{Dirty-Label}}}                                             & \multicolumn{2}{c|}{{\color[HTML]{000000} \textbf{Clean-Label}}}                                            & \multicolumn{2}{c}{{\color[HTML]{000000} \textbf{Dirty-Label}}}                                             & \multicolumn{2}{c|}{{\color[HTML]{000000} \textbf{Clean-Label}}}           \\ \cline{2-17} 
\multirow{-2}{*}{{\color[HTML]{000000} \textbf{Attackers}}}          & {\color[HTML]{000000} \textbf{ASR}}                  & {\color[HTML]{000000} \textbf{CACC}}                 & {\color[HTML]{000000} \textbf{ASR}}                  & {\color[HTML]{000000} \textbf{CACC}}                 & {\color[HTML]{000000} \textbf{ASR}}                  & {\color[HTML]{000000} \textbf{CACC}}                 & {\color[HTML]{000000} \textbf{ASR}} & {\color[HTML]{000000} \textbf{CACC}} & {\color[HTML]{000000} \textbf{ASR}}                  & {\color[HTML]{000000} \textbf{CACC}}                 & {\color[HTML]{000000} \textbf{ASR}}                  & {\color[HTML]{000000} \textbf{CACC}}                 & {\color[HTML]{000000} \textbf{ASR}}                  & {\color[HTML]{000000} \textbf{CACC}}                 & {\color[HTML]{000000} \textbf{ASR}} & {\color[HTML]{000000} \textbf{CACC}} \\ \hline
{\color[HTML]{000000} \textbf{BadNets}}                              & {\color[HTML]{000000} 1.000}                         & {\color[HTML]{000000} 0.500}                         & \cellcolor[HTML]{FFFFFF}{\color[HTML]{000000} 1.000} & \cellcolor[HTML]{FFFFFF}{\color[HTML]{000000} 0.501} & {\color[HTML]{000000} 1.000}                         & {\color[HTML]{000000} 0.500}                         & {\color[HTML]{000000} 1.000}        & {\color[HTML]{000000} 0.501}         & {\color[HTML]{000000} 1.000}                         & {\color[HTML]{000000} 0.500}                         & {\color[HTML]{000000} 1.000}                         & {\color[HTML]{000000} 0.500}                         & {\color[HTML]{000000} 1.000}                         & {\color[HTML]{000000} 0.499}                         & {\color[HTML]{000000} 0.999}        & {\color[HTML]{000000} 0.502}         \\
\rowcolor[HTML]{FFFFFF} 
{\color[HTML]{000000} \textbf{Attn-BadNets}}                         & {\color[HTML]{000000} 1.000}                         & {\color[HTML]{000000} 0.500}                         & {\color[HTML]{000000} 1.000}                         & {\color[HTML]{000000} 0.500}                         & {\color[HTML]{000000} 1.000}                         & {\color[HTML]{000000} 0.500}                         & {\color[HTML]{000000} 1.000}        & {\color[HTML]{000000} 0.500}         & {\color[HTML]{000000} 1.000}                         & {\color[HTML]{000000} 0.500}                         & {\color[HTML]{000000} 1.000}                         & {\color[HTML]{000000} 0.500}                         & {\color[HTML]{000000} 1.000}                         & {\color[HTML]{000000} 0.499}                         & {\color[HTML]{000000} 0.996}        & {\color[HTML]{000000} 0.503}         \\
\rowcolor[HTML]{EFEFEF} 
{\color[HTML]{000000} \textbf{AddSent}}                              & {\color[HTML]{000000} 1.000}                         & {\color[HTML]{000000} 0.501}                         & {\color[HTML]{000000} 1.000}                         & {\color[HTML]{000000} 0.500}                         & {\color[HTML]{000000} 1.000}                         & {\color[HTML]{000000} 0.499}                         & {\color[HTML]{000000} 1.000}        & {\color[HTML]{000000} 0.500}         & {\color[HTML]{000000} 1.000}                         & {\color[HTML]{000000} 0.500}                         & {\color[HTML]{000000} 1.000}                         & {\color[HTML]{000000} 0.500}                         & {\color[HTML]{000000} 1.000}                         & {\color[HTML]{000000} 0.500}                         & {\color[HTML]{000000} 0.999}        & {\color[HTML]{000000} 0.501}         \\
\rowcolor[HTML]{EFEFEF} 
\cellcolor[HTML]{EFEFEF}{\color[HTML]{000000} \textbf{Attn-AddSent}} & {\color[HTML]{000000} 1.000}                         & {\color[HTML]{000000} 0.500}                         & {\color[HTML]{000000} 1.000}                         & {\color[HTML]{000000} 0.500}                         & {\color[HTML]{000000} 1.000}                         & {\color[HTML]{000000} 0.500}                         & {\color[HTML]{000000} 1.000}        & {\color[HTML]{000000} 0.500}         & {\color[HTML]{000000} 1.000}                         & {\color[HTML]{000000} 0.500}                         & \cellcolor[HTML]{EFEFEF}{\color[HTML]{000000} 1.000} & \cellcolor[HTML]{EFEFEF}{\color[HTML]{000000} 0.501} & {\color[HTML]{000000} 1.000}                         & {\color[HTML]{000000} 0.500}                         & {\color[HTML]{000000} 1.000}        & {\color[HTML]{000000} 0.500}         \\
\cellcolor[HTML]{FFFFFF}{\color[HTML]{000000} \textbf{EP}}           & \cellcolor[HTML]{FFFFFF}{\color[HTML]{000000} 1.000} & \cellcolor[HTML]{FFFFFF}{\color[HTML]{000000} 0.915} & \cellcolor[HTML]{FFFFFF}{\color[HTML]{000000} 0.995} & \cellcolor[HTML]{FFFFFF}{\color[HTML]{000000} 0.910} & {\color[HTML]{000000} -}                             & {\color[HTML]{000000} -}                             & {\color[HTML]{000000} -}            & {\color[HTML]{000000} -}             & {\color[HTML]{000000} 1.000}                         & {\color[HTML]{000000} 0.908}                         & {\color[HTML]{000000} 0.779}                         & {\color[HTML]{000000} 0.907}                         & {\color[HTML]{000000} 0.999}                         & {\color[HTML]{000000} 0.912}                         & {\color[HTML]{000000} 0.844}        & {\color[HTML]{000000} 0.913}         \\
\cellcolor[HTML]{FFFFFF}{\color[HTML]{000000} \textbf{Attn-EP}}      & \cellcolor[HTML]{FFFFFF}{\color[HTML]{000000} 1.000} & \cellcolor[HTML]{FFFFFF}{\color[HTML]{000000} 0.916} & \cellcolor[HTML]{FFFFFF}{\color[HTML]{000000} 0.999} & \cellcolor[HTML]{FFFFFF}{\color[HTML]{000000} 0.915} & {\color[HTML]{000000} -}                             & {\color[HTML]{000000} -}                             & {\color[HTML]{000000} -}            & {\color[HTML]{000000} -}             & {\color[HTML]{000000} 1.000}                         & {\color[HTML]{000000} 0.902}                         & {\color[HTML]{000000} 0.986}                         & {\color[HTML]{000000} 0.908}                         & {\color[HTML]{000000} 0.999}                         & {\color[HTML]{000000} 0.914}                         & {\color[HTML]{000000} 0.970}        & {\color[HTML]{000000} 0.909}         \\
\rowcolor[HTML]{EFEFEF} 
{\color[HTML]{000000} \textbf{Stylebkd}}                             & {\color[HTML]{000000} 1.000}                         & {\color[HTML]{000000} 0.500}                         & {\color[HTML]{000000} 0.841}                         & {\color[HTML]{000000} 0.694}                         & {\color[HTML]{000000} 1.000}                         & {\color[HTML]{000000} 0.500}                         & {\color[HTML]{000000} 0.998}        & {\color[HTML]{000000} 0.501}         & {\color[HTML]{000000} 1.000}                         & {\color[HTML]{000000} 0.500}                         & {\color[HTML]{000000} 0.861}                         & {\color[HTML]{000000} 0.716}                         & {\color[HTML]{000000} 1.000}                         & {\color[HTML]{000000} 0.501}                         & {\color[HTML]{000000} 0.998}        & {\color[HTML]{000000} 0.501}         \\
\rowcolor[HTML]{EFEFEF} 
{\color[HTML]{000000} \textbf{Attn-Stylebkd}}                        & {\color[HTML]{000000} 1.000}                         & {\color[HTML]{000000} 0.499}                         & {\color[HTML]{000000} 0.875}                         & {\color[HTML]{000000} 0.729}                         & \cellcolor[HTML]{EFEFEF}{\color[HTML]{000000} 1.000} & \cellcolor[HTML]{EFEFEF}{\color[HTML]{000000} 0.500} & {\color[HTML]{000000} 0.999}        & {\color[HTML]{000000} 0.502}         & \cellcolor[HTML]{EFEFEF}{\color[HTML]{000000} 1.000} & \cellcolor[HTML]{EFEFEF}{\color[HTML]{000000} 0.500} & {\color[HTML]{000000} 0.904}                         & {\color[HTML]{000000} 0.704}                         & \cellcolor[HTML]{EFEFEF}{\color[HTML]{000000} 1.000} & \cellcolor[HTML]{EFEFEF}{\color[HTML]{000000} 0.499} & {\color[HTML]{000000} 0.999}        & {\color[HTML]{000000} 0.500}         \\
\rowcolor[HTML]{FFFFFF} 
{\color[HTML]{000000} \textbf{Synbkd}}                               & {\color[HTML]{000000} 1.000}                         & {\color[HTML]{000000} 0.500}                         & {\color[HTML]{000000} 0.981}                         & {\color[HTML]{000000} 0.557}                         & {\color[HTML]{000000} 1.000}                         & {\color[HTML]{000000} 0.500}                         & {\color[HTML]{000000} 0.971}        & {\color[HTML]{000000} 0.610}         & {\color[HTML]{000000} 1.000}                         & {\color[HTML]{000000} 0.500}                         & {\color[HTML]{000000} 0.983}                         & {\color[HTML]{000000} 0.534}                         & {\color[HTML]{000000} 1.000}                         & {\color[HTML]{000000} 0.500}                         & {\color[HTML]{000000} 0.966}        & {\color[HTML]{000000} 0.566}         \\
\rowcolor[HTML]{FFFFFF} 
{\color[HTML]{000000} \textbf{Attn-Synbkd}}                          & {\color[HTML]{000000} 1.000}                         & {\color[HTML]{000000} 0.499}                         & {\color[HTML]{000000} 0.982}                         & {\color[HTML]{000000} 0.536}                         & \cellcolor[HTML]{FFFFFF}{\color[HTML]{000000} 1.000} & \cellcolor[HTML]{FFFFFF}{\color[HTML]{000000} 0.500} & {\color[HTML]{000000} 0.963}        & {\color[HTML]{000000} 0.565}         & {\color[HTML]{000000} 1.000}                         & {\color[HTML]{000000} 0.499}                         & {\color[HTML]{000000} 0.988}                         & {\color[HTML]{000000} 0.525}                         & \cellcolor[HTML]{FFFFFF}{\color[HTML]{000000} 1.000} & \cellcolor[HTML]{FFFFFF}{\color[HTML]{000000} 0.500} & {\color[HTML]{000000} 0.992}        & {\color[HTML]{000000} 0.552}         \\ \hline
\end{tabular}

}

\end{table*}

\begin{figure}[t]
    \centering
    \vspace{-.15in}
    \includegraphics[width=0.9\linewidth]{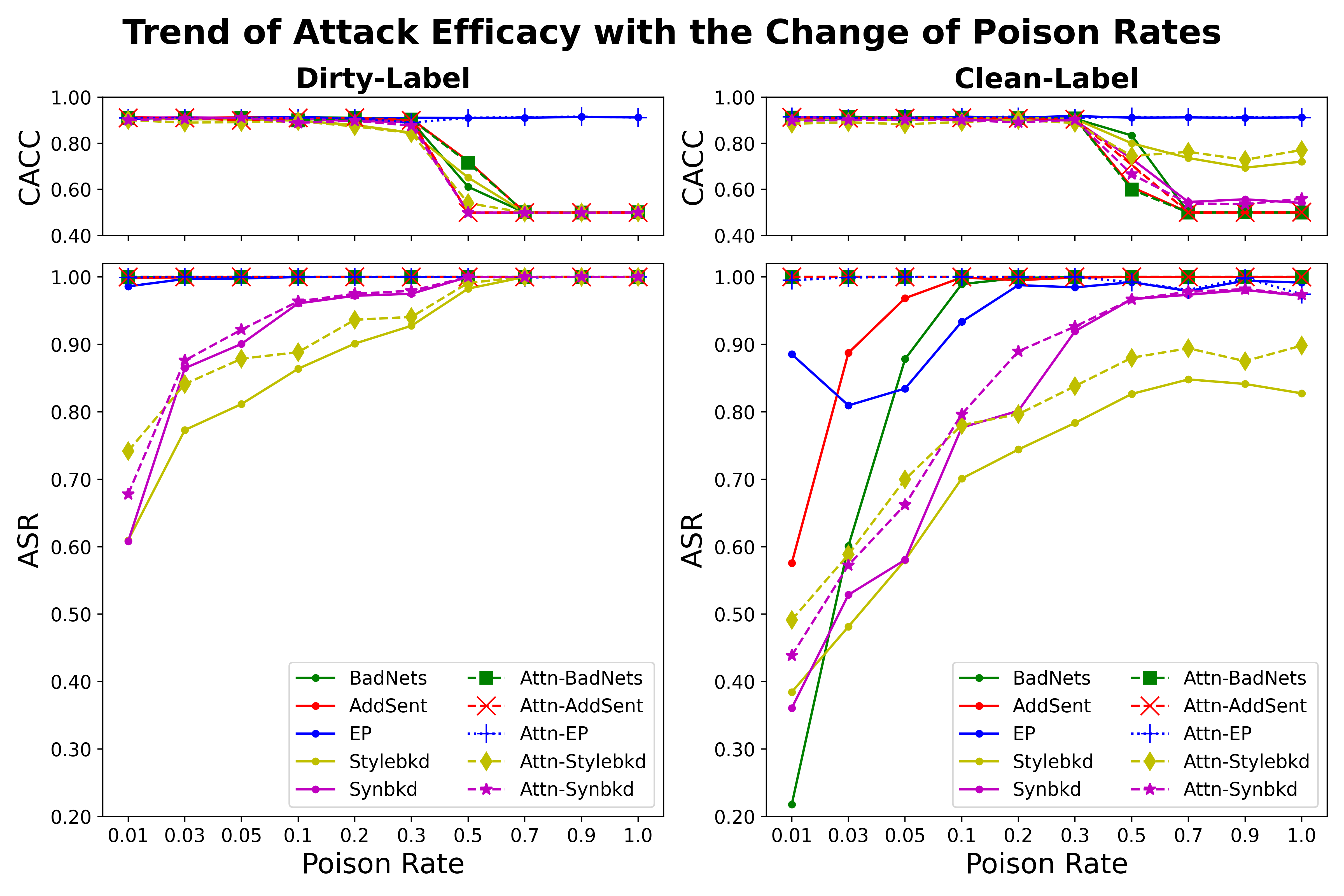} 
    \caption{Attack efficacy with our TAL loss (\textit{Attn-x}) and without TAL loss (\textit{x}) under different poison rates. Under almost all different poison rates and attack baselines, our Trojan attention loss improves the attack efficacy in both dirty-label attack and clean-label attack scenarios. Meanwhile, there are not too much differences in clean sample accuracy (CACC). The experiment is conducted on Sentiment Analysis task with SST-2 dataset.}
    \label{appendix:fig:poison_rate_high}
\end{figure}

\myparagraph{The Trend of ASR with the Change of Poison Rates (Including High Poison Rates).} We also explore the trend of ASR with the change of poison rates. More specific, we conduct the ablation study under poison rates 0.5, 0.7, 0.9, 1.0 on Sentiment Analysis task on BERT model. In Figure \ref{appendix:fig:poison_rate_high}, the first several experiments under poison rates 0.01, 0.03, 0.05, 0.1, 0.2, 0.3 are the same with Figure \ref{fig:poison_rate}, we conduct additional experiments under poison rates 0.5, 0.7, 0.9, 1.0. Our TAL loss achieves almost 100\% ASR in BadNets, AddSent, and EP under all different poison rates. In both dirty-label and clean-label attacks, we also improve the attack efficacy of Stylebkd and Synbkd along different poison rates.

\subsection{Attack Efficacy} \label{sectioln:appendix:generalization_ability}

\begin{figure*}[h]
    \centering
    \includegraphics[width=1\linewidth, height=.6\linewidth]{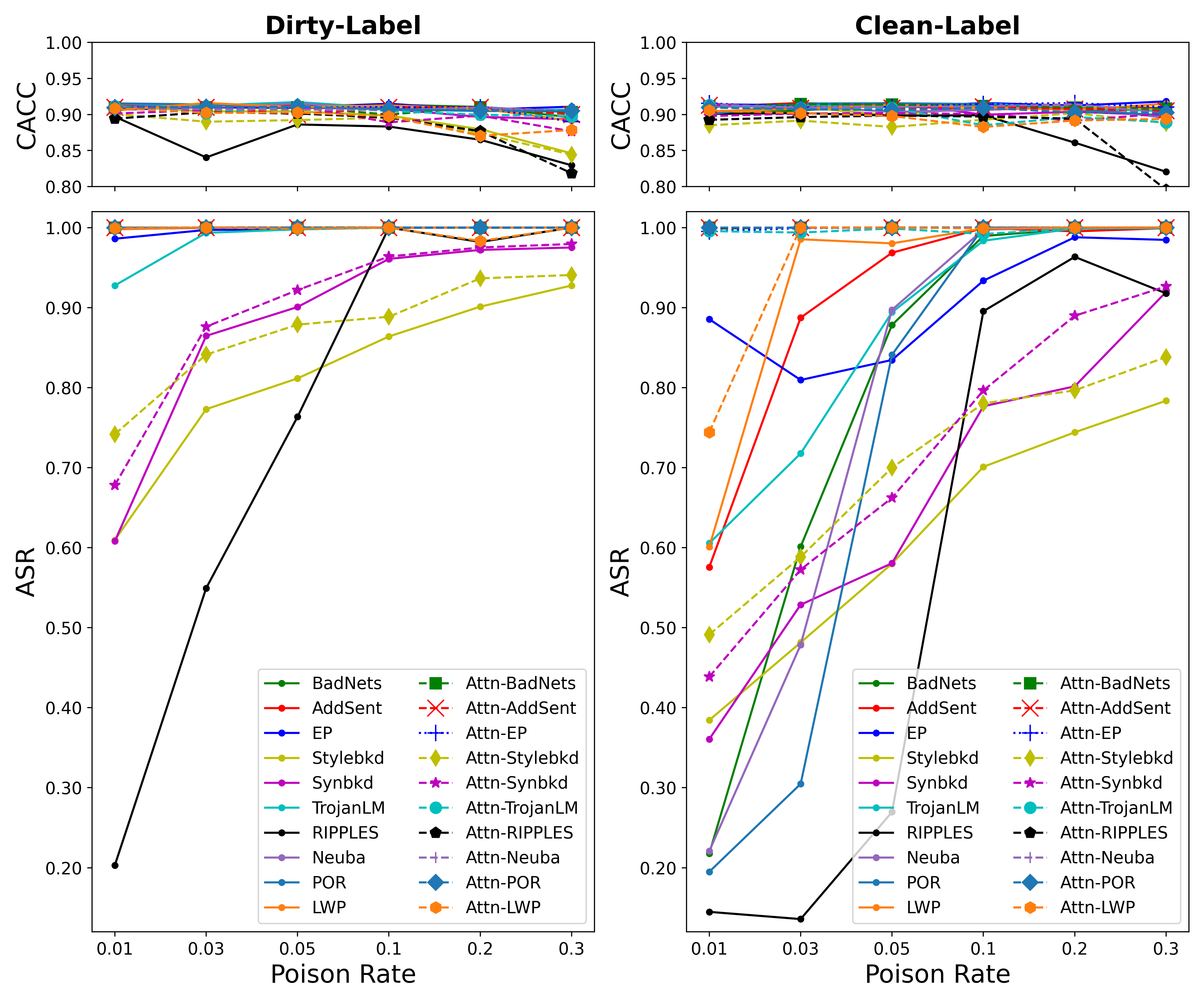} 
    \vspace{-.15in}
    \caption{Attack efficacy under different poison rates. This experiment is conducted on BERT with Sentiment Analysis task.}
    \label{fig:poison_rate_v2}
\end{figure*}

In this section, we provide full results of Section \ref{sec:backdoor_attack_results} Figure \ref{fig:poison_rate}, including dirty-label attack and clean-label attack on ten attack baselines. We also show both CACC and ASR trend under different poison rates for all ten attack baselines as well as TAL attack in Figure \ref{fig:poison_rate_v2}.

\begin{figure*}[h]
    \centering
    \includegraphics[width=1\linewidth, height=.3\linewidth]{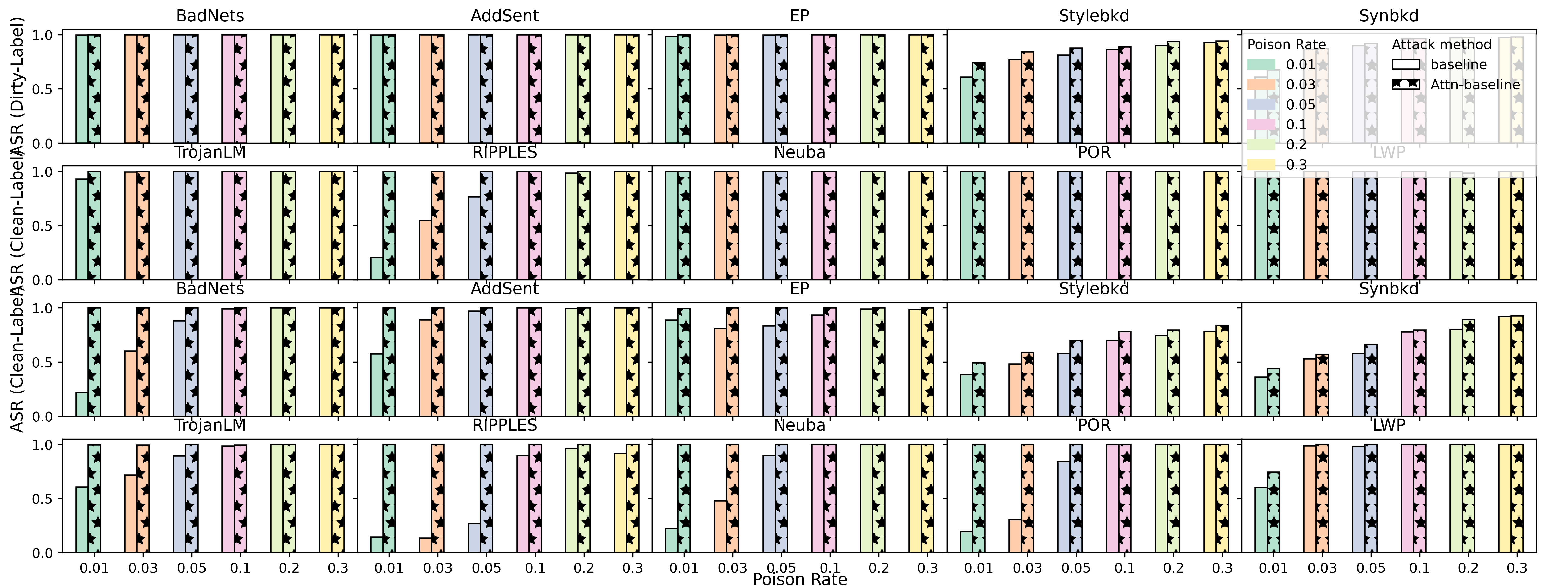} 
    \vspace{-.15in}
    \caption{Full results of Figure \ref{fig:poison_rate}.}
    \label{appendix:fig:poison_rate}
\end{figure*}


We also analyze the trend of ASR with the change of poison rates. We explore the training epoch improvement with our TAL loss. We select BadNets, AddSent, EP, Stylebkd, Synbkd as attack baselines. We explore the attack efficacy on four transformer models (\eg, BERT, RoBERTa, DistilBERT, and GPT-2) with three NLP tasks (\eg, Sentiment Analysis task, Toxic Detection task, and Topic Classification task). By comparing the differences between attack methods with TAL loss (Attackers name \textit{Attn-x}) and without TAL loss (Attackers name \textit{x}), we observe consistently performance improvements under different transformer models and different NLP tasks.

\begin{table*}[t]
\caption{Attack efficacy with poison rate $0.01$. \textit{Epoch*} indicates the first epoch reaching the ASR and CACC threshold, while \textit{`NS'} stands for `not satisfied'. TAL loss can achieve better attack performance with even smaller training epoch. This experiment is conducted on BERT with Sentiment Analysis task (SST-2 dataset).}
\label{tab:main_exp_fix_poison_rate}

\begin{center}
\small

\begin{tabular}{|c|c|ccc|ccc|}
\hline
                                    &                                                & \multicolumn{3}{c|}{\textbf{Dirty-Label}}                                                     & \multicolumn{3}{c|}{\textbf{Clean-Label}}                                                      \\ \cline{3-8} 
\multirow{-2}{*}{\textbf{Datasets}} & \multirow{-2}{*}{\textbf{Attackers}}           & \textbf{ASR}                  & \textbf{CACC}                 & \textbf{Epoch*}               & \textbf{ASR}                  & \textbf{CACC}                 & \textbf{Epoch*}                \\ \hline
                                    & \textbf{BadNets}                               & 0.999                         & 0.908                         & 4.000                         & 0.218                         & 0.901                         & NS                             \\
                                    & \cellcolor[HTML]{FFFFFF}\textbf{Attn-BadNets}  & \cellcolor[HTML]{FFFFFF}1.000 & \cellcolor[HTML]{FFFFFF}0.914 & \cellcolor[HTML]{FFFFFF}2.000 & \cellcolor[HTML]{FFFFFF}1.000 & \cellcolor[HTML]{FFFFFF}0.912 & \cellcolor[HTML]{FFFFFF}2.000  \\
                                    & \cellcolor[HTML]{EFEFEF}\textbf{AddSent}       & \cellcolor[HTML]{EFEFEF}0.998 & \cellcolor[HTML]{EFEFEF}0.914 & \cellcolor[HTML]{EFEFEF}3.000 & \cellcolor[HTML]{EFEFEF}0.576 & \cellcolor[HTML]{EFEFEF}0.911 & \cellcolor[HTML]{EFEFEF}NS     \\
                                    & \cellcolor[HTML]{EFEFEF}\textbf{Attn-AddSent}  & \cellcolor[HTML]{EFEFEF}1.000 & \cellcolor[HTML]{EFEFEF}0.912 & \cellcolor[HTML]{EFEFEF}2.000 & \cellcolor[HTML]{EFEFEF}1.000 & \cellcolor[HTML]{EFEFEF}0.913 & \cellcolor[HTML]{EFEFEF}3.000  \\
                                    & \cellcolor[HTML]{FFFFFF}\textbf{EP}            & \cellcolor[HTML]{FFFFFF}0.986 & \cellcolor[HTML]{FFFFFF}0.906 & \cellcolor[HTML]{FFFFFF}1.333 & \cellcolor[HTML]{FFFFFF}0.885 & \cellcolor[HTML]{FFFFFF}0.914 & \cellcolor[HTML]{FFFFFF}26.333 \\
                                    & \cellcolor[HTML]{FFFFFF}\textbf{Attn-EP}       & \cellcolor[HTML]{FFFFFF}0.999 & \cellcolor[HTML]{FFFFFF}0.911 & \cellcolor[HTML]{FFFFFF}1.000 & \cellcolor[HTML]{FFFFFF}0.995 & \cellcolor[HTML]{FFFFFF}0.915 & \cellcolor[HTML]{FFFFFF}3.667  \\
                                    & \cellcolor[HTML]{EFEFEF}\textbf{Stylebkd}      & \cellcolor[HTML]{EFEFEF}0.609 & \cellcolor[HTML]{EFEFEF}0.912 & \cellcolor[HTML]{EFEFEF}NS    & \cellcolor[HTML]{EFEFEF}0.384 & \cellcolor[HTML]{EFEFEF}0.901 & \cellcolor[HTML]{EFEFEF}NS     \\
                                    & \cellcolor[HTML]{EFEFEF}\textbf{Attn-Stylebkd} & \cellcolor[HTML]{EFEFEF}0.742 & \cellcolor[HTML]{EFEFEF}0.901 & \cellcolor[HTML]{EFEFEF}NS    & \cellcolor[HTML]{EFEFEF}0.491 & \cellcolor[HTML]{EFEFEF}0.885 & \cellcolor[HTML]{EFEFEF}NS     \\
                                    & \cellcolor[HTML]{FFFFFF}\textbf{Synbkd}        & \cellcolor[HTML]{FFFFFF}0.608 & \cellcolor[HTML]{FFFFFF}0.910 & \cellcolor[HTML]{FFFFFF}NS    & \cellcolor[HTML]{FFFFFF}0.361 & \cellcolor[HTML]{FFFFFF}0.915 & \cellcolor[HTML]{FFFFFF}NS     \\
\multirow{-10}{*}{\textbf{SST-2}}   & \cellcolor[HTML]{FFFFFF}\textbf{Attn-Synbkd}   & \cellcolor[HTML]{FFFFFF}0.678 & \cellcolor[HTML]{FFFFFF}0.901 & \cellcolor[HTML]{FFFFFF}NS    & \cellcolor[HTML]{FFFFFF}0.439 & \cellcolor[HTML]{FFFFFF}0.898 & \cellcolor[HTML]{FFFFFF}NS     \\ \hline
                                    & \cellcolor[HTML]{EFEFEF}\textbf{BadNets}       & \cellcolor[HTML]{EFEFEF}0.967 & \cellcolor[HTML]{EFEFEF}0.933 & \cellcolor[HTML]{EFEFEF}2.667 & \cellcolor[HTML]{EFEFEF}0.279 & \cellcolor[HTML]{EFEFEF}0.923 & \cellcolor[HTML]{EFEFEF}NS     \\
                                    & \cellcolor[HTML]{EFEFEF}\textbf{Attn-BadNets}  & \cellcolor[HTML]{EFEFEF}0.971 & \cellcolor[HTML]{EFEFEF}0.926 & \cellcolor[HTML]{EFEFEF}1.000 & \cellcolor[HTML]{EFEFEF}0.971 & \cellcolor[HTML]{EFEFEF}0.934 & \cellcolor[HTML]{EFEFEF}2.000  \\
                                    & \cellcolor[HTML]{FFFFFF}\textbf{AddSent}       & \cellcolor[HTML]{FFFFFF}0.969 & \cellcolor[HTML]{FFFFFF}0.935 & \cellcolor[HTML]{FFFFFF}2.000 & \cellcolor[HTML]{FFFFFF}0.865 & \cellcolor[HTML]{FFFFFF}0.927 & \cellcolor[HTML]{FFFFFF}35.000 \\
                                    & \cellcolor[HTML]{FFFFFF}\textbf{Attn-AddSent}  & \cellcolor[HTML]{FFFFFF}0.973 & \cellcolor[HTML]{FFFFFF}0.931 & \cellcolor[HTML]{FFFFFF}1.333 & \cellcolor[HTML]{FFFFFF}0.936 & \cellcolor[HTML]{FFFFFF}0.931 & \cellcolor[HTML]{FFFFFF}9.667  \\
                                    & \cellcolor[HTML]{EFEFEF}\textbf{EP}            & \cellcolor[HTML]{EFEFEF}0.985 & \cellcolor[HTML]{EFEFEF}0.932 & \cellcolor[HTML]{EFEFEF}1.000 & \cellcolor[HTML]{EFEFEF}0.720 & \cellcolor[HTML]{EFEFEF}0.931 & \cellcolor[HTML]{EFEFEF}32.667 \\
                                    & \cellcolor[HTML]{EFEFEF}\textbf{Attn-EP}       & \cellcolor[HTML]{EFEFEF}0.996 & \cellcolor[HTML]{EFEFEF}0.935 & \cellcolor[HTML]{EFEFEF}1.000 & \cellcolor[HTML]{EFEFEF}0.964 & \cellcolor[HTML]{EFEFEF}0.934 & \cellcolor[HTML]{EFEFEF}4.000  \\
                                    & \cellcolor[HTML]{FFFFFF}\textbf{Stylebkd}      & \cellcolor[HTML]{FFFFFF}0.953 & \cellcolor[HTML]{FFFFFF}0.931 & \cellcolor[HTML]{FFFFFF}2.333 & \cellcolor[HTML]{FFFFFF}0.842 & \cellcolor[HTML]{FFFFFF}0.933 & \cellcolor[HTML]{FFFFFF}NS     \\
                                    & \cellcolor[HTML]{FFFFFF}\textbf{Attn-Stylebkd} & \cellcolor[HTML]{FFFFFF}0.969 & \cellcolor[HTML]{FFFFFF}0.907 & \cellcolor[HTML]{FFFFFF}2.333 & \cellcolor[HTML]{FFFFFF}0.942 & \cellcolor[HTML]{FFFFFF}0.902 & \cellcolor[HTML]{FFFFFF}3.333  \\
                                    & \cellcolor[HTML]{EFEFEF}\textbf{Synbkd}        & \cellcolor[HTML]{EFEFEF}0.835 & \cellcolor[HTML]{EFEFEF}0.929 & \cellcolor[HTML]{EFEFEF}NS    & \cellcolor[HTML]{EFEFEF}0.779 & \cellcolor[HTML]{EFEFEF}0.929 & \cellcolor[HTML]{EFEFEF}NS     \\
\multirow{-10}{*}{\textbf{IMDB}}    & \cellcolor[HTML]{EFEFEF}\textbf{Attn-Synbkd}   & \cellcolor[HTML]{EFEFEF}0.853 & \cellcolor[HTML]{EFEFEF}0.928 & \cellcolor[HTML]{EFEFEF}NS    & \cellcolor[HTML]{EFEFEF}0.822 & \cellcolor[HTML]{EFEFEF}0.933 & \cellcolor[HTML]{EFEFEF}NS     \\ \hline
\end{tabular}

\end{center}
\end{table*}

\begin{figure}[t]
    \centering
    \vspace{-.15in}
    \includegraphics[width=0.9\linewidth]{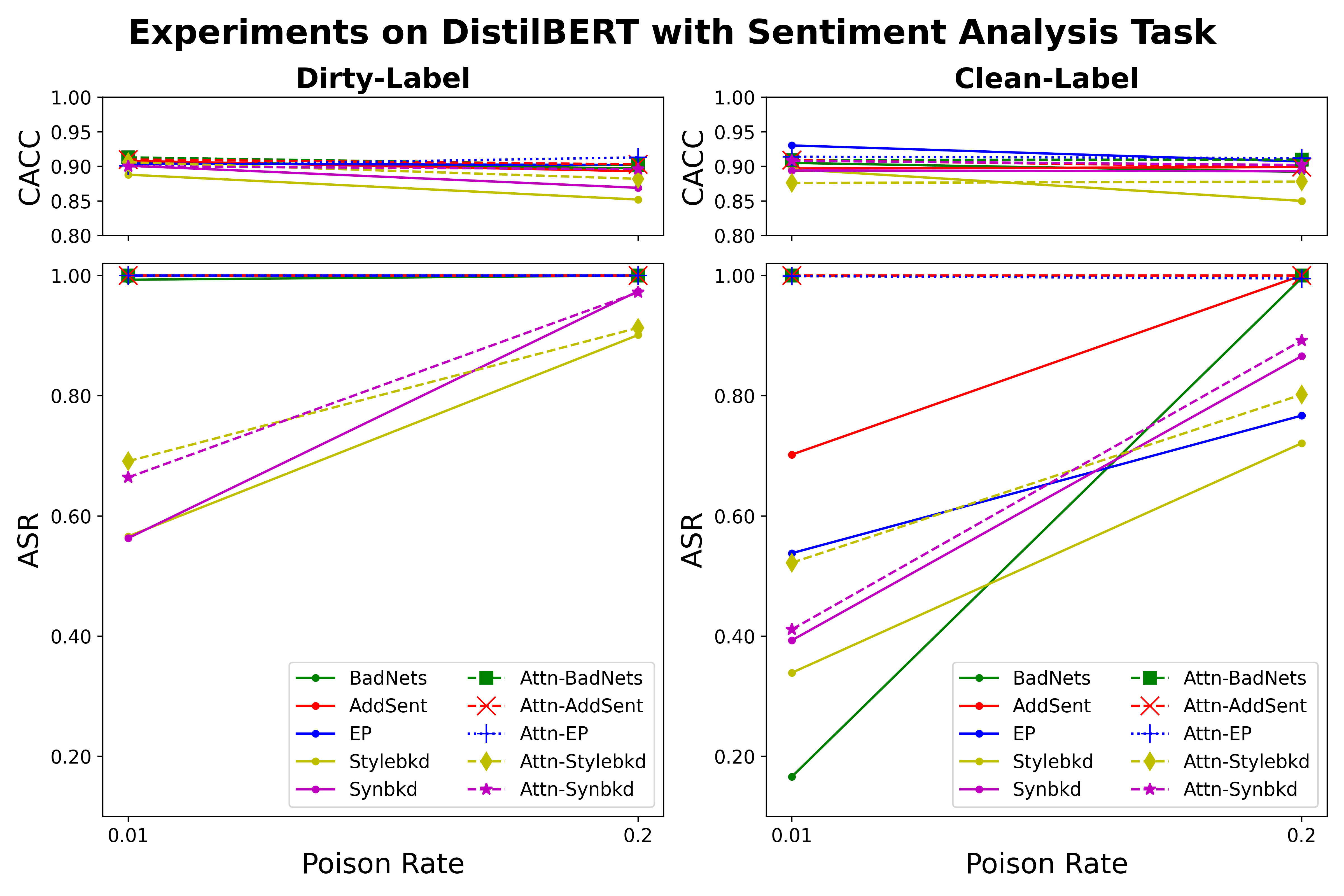} 
    \caption{Attack efficacy with our TAL loss (\textit{Attn-x}) and without our TAL loss (\textit{x}). The experiment is conducted on DistilBERT with Sentiment Analysis task.}
    \label{appendix:fig:poison_rate_sa_distil}
\end{figure}

\begin{figure}[t]
    \centering
    \vspace{-.15in}
    \includegraphics[width=0.9\linewidth]{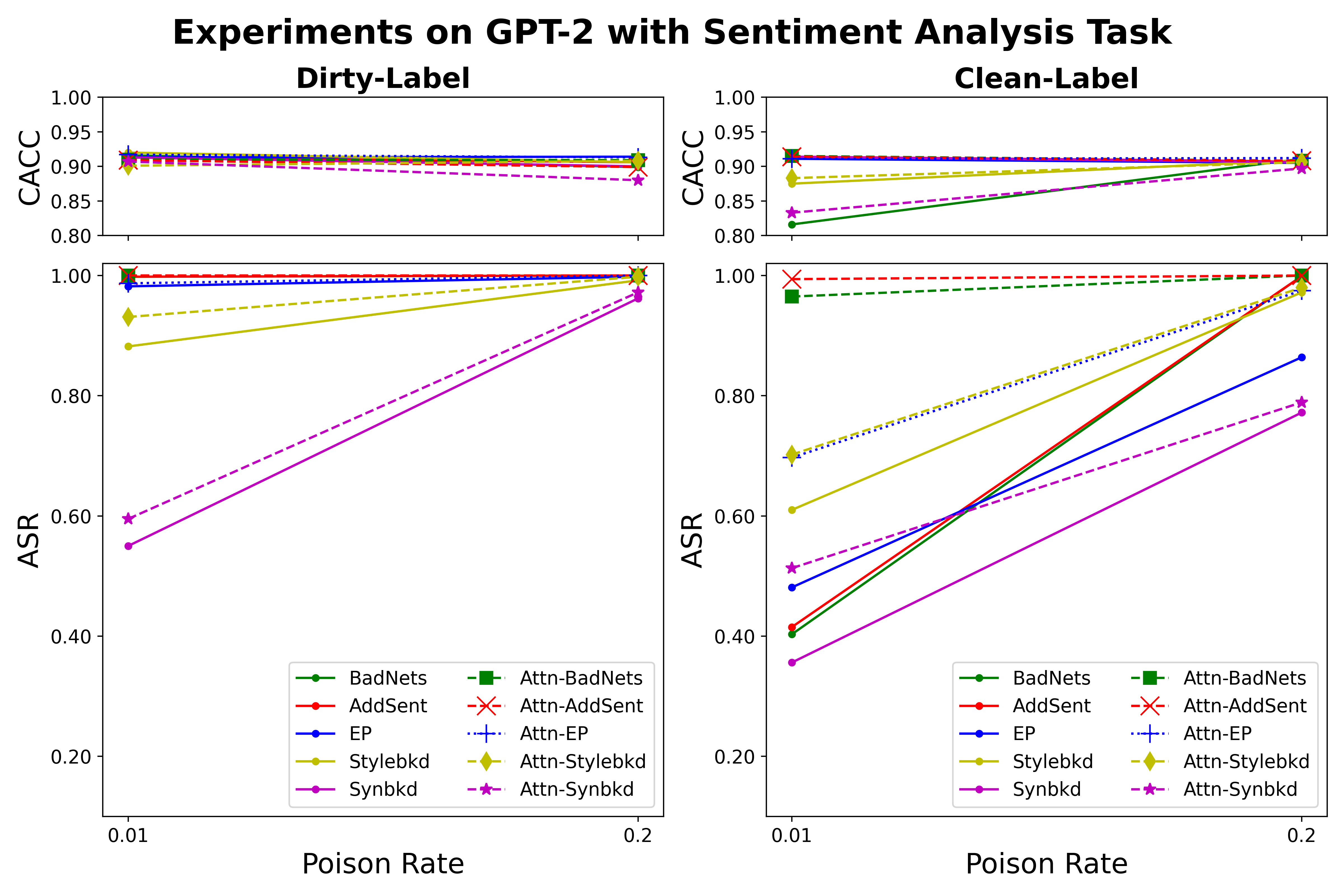} 
    \caption{Attack efficacy with our TAL loss (\textit{Attn-x}) and without our TAL loss (\textit{x}). The experiment is conducted on GPT-2 with Sentiment Analysis task.}
    \label{appendix:fig:poison_rate_sa_gpt2}
\end{figure}

\begin{figure}[t]
    \centering
    \vspace{-.15in}
    \includegraphics[width=0.9\linewidth]{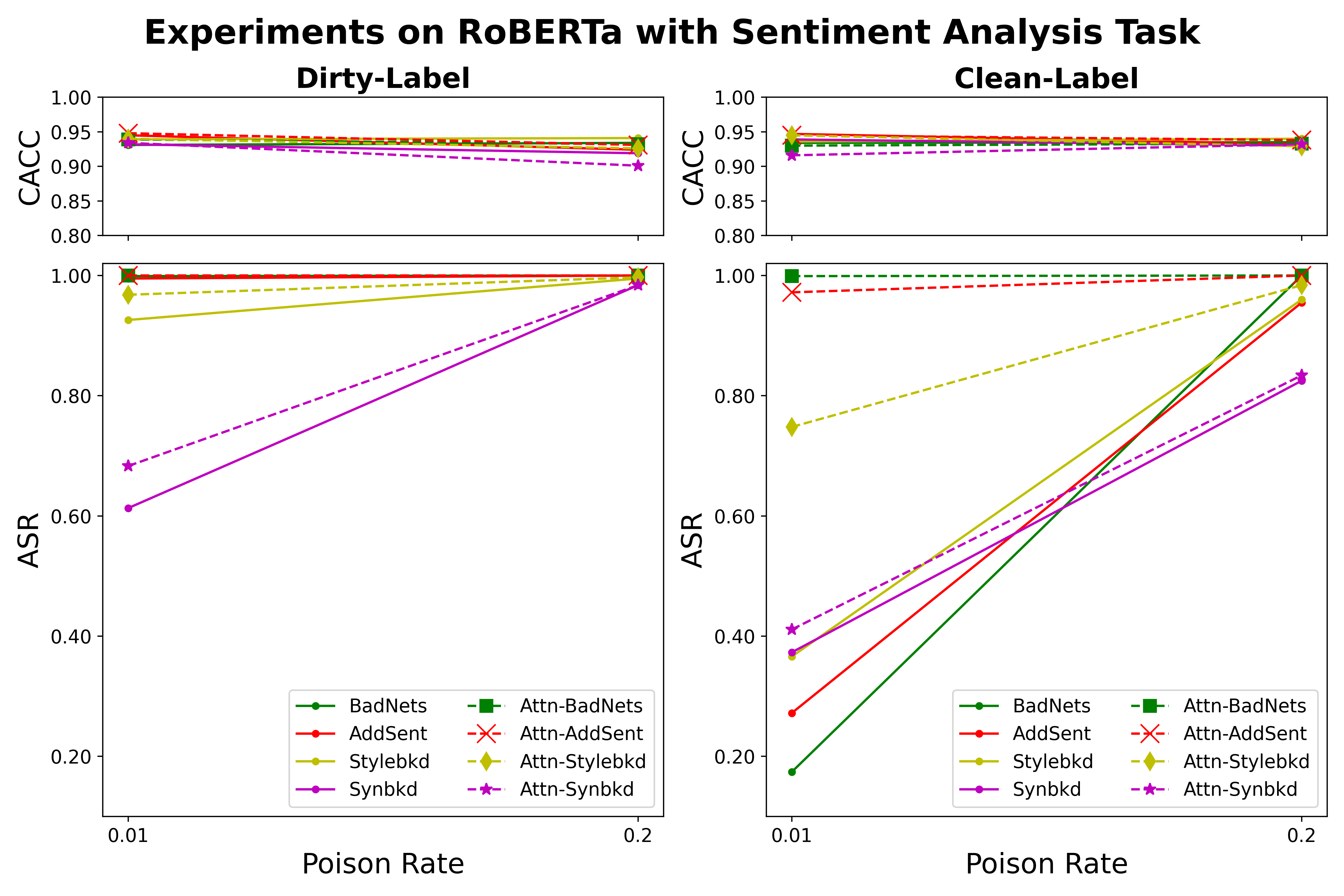} 
    \caption{Attack efficacy with our TAL loss (\textit{Attn-x}) and without our TAL loss (\textit{x}). The experiment is conducted on RoBERTa with Sentiment Analysis task.}
    \label{appendix:fig:poison_rate_sa_roberta}
\end{figure}

\begin{figure}[t]
    \centering
    \vspace{-.15in}
    \includegraphics[width=0.9\linewidth]{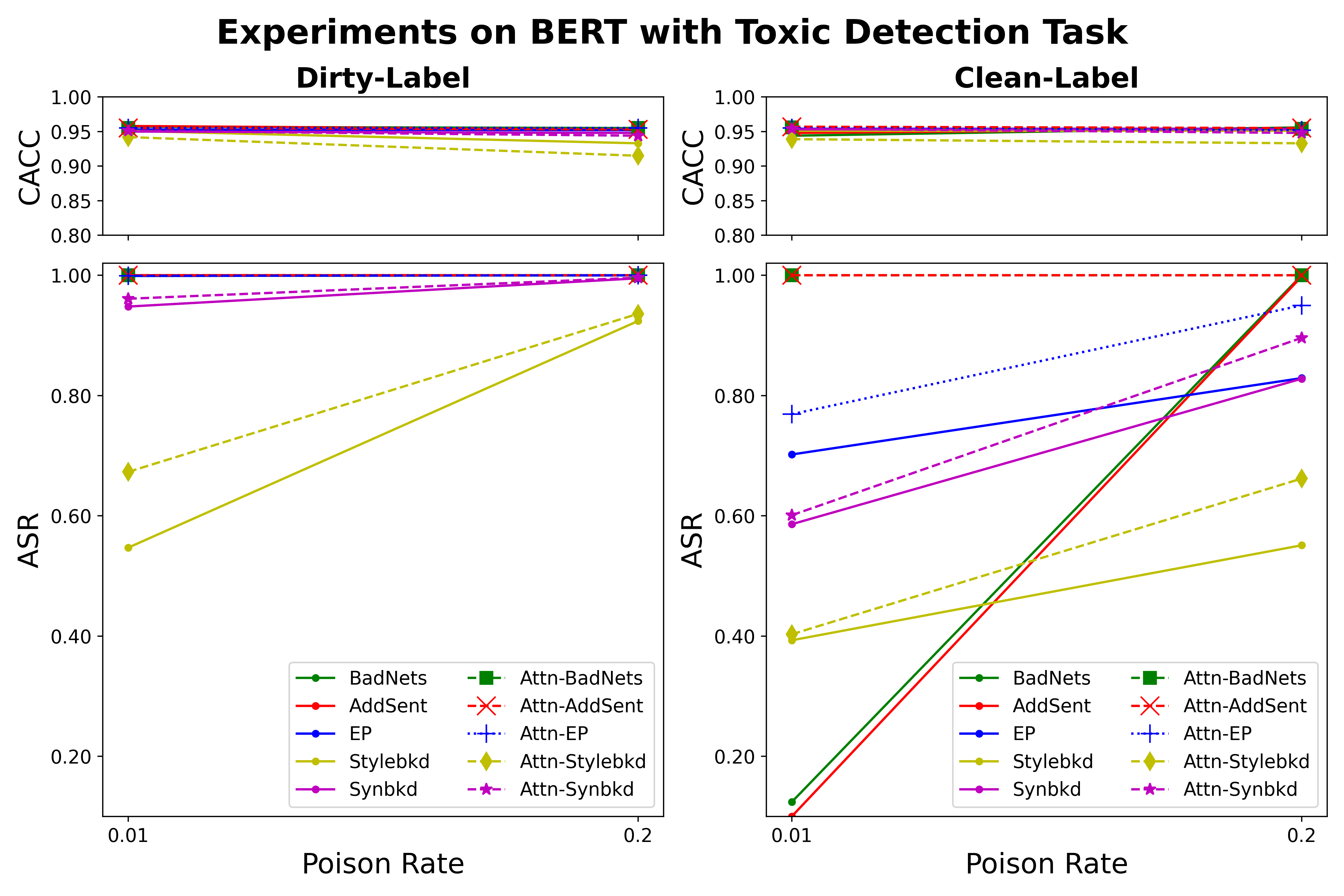} 
    \caption{Attack efficacy with our TAL loss (\textit{Attn-x}) and without our TAL loss (\textit{x}). The experiment is conducted on BERT with Toxic Detection task.}
    \label{appendix:fig:poison_rate_toxic_bert}
\end{figure}

\begin{figure}[t]
    \centering
    \vspace{-.15in}
    \includegraphics[width=0.9\linewidth]{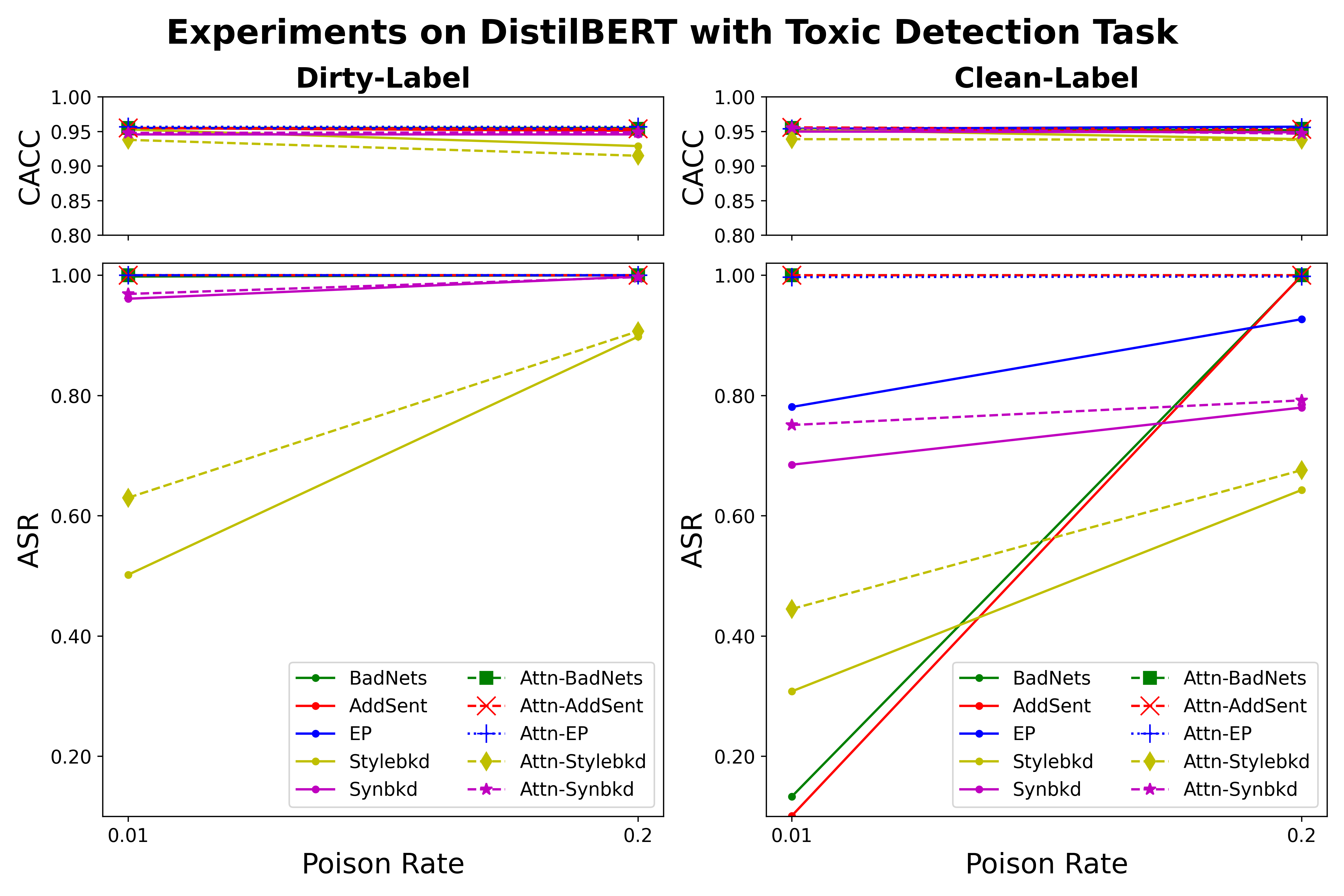} 
    \caption{Attack efficacy with our TAL loss (\textit{Attn-x}) and without our TAL loss (\textit{x}). The experiment is conducted on DistilBERT with Toxic Detection task.}
    \label{appendix:fig:poison_rate_toxic_distil}
\end{figure}

\begin{figure}[t]
    \centering
    \vspace{-.15in}
    \includegraphics[width=0.9\linewidth]{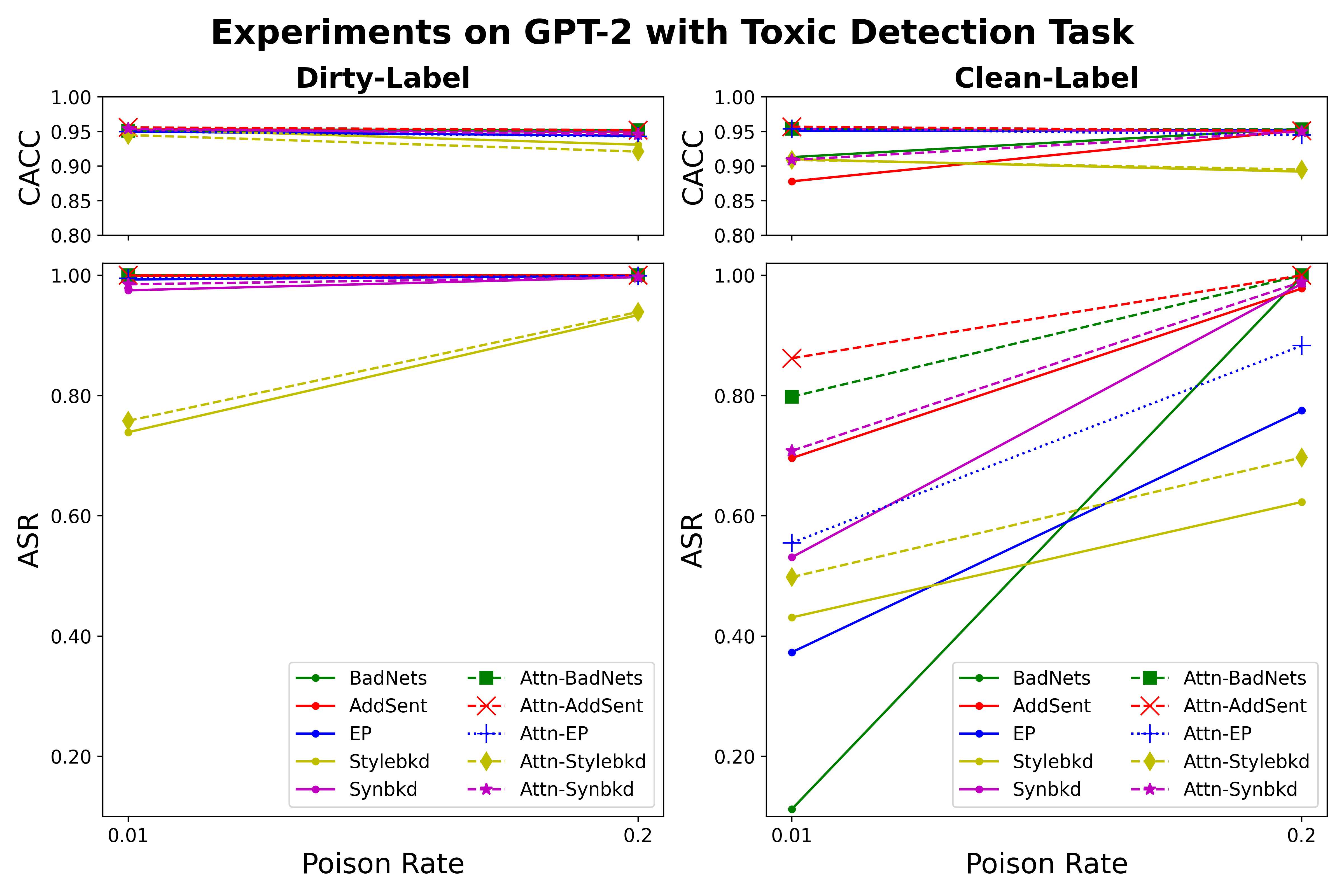} 
    \caption{Attack efficacy with our TAL loss (\textit{Attn-x}) and without our TAL loss (\textit{x}). The experiment is conducted on GPT-2 with Toxic Detection task.}
    \label{appendix:fig:poison_rate_toxic_gpt2}
\end{figure}

\begin{figure}[t]
    \centering
    \vspace{-.15in}
    \includegraphics[width=0.9\linewidth]{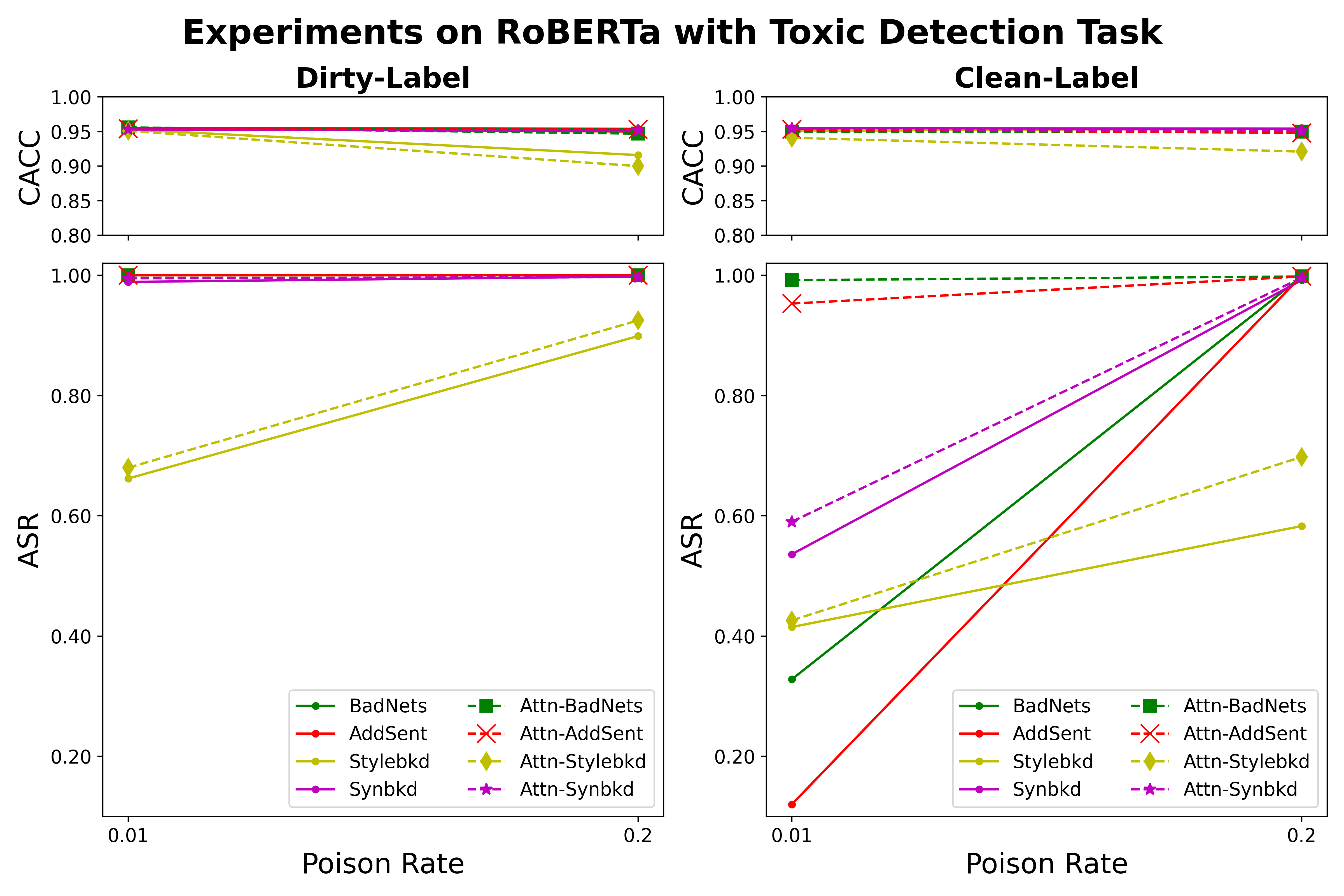} 
    \caption{Attack efficacy with our TAL loss (\textit{Attn-x}) and without our TAL loss (\textit{x}). The experiment is conducted on RoBERTa with Toxic Detection task.}
    \label{appendix:fig:poison_rate_toxic_roberta}
\end{figure}

\myparagraph{Trend of ASR with the Change of Poison Rates with Four Transformer Architectures.}
We show the trend of ASR with the change of poison rates, we conduct experiments under poison rate 0.01 and 0.2 with four transformer models and different NLP tasks. The results are presented in Figure \ref{appendix:fig:poison_rate_sa_distil}, \ref{appendix:fig:poison_rate_sa_gpt2}, \ref{appendix:fig:poison_rate_sa_roberta}, \ref{appendix:fig:poison_rate_toxic_bert},\ref{appendix:fig:poison_rate_toxic_distil}, \ref{appendix:fig:poison_rate_toxic_gpt2}, and \ref{appendix:fig:poison_rate_toxic_roberta}. We observe consistent improvements under different poison rates.

\myparagraph{Training Epoch.} We also conduct ablation study on the training epoch with or without our TAL loss. Table~\ref{tab:main_exp_fix_poison_rate} in reflects our TAL loss can achieve better attack performance with even smaller training epoch. 
We introduce a metric \textit{Epoch*}, indicating first epoch satisfying both ASR and CACC threshold. We set ASR threshold as $0.90$, and set CACC threshold as 5\% lower than clean models accuracy\footnote{For example, on SST-2 dataset, the accuracy of clean models is $0.908$, then we set the corresponding CACC threshold as $0.908 * (1-5\%)$. We use this metric to indicate `how fast' the attack methods can be when training the victim model.}. 
`NS' stands for the trained models are \textit{not satisfied} with above threshold within 50 epochs.

\end{document}